\documentclass[pdflatex,sn-mathphys-num]{sn-jnl}

\usepackage{graphicx}%
\usepackage{multirow}%
\usepackage{amsmath,amssymb,amsfonts}%
\usepackage{amsthm}%
\usepackage{mathrsfs}%
\usepackage[title]{appendix}%
\usepackage{xcolor}%
\usepackage{textcomp}%
\usepackage{manyfoot}%
\usepackage{booktabs}%
\usepackage{algorithm}%
\usepackage{algorithmicx}%
\usepackage{algpseudocode}%
\usepackage{listings}%
\usepackage{tabularx}
\usepackage[T5]{fontenc}
\usepackage[utf8]{inputenc}
\usepackage{makecell}
\usepackage{pifont}
\usepackage{tablefootnote}
\usepackage{subcaption}
\usepackage{float}
\usepackage{dblfloatfix}
\usepackage{arydshln}

\theoremstyle{thmstyleone}%
\theoremstyle{thmstyletwo}%

\theoremstyle{thmstylethree}%

\begin{document}

\title[LiGT: Layout-infused Generative Transformer for Visual Question Answering on Vietnamese Receipts]{LiGT: Layout-infused Generative Transformer for Visual Question Answering on Vietnamese Receipts}


\author[1,2]{\fnm{Thanh-Phong} \sur{Le}}\email{21520395@gm.uit.edu.vn}

\author[1,2]{\fnm{Trung Le Chi} \sur{Phan}}\email{21522725@gm.uit.edu.vn}

\author[1,2]{\fnm{Nghia Hieu} \sur{Nguyen}}\email{nghiangh@uit.edu.vn}

\author*[1,2]{\fnm{Kiet Van} \sur{Nguyen}}\email{kietnv@uit.edu.vn}

\affil[1]{\orgdiv{Faculty of Information Science and Engineering}, \orgname{University of Information Technology}, \orgaddress{\city{Ho Chi Minh City}, \country{Vietnam}}}

\affil[2]{\orgname{Vietnam National University}, \orgaddress{\city{Ho Chi Minh City}, \country{Vietnam}}}


\abstract{\textbf{Purpose:} Document Visual Question Answering (document VQA) challenges multimodal systems to holistically handle textual, layout, and visual modalities to provide appropriate answers. Document VQA has gained popularity in recent years due to the increasing amount of documents and the high demand for digitization. Nonetheless, most of document VQA datasets are developed in high-resource languages such as English.
 
\textbf{Methods:} In this paper, we present ReceiptVQA (\textbf{Receipt} \textbf{V}isual \textbf{Q}uestion \textbf{A}nswering), the initial large-scale document VQA dataset in Vietnamese dedicated to receipts, a document kind with high commercial potentials. The dataset encompasses \textbf{9,000+} receipt images and \textbf{60,000+} manually annotated question-answer pairs. In addition to our study, we introduce LiGT (\textbf{L}ayout-\textbf{i}nfused \textbf{G}enerative \textbf{T}ransformer), a layout-aware encoder-decoder architecture designed to leverage embedding layers of language models to operate layout embeddings, minimizing the use of additional neural modules.
 
\textbf{Results:} Experiments on ReceiptVQA show that our architecture yielded promising performance, achieving competitive results compared with outstanding baselines. Furthermore, throughout analyzing experimental results, we found evident patterns that employing encoder-only model architectures has considerable disadvantages in comparison to architectures that can generate answers. We also observed that it is necessary to combine multiple modalities to tackle our dataset, despite the critical role of semantic understanding from language models.
 
\textbf{Conclusion:} We hope that our work will encourage and facilitate future development in Vietnamese document VQA, contributing to a diverse multimodal research community in the Vietnamese language.}

\keywords{Visual Question Answering, Low-resource Languages, Multimodal Learning, Document Understanding}



\maketitle

\section{Introduction} \label{sec:Introduction}

Visual Question Answering (VQA) \cite{Antol_2015_ICCV} is a multimodal task aiming to create systems capable of answering questions based on visual information. VQA research, over the past few years, has gained remarkable popularity, involving various domains and variants \cite{SHARMA2021104327,Fromimagetolanguage}. Correspondingly, there has been an increasing interest in VQA on document images (document VQA) \cite{mathew2021docvqa} due to the proliferation of documents, e.g. forms, receipts, and resumes, and the high digitization demand in modern environments. 

Document VQA has become not only a crucial task for the industry but also a multimodal challenge, wherein a machine needs to simultaneously process visual, textual, and layout information from document images while referring to the questions. Document images are specifically characterized by a remarkable density of text and a variety of formats. The text is structurally allocated in different areas, constituting a diversity of 2D spatial layouts depending on its document type. For this reason, studies in document VQA have proposed many datasets, often designed to tackle some particular kinds of documents \cite{mathew2021docvqa,tanaka2021visualmrc,InfographicVQA,masry2022chartqa}.

Among various documents, receipts, the papers proving that payment or delivery has been completed, have great commercial potential. Receipt images can be utilized for many purposes, such as efficient archiving on family/personal scales, document retrieval, and document analytics. To this end, there have been studies focusing on localizing, extracting, and parsing text on receipt images \cite{huang2019sroie,park2019cord}. However, in practice, most individuals opt for interactive features such as asking questions to get particular information rather than seeking texts extracted from the image. Thus, the VQA task stands as a potential extension toward user-friendly applications and services for documents, specifically for receipts.

Recently, inspired by the rapid development of multimodal learning (MML), many studies have tackled VQA task in Vietnamese, a low-resource language, encompassing dataset creations \cite{vivqa,openvivqa}, competitions \cite{Luu_Thuy_Nguyen_2023}, and methodologies \cite{phuc2020vican,openvivqa,tran2023bartphobeit}. Nonetheless, much of the current literature pays particular attention to either conventional VQA \cite{Antol_2015_ICCV} or scene text VQA \cite{singh2019towards,biten2019scene}, and there have been few attempts to investigate VQA on document images.

In this study, we introduce a large-scale Vietnamese document VQA dataset for receipts, called ReceiptVQA, including \textbf{64,812} question-answer (QA) pairs annotated from \textbf{9,500} receipt images. All dataset samples are manually annotated by a large resource of crowd-workers, and investigated on a scheme of answering natural questions about receipt content by using verbatim text spans extracted from their images. In addition, most of the images are \textit{unscanned} and captured by mobile phones that align with the context of users analyzing documents with portable devices. Also, further statistical analysis of the dataset found its specific features and interesting linguistic phenomena when people inquire about receipt content.

In addition to our work, we present \textbf{L}ayout-\textbf{i}nfused \textbf{G}enerative \textbf{T}ransformer (LiGT), a novel encoder-decoder transformer-based architecture extending the layout understanding capability of pretrained language models. Due to the text-dense characteristic of the document domain, we perceive that the relationship between texts and their 2D locations could also be associated with the 2D areas containing them. For this reason, instead of using numeric locations like previous layout approaches \cite{LayoutLM,layoutlmv2,tilt,docformer,udop,docformerv2}, we separate the examined 2D area into quarters and transform the OCR (Optical Character Recognition) texts' locations into hashed values for representing the area they belong to. The process is recursively executed throughout multiple hashing levels, constituting layout representations from more general to more specific. Furthermore, as pretrained language models could endure instability when being added with new neural modules, we minimize the usage of additional neural factors via leveraging pretrained embedding layers to operate 2D layout embeddings. This approach is projected to enhance the feasibility of our architecture, facilitating the adaptability to process the layout modality.

Our principal contributions are as follows.
\begin{enumerate}
\item We propose ReceiptVQA, the first large-scale document VQA dataset for the Vietnamese multimodal learning community, manually annotated for Vietnamese VQA on the receipt domain. Our dataset is publicly available exclusively for research purposes \footnote{We will provide the link as soon as our paper is accepted}. 

\item We developed a novel architecture, LiGT, infusing layout understanding ability into a pretrained encoder-decoder model while minimizing additional neural modules. Experimental evaluation shows our model achieved competitive results compared with other multimodal baselines. Furthermore, employing our LiGT's layout component on encoder-only models also yielded remarkable enhancements.

\item Based on the experimental results, we observed that models capable of generating answers superseded those using encoder-only methods. 

\item In addition, while semantic understanding plays a vital role in tackling our dataset, applying additional modalities shows substantial performance compared to using only text modality.

\end{enumerate}

In the following sections, we first provide information about related research in Section \ref{sec:RelatedWorks}. Then, Section \ref{sec:RVQADataset} presents our ReceiptVQA dataset and dataset analysis. Our proposed method is illustrated in Section \ref{sec:ProposedMethod}. Subsequently, we present our evaluation process in Section \ref{sec:Evaluation}, including chosen baselines, metrics, experimental results, and in-depth analysis. Finally, our conclusions and future plans are conveyed in Section \ref{sec:Conclusions}.


\section{Related Works} \label{sec:RelatedWorks}

In our research scope, the ReceiptVQA task is defined as extractive document VQA, wherein given a question written in natural language and a receipt image, a system answers by using verbatim text spans extracted from the image. The following subsections provide a brief overview of VQA in the document domain, receipt understanding, and current VQA research in Vietnamese. 


\subsection{Document VQA dataset} \label{sec:rel_docvqa_data}

Many VQA datasets on document images have been introduced in various document types. DocVQA \cite{mathew2021docvqa} encompasses images of industry/business documents, such as letters, forms, tables, and graphics. VisualMRC \cite{tanaka2021visualmrc} contains images that are screenshots of web pages. InfographicVQA \cite{InfographicVQA} and ChartQA \cite{masry2022chartqa} particularly focus on diagrams.

Furthermore, some studies extend the current scale of visual input from single to multiple images. SlideVQA \cite{SlideVQA2023} requires machines to handle a slide deck composed of multiple images, while PDF-VQA \cite{pdfvqa} challenges systems to elicit information from multiple pages of a full PDF document collected from published scientific articles. Also, recently, DUDE \cite{dude} presents as not only a \textit{multi-industry, multi-domain, and multi-page} VQA dataset but also a challenging benchmark with various types of questions and answers, aiming to provide a more real-world evaluation scheme for current and future methodologies.

\subsection{Document VQA methodology} \label{sec:rel_docvqa_med}

Generally speaking, there are two approaches to producing document VQA answers: extractive and generative. The extractive approach (or sequence labeling) produces answers by marking out the start and end indices of a sub-sequence from the context, which is serialized OCR tokens in reading order. This approach has largely been applied via BERT-like layout-aware models, which are pioneered by LayoutLM \cite{LayoutLM}. The introduction of the following pretrained layout models, e.g., LayoutLMv2 \cite{layoutlmv2}, Docformer \cite{docformer}, and LayoutLMv3 \cite{layoutlmv3}, has shown a remarkable enhancement over the past few years.

Nevertheless, as Powalski et al. \cite{tilt} addressed, the extractive approach is not only sensitive to error-prone OCR systems, and output tagging procedures but also inflexible in terms of producing answers. Therefore, the generative approach has been investigated with the addition of a decoder module. With an encoder-decoder backbone (often a T5-based architecture \cite{t5model}), generative models have the ability to auto-regressively generate answers by a \textit{built-in} vocabulary, which is not solely dependent on the OCR token inputs. Therefore, many architectures and pretrained models, such as, TILT \cite{tilt}, UDOP \cite{udop}, and Docformerv2 \cite{docformerv2}, have been presented and achieved state-of-the-art performance on strong benchmarks.

In this paper, we conduct experiments on outstanding baselines based on the extractive approach and generative approach and assess their feasibility on our dataset. Additionally, regarding the experiment scope, we mainly focus on the understanding of document content. Thus, OCR information is first extracted by using an off-the-shelf OCR engine and then used for training the examined models.


\subsection{Receipt Understanding} \label{sec:rel_ru}

In recent years, research on receipts has been primarily addressed in the Optical Character Recognition (OCR) field, specifically with the two datasets SROIE \cite{huang2019sroie} and CORD \cite{park2019cord}. SROIE \cite{huang2019sroie}, considered the first public receipt dataset, contains 1,000 scanned receipt images enclosed by special challenges, e.g., poor paper/printing quality, low-resolution scanners, long texts, and small font sizes. Subsequently, CORD \cite{park2019cord} was proposed for the post-OCR parsing task with more than 11,000 Indonesian receipt images. In addition, a receipt dataset for the VQA task, named Receipt AVQA, was introduced for the RECEIPT-AVQA-2023 shared task \cite{ravqa}, encompasses 21,835 English questions annotated from 1,957 receipt images.

Regarding the Vietnamese language, studies on receipt images, likewise, primarily investigate OCR-related tasks. The Mobile Captured Receipt Recognition Challenge (MC-OCR) \cite{mcocr} at the RIVF conference 2021 drew the attention of 105 participants and about 1,285 submission entries. MC-OCR used 2,436 Vietnamese mobile captured receipts to create the dataset that challenges two tasks, \textit{Image Quality Assessment} and \textit{Key Information Extraction}. Recently, in 2024, UIT-MLReceipts dataset was proposed for extracting key information on more than 2,000 receipt images annotated with four classes of bounding box, which are similar to MC-OCR \textit{Key Information Extraction} classes.


\subsection{VQA in Vietnamese} \label{sec:rel_vqavi}

Vietnamese VQA has been popularized since the introduction of the ViVQA dataset \cite{vivqa} in 2019. ViVQA dataset, with 10,328 images and 15,000 QA pairs, was conducted in a semi-automatic process, wherein questions and answers from the COCO-QA dataset \cite{cocoqa} were translated to Vietnamese before being revised by crowd-workers. Successively, the first large-scale and manually annotated dataset for VQA in Vietnamese, called OpenViVQA \cite{openvivqa}, was published encompassing 11,199 images of Vietnamese natural scenes and 37,914 open-ended question-answer (QA) pairs. OpenViVQA authors also tackled the scene-text property of the image by classifying each QA pair into \textit{Non-text QA} or \textit{Text QA} label and evaluating strong scene-text VQA baselines as well as their proposed architectures. Besides, the EVJVQA Challenge \cite{Luu_Thuy_Nguyen_2023} of VLSP 2022, handling VQA task in three languages English, Vietnamese, and Japanese, and the VLSP 2023 challenge on Visual Reading Comprehension for Vietnamese\footnote{\url{https://vlsp.org.vn/vlsp2023/eval/vrc}}, tackling the scene text factor in images, have gained attention from Vietnamese research community. Meanwhile, studies on methodologies for VQA in Vietnamese have started to be investigated \cite{phuc2020vican,nguyen2023pat,tran2023bartphobeit}.

To the best of our knowledge, ReceiptVQA is the initial manually-annotated dataset to explore Vietnamese VQA in document images, or receipt images in particular. We hope that our work will encourage and facilitate future extensions and contributions in Vietnamese VQA for document domains. 


\section{ReceiptVQA Dataset} \label{sec:RVQADataset}

\begin{figure}[hbt]
    \centering
    \includegraphics[width=1\linewidth]{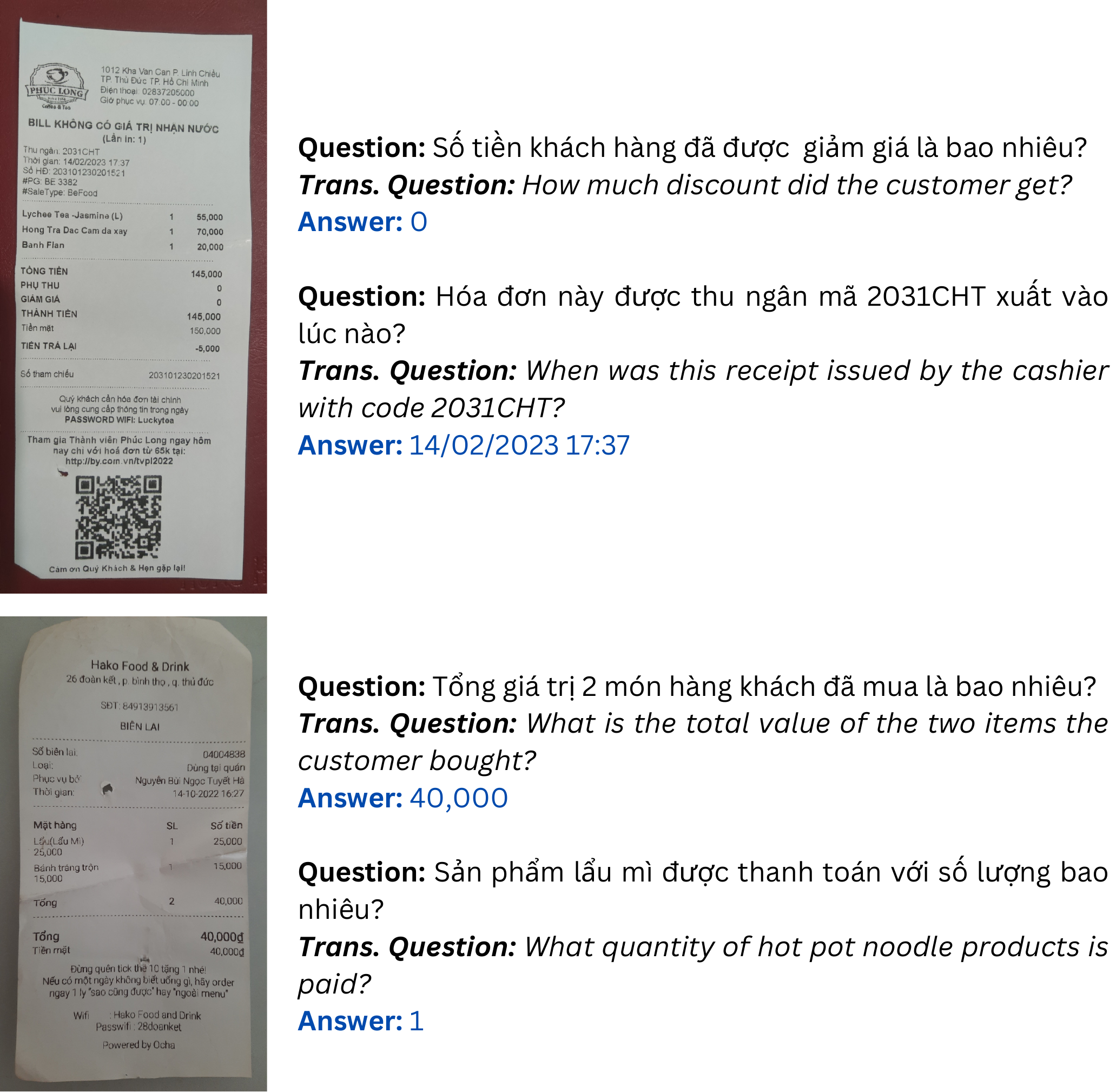}
    \caption{Examples of ReceiptVQA samples}
    \label{fig:example}
\end{figure}

Figure \ref{fig:example} shows examples of ReceiptVQA. Each sample of our dataset includes a Vietnamese receipt image, a question inquired by natural language in Vietnamese, and an answer of text spans extracted from the image. In this section, we illustrate the creation process, question classification, and statistical analysis of our dataset.

\subsection{Dataset Creation} \label{sec:datacre}

\begin{figure*}
  \begin{minipage}[c]{0.17\linewidth}
    \centering
    \caption{
       Overview of ReceiptVQA dataset creation
    } \label{fig:pipeline}
  \end{minipage}
  \hspace{0.02\linewidth}
  \begin{minipage}[c]{0.75\linewidth}
    \centering
    \includegraphics[width=0.8\linewidth]{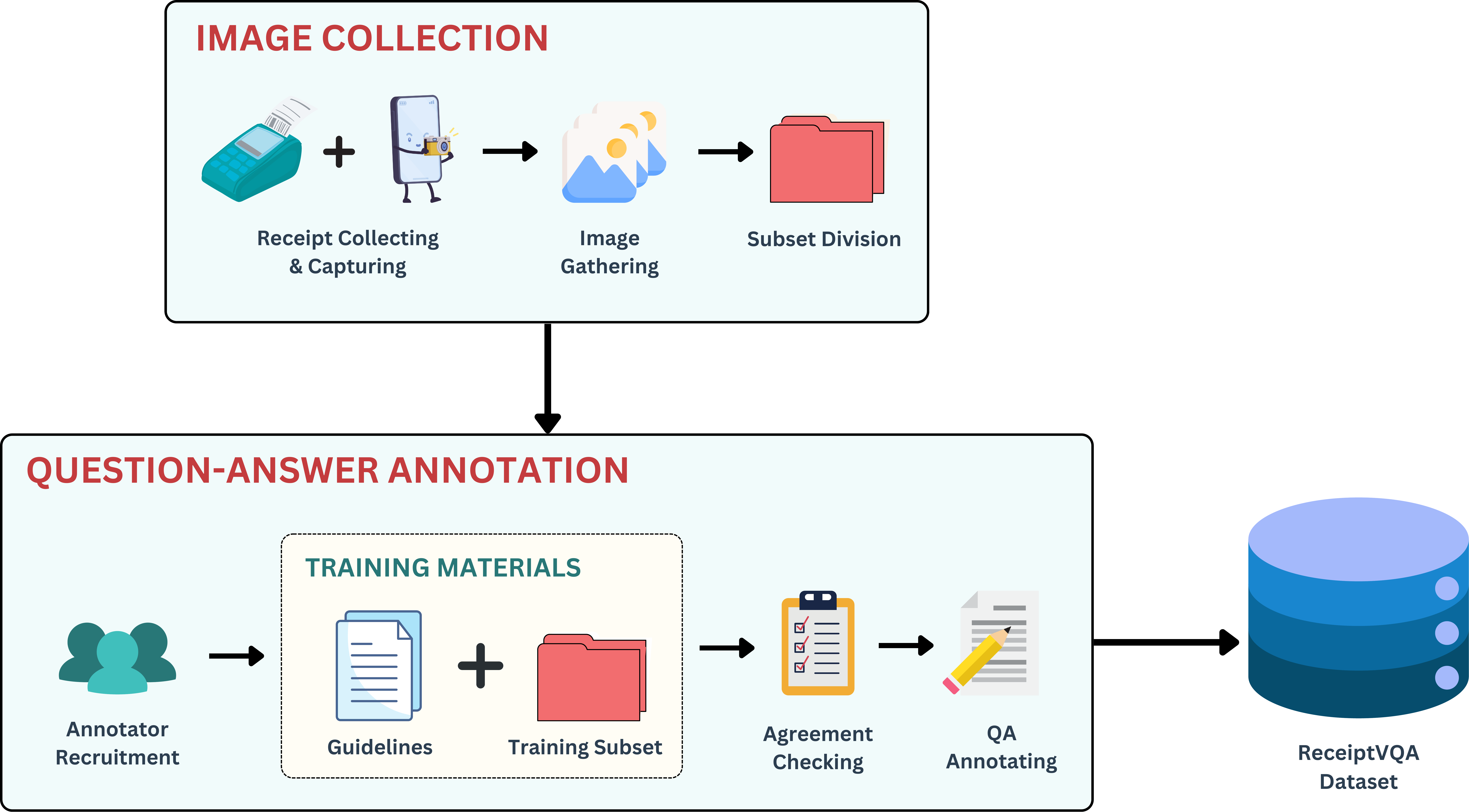}
  \end{minipage}
\end{figure*}

Figure \ref{fig:pipeline} illustrates the overview of our dataset creation. Overall, ReceiptVQA underwent two phases: image collection, and question-answer annotation. The details of each phase are as follows.

\subsubsection{Image Collection} \label{sec:datacre_img}

Receipts used for the dataset were collected from multiple sources, e.g., coffee shops, grocers, supermarkets, and people, and varied from their sizes to brands. Also, faded receipts which most of the texts still can be seen by human eyes were retained to cover the common phenomenon of gradual fading of the thermal paper. We then captured the collected receipts by personal phones, with one receipt for one image. Each receipt was captured once. 

The images were created with various backgrounds and different angles, while ensuring their textual contents were adequately readable. In addition, we utilized images from the MC-OCR dataset \cite{mcocr} by selectively eliminating its images that have texts virtually can not be read and combining the rest with our acquired images. After the collecting process, we obtained 9,500 receipt images. The images were then divided into 100-image subsets for ease of management in the annotation process.


\subsubsection{Question-Answer Annotation} \label{sec:datacre_ann}

There were 41 people, including 39 hired annotators and two of the authors, who participated in the annotation process. Hired annotators are Vietnamese students from the ages of 17 to 23. The annotators were first instructed with a guideline, which covers our task definition, requirements, and restrictions. In particular, annotators were asked to produce at least six unique QA pairs for each image, and only focus on printed elements, i.e., QA pairs about hand-written texts do not belong to our scope. The question is a Vietnamese natural question inquiring about one or more information about texts on the receipt, and the answer is verbatim text span from the receipt which is enough information to satisfy the inquiry. 

Additionally, a question, by preference, should be answered by specific text spans, i.e., one unambiguous answer for one question, so as to enhance the reliability, and applicability of the document task. Moreover, while encouraging participants to use their own lexical resources and grammar to create questions, we restricted them to use canonical forms of words, except leveraging texts on the receipt such as brand names, or product codes as the intermediary for asking more specifically. Furthermore, the questions must belong to neither yes/no question (e.g., \textit{Did the customer buy product A?}) nor selective question (e.g., \textit{Did the customer arrive at 10 am or 10 pm?}) since they have limitations of challenging factors and informative prospects for such a document-related task. Besides, if the annotator asks questions that need to be answered in the listed manner, \textit{<sep>} token must be applied between each two items in the response.

To ensure the dataset quality, 150 images were selected randomly to create a subset used for training annotators. The training subset contains 1,000 QA pairs annotated by the authors. Due to the open-ended attribute of the question, which is difficult to evaluate automatically, we focused on how the annotator answers the questions. After the instruction of the guideline, crowd-workers were asked to provide answers for 1,000 questions from 150 images of the training subset. F1 score, presented in Section \ref{sec:f1metric}, was utilized for the evaluation. We computed the similarity between answers created by hired annotators and pre-annotated answers from the authors on the F1 score, considering that the pre-annotated answers are the ground truth. The average of all crowd-workers' scores gained 80.9\% overall. Subsequently, hired annotators began to annotate QA pairs on 100-image subsets with the payment at \$0.014 per pair. Finally, we gained a total of 64,812 QA pairs of 9,500 receipt images for the ReceiptVQA dataset.


\subsection{Question Classification} \label{sec:Q_classify}

To analyze the language perspective of our dataset, we categorized the questions of our annotations, since questions are projected to give more linguistic insights compared with varied text spans in the answers. Prioritizing the quality of QA content, we decided to classify the questions automatically after the QA annotation process was finished.

According to the study of interrogative construction in the paper \cite{siemund2001interrogative}, most of the questions in our dataset fall into constituent interrogatives due to the confinement of yes/no and selective questions in the annotation process. Constituent interrogative sentences are primarily determined by interrogative words, which are dedicated to provide \textit{meta-information} to elicit desired answers. For instance, interrogative words in English, also called wh-words, include what, which, when, where, who, etc.

In contrast to English, wherein the majority of interrogative words are placed at the beginning of a sentence, Vietnamese interrogative words appear in various positions. For instance, sentence \textit{Khách hàng đã mua \textbf{gì}?} (\textit{\textbf{What} did the customer buy?}) has the interrogative word \textit{\textbf{gì}} (\textit{what}) at the end position, while sentence \textit{\textbf{Ai} là nhân viên phụ trách hóa đơn này?} (\textit{\textbf{Who} is the employee in charge of this receipt?}) places \textit{\textbf{ai}} (\textit{who}) at the initial position. In addition, Vietnamese is a monosyllabic language that a word can consist of multiple syllables separated by spaces, e.g, the word \textit{\textbf{khách hàng}} (\textit{customer}) is constructed by \textit{\textbf{khách}} (\textit{guest}) and \textit{\textbf{hàng}} (\textit{goods/product}), and some interrogative words also have this attribute, e.g., \textit{tại sao} (\textit{why}), \textit{như thế nào}/\textit{thế nào} (\textit{how}), \textit{khi nào}/\textit{lúc nào} (\textit{when}), and \textit{ở đâu} (\textit{where}). Thus, solely relying on the first words in the sentence to classify Vietnamese questions is infeasible. 

Following \cite{siemund2001interrogative,ngo2020vietnamese}, and considering our receipt domain, we categorized our questions into eight classes including Location, Object, Quantity, Time, Reason, Manner, Person/People, and Other, which were designed to represent types of information that expected to be answered. We generalized the classifying process by making use of typical keywords, a part of or a whole interrogative word, which are the minimal interrogative parts that can be categorized. For example, \textit{cái gì} (\textit{what thing(s)}), and \textit{là gì} (\textit{am/is/are what}) can be classified by one keyword \textit{gì} (\textit{what}), and \textit{ở đâu} (\textit{where}) can be recognized by keyword \textit{đâu} (also means \textit{where}). Table \ref{tab:qa_class} illustrates the chosen keywords and their classes respectively. 

\begin{table*}[hbt]
\caption{Question-Answer classes with their respective keywords and glosses }
\setlength{\tabcolsep}{0.5\tabcolsep}
\resizebox{\textwidth}{!}{%
\begin{tabular}{ccc}
\hline
{\textbf{Class}} & \textbf{Keyword} & \textbf{Gloss}\\ 
        \hline
        Location & đâu & where\\
        \hline
        Object & gì, nào & what/which\\
        \hline
        Quantity & mấy, nhiêu & how much/how many/what quantity\\
        \hline
        \multirow{6}{*}{Time} & khi nào, lúc nào, thời gian nào & when \\ 
                            & ngày nào, ngày mấy, ngày bao nhiêu, mùng nào, mùng mấy & what day/what date\\
                            & tháng nào, tháng mấy & what month \\ 
                            & giờ nào, mấy giờ & what time \\ 
                            & năm nào & what year \\ 
                            & thứ mấy & what week day \\
        \hline
        Reason & vì sao, tại sao, để làm gì & why/what purpose\\
        \hline
        Manner & thế nào, bằng cách nào, làm cách nào, làm sao, như nào & how \\
        \hline
        Person/People & ai & who\\
        \hline
        Other & - & - \\
        \hline
    \end{tabular}%
}
\label{tab:qa_class}
\end{table*}

We set rules to check whether a keyword is in the given question, and which category it belongs to. Due to the Vietnamese monosyllabic attribute, some keywords having one syllable when combined with other special syllables can become keywords in other classes. For instance, keyword \textit{nào} (\textit{what/which}) in Object class gotten preceded by \textit{ngày} (\textit{day/date}) becomes keyword \textit{ngày nào} (\textit{what day/what date}) in Time class, while \textit{nào} (\textit{what/which}) combining with syllable \textit{như} (\textit{like/similar to}) turns into keyword \textit{như nào} (\textit{how}) in Manner class. There are such four conflict single-syllable keywords: \textit{gì} and \textit{nào} (\textit{what/which}) in Object and \textit{mấy} and \textit{nhiêu} (\textit{how much/how many}) in Quantity. To avoid these problems, we prioritized checking multi-syllable keywords having conflict syllables, which create more particular contexts. If there was no match, the single-syllable keywords, which cover the remaining contexts, were then checked. The conflict cases must return only one out of the conflict classes.


\subsection{Dataset Statistics and Analysis}

\subsubsection{General Statistics}

ReceiptVQA dataset contains 9,500 receipt images, and 64,812 QA pairs. For evaluation, we randomly split the dataset into three sets train, dev, and test with the ratio 8:1:1 respectively. Since the questions and answers are created from the content of the images, we ensured that there is no overlap image in each set to the others. Table~\ref{tab:train_dev_test} shows details of split images as well as QA pairs in each set.

\begin{table}[hbt]
    \centering
    \caption{Data split counts}
    \begin{tabularx}{\columnwidth}{>{\centering\arraybackslash}X >{\centering\arraybackslash}X >{\centering\arraybackslash}X}
        \hline
         Set name & Num. image & Num. QA pair \\
        \hline
        train & 7,600 & 51,886\\
        \hline
        dev & 950 & 6,426\\
        \hline
        test & 950 & 6,500\\
        \hline
    \end{tabularx}
\label{tab:train_dev_test}
\end{table}

\begin{figure}[hbt]
    \centering
  \includegraphics[width=.9\linewidth]{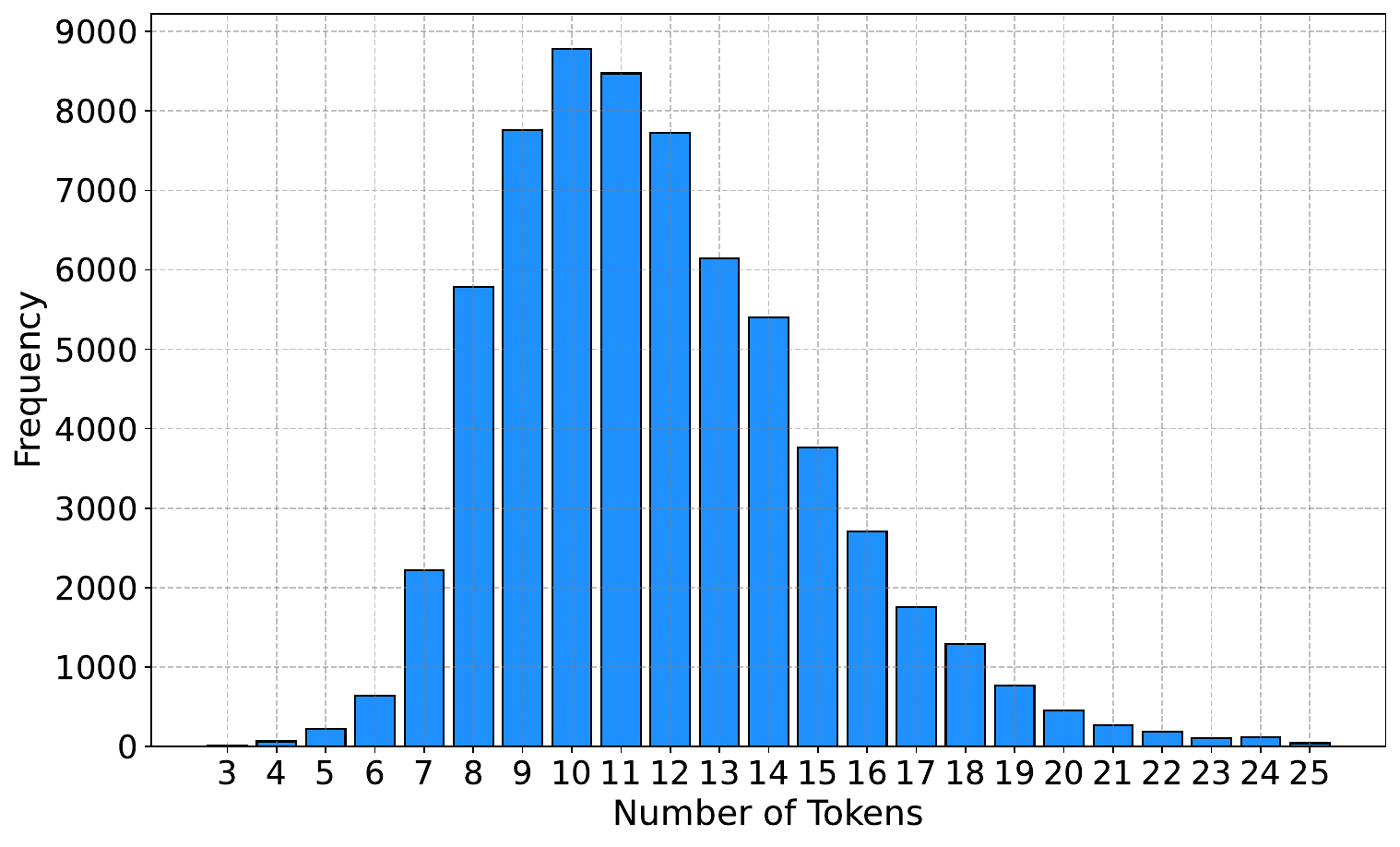}
  \caption{Number of questions with a particular question length}
  \label{fig:Qlengths}
\end{figure}

We conducted a fundamental statistical analysis of ReceiptVQA and two former Vietnamese VQA datasets, OpenViVQA, and ViVQA. Table \ref{tab:stats_comparison} shows that our number of QA is much larger than the others, although our dataset has fewer images. This results in a significantly higher rate of average QA pairs per image, with 6.8 pairs of ReceiptVQA compared to 4.3 pairs of OpenViVQA and 1.5 pairs of ViVQA. This indicates that machine learning systems could have more chances to comprehend our data in the training process. In addition, we calculated the lengths of the questions and answers by splitting them into sequences of tokens by white spaces and counting the number of tokens in each sequence. The average question length of our dataset is higher than the two others. However, our average answer length, moderately higher than ViVQA's, is much lower than that of OpenViVQA, which can be explained by the open-ended characteristic of the OpenViVQA answers. Furthermore, Figure \ref{fig:Qlengths} shows that our questions cover a wide range of lengths with the majority around 8 to 14 tokens, while the lengths of the answers are mainly one token. Overall, our ReceiptVQA presents a VQA dataset of natural questions with answers that reflect the conciseness of an extractive VQA task.

\begin{table*}[hbt]
  \begin{minipage}[c]{0.24\linewidth}
    \caption{Statistics of questions, answers and images in OpenViVQA \cite{openvivqa}, ViVQA \cite{vivqa} and our ReceiptVQA dataset}\label{tab:stats_comparison}
  \end{minipage}
  \hspace{0.02\linewidth}
  \begin{minipage}[c]{0.74\linewidth}
    \centering
    \small
    \resizebox{0.98\textwidth}{!}{%
    \begin{tabular}{cccc}
         \hline
         &  OpenViVQA & ViVQA & ReceiptVQA (ours)\\
          \hline
        Num. image & 11,199 & 10,328 & 9,500\\
         \hline
        Num. question & 37,914 & 15,000 & 68,412\\
         \hline
        Num. unique question & 30,809 & 11,826 & 47,674\\
         \hline
        Avg. QA/image & 4.3 & 1.5 & 6.8\\
         \hline
        Avg. question length & 10.1 & 9.5 & 11.8\\
         \hline
        Avg. answer length & 5.4 & 1.8 & 2.1\\
         \hline
    \end{tabular}%
    }
  \end{minipage}
\end{table*}

\begin{figure}[hbt]
   \centering
  \includegraphics[width=.9\linewidth]{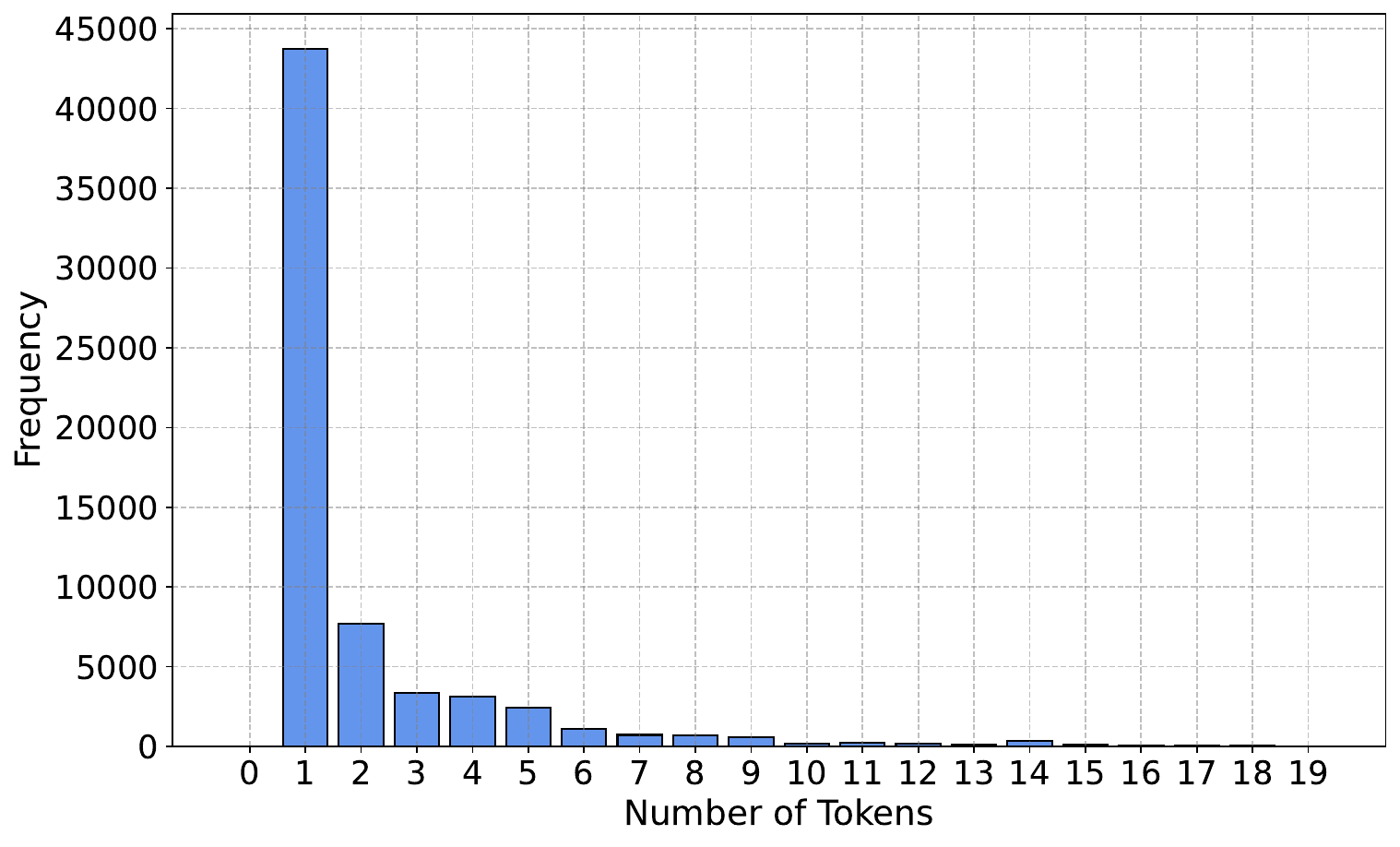}
  \caption{Number of answers with a particular answer length}
  \label{fig:Alengths}
\end{figure}

\subsubsection{Further Analysis}\label{sec:Further_Stat}

\begin{figure}[hbt]
    \centering
    \includegraphics[width=.9\linewidth]{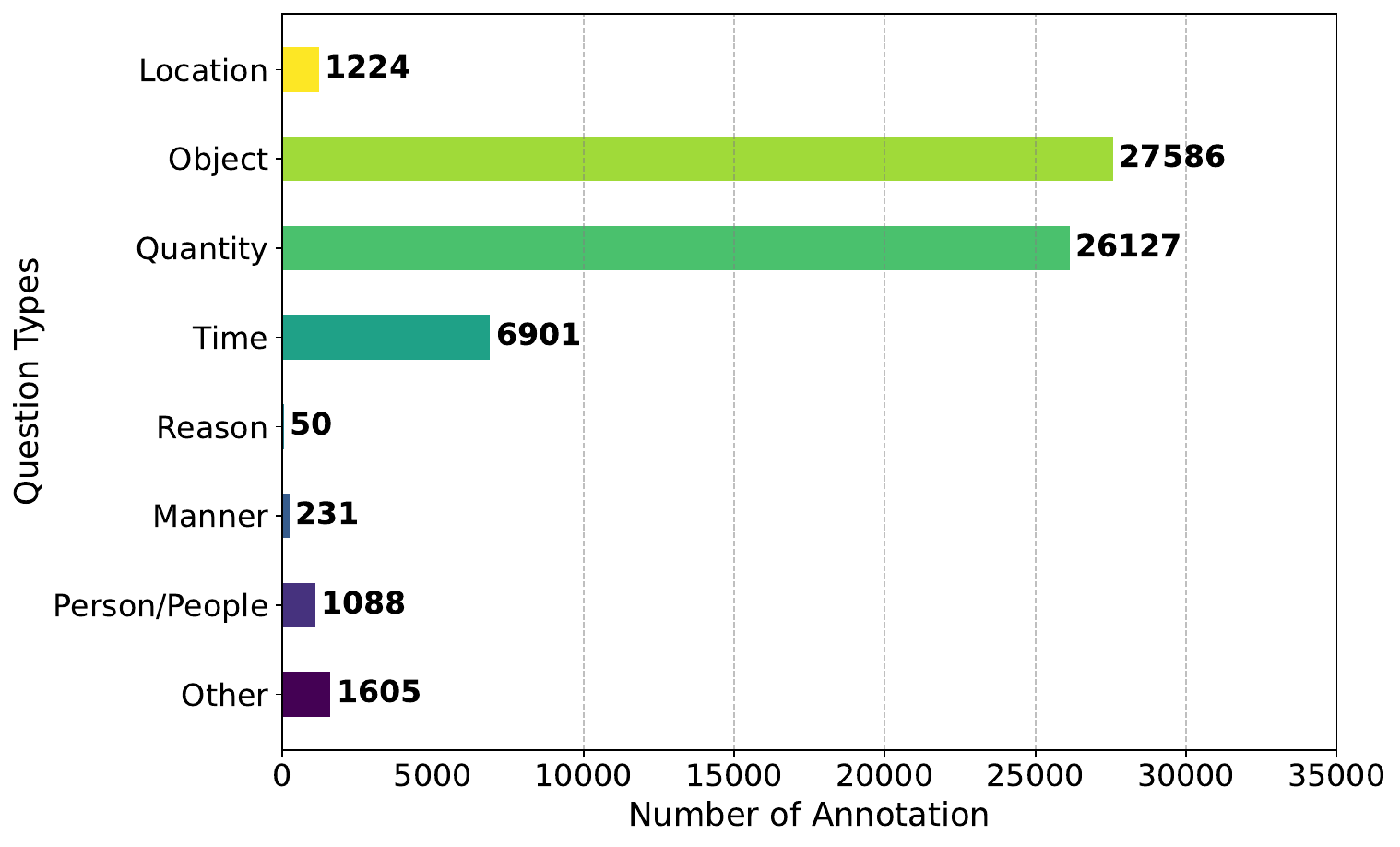}
    \caption{Number of annotations in each question type}
    \label{fig:Questiontypes_Questionnums}
\end{figure}

Regarding the question types mentioned in Section \ref{sec:Q_classify}, Figure \ref{fig:Questiontypes_Questionnums} shows that Object (27,586 annotations) and Quantity (26,127 annotations) are the most populous question types in the dataset, followed by Time with 6,901 annotations. Additionally, no other question type has a number of samples exceeding 2,000 annotations, and the Reason type has only 50 annotations, as expected given the nature of the domain. Besides, the type Other has 1,605 annotations. We observed that these Other questions generally fall into those containing more than one keyword, e.g., question \textit{Sản phẩm đầu tiên trong hóa đơn này có tên là \textbf{gì} và số lượng mua bao \textbf{nhiêu}?} (\textit{\textbf{What} is the name of the first product on this receipt and \textbf{how many} is it purchased?}) with the answer \textit{Longan Tea- Jasmine (L) <sep> 2}, or questions do not have any listed keywords, e.g., \textit{Số tham chiếu của hóa đơn này là?} (\textit{The reference number of this receipt is?}). In addition, the number of questions requiring answers to include \textit{<sep>} token is 1,068; we ensured that each set in Table \ref{tab:train_dev_test} contains a certain amount of this kind of answer.

We analyzed the lengths of questions and answers in each type of question. We separated questions into tokens, which is similar to the way used to assess question lengths in Figure \ref{fig:Qlengths}. Answers, however, were separated into characters to differentiate them more specifically, since Figure \ref{fig:Alengths} shows they mainly contain one token.

\begin{figure}[hbt]
    \centering
    \includegraphics[width=.9\linewidth]{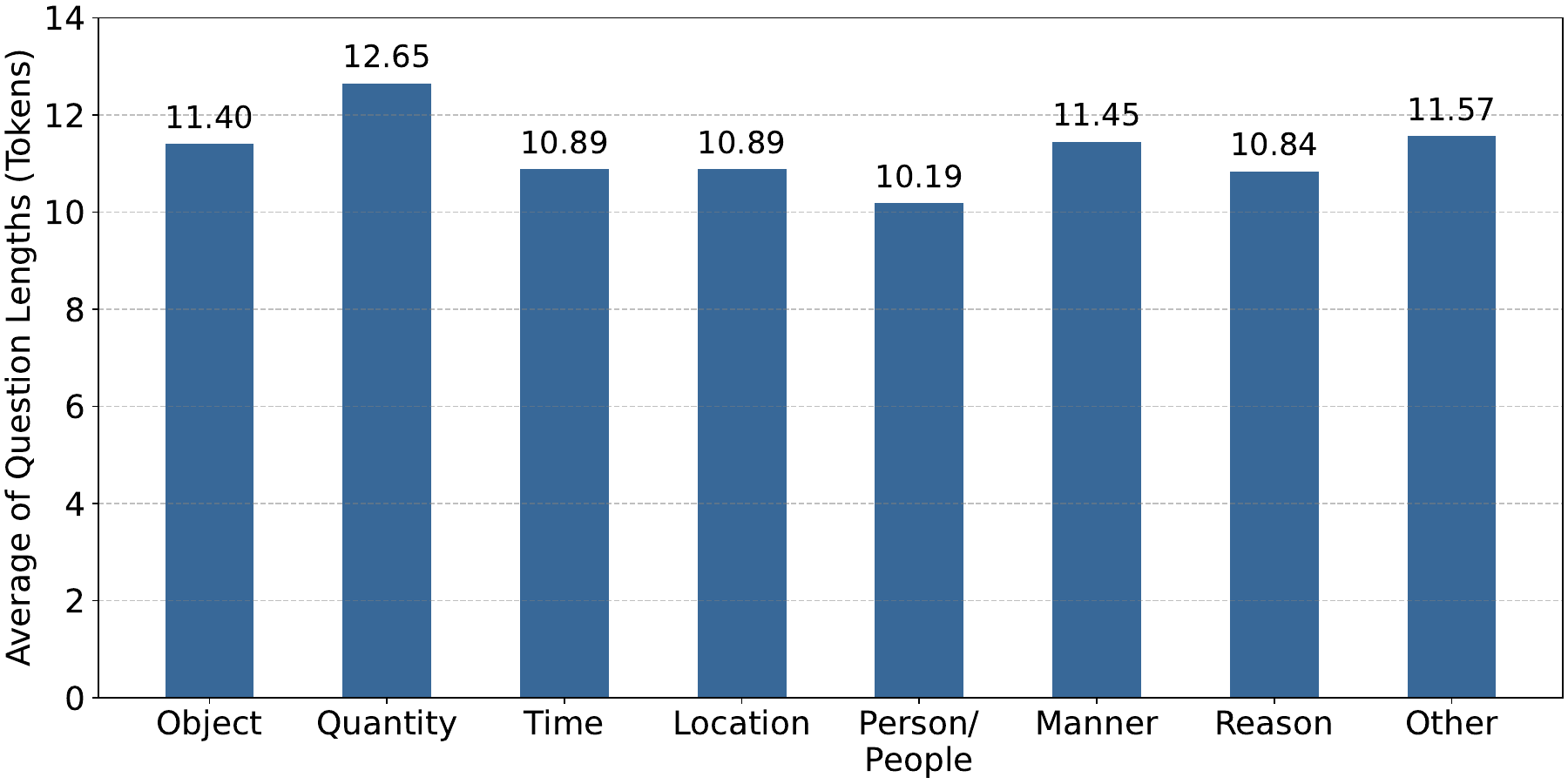}
    \caption{Average of question lengths, counted in tokens, in each question type}
    \label{fig:Qlen_Qtype}
\end{figure}

Figure \ref{fig:Qlen_Qtype} shows that the lengths of questions are roughly evenly distributed among the question types, spanning from 10 tokens to 13 tokens. Questions about people or a person are relatively lower than the others, while asking about quantity is used relatively longer questions. Contrary to the questions, the length distribution of the answers has certain differences among the types. In particular, answers for Location questions have the longest average length with nearly 40 characters, while the lowest average length belongs to answers for Quantity questions. This could be explained that Quantity questions tend to require numbers, or codes, which are often a couple of characters, whereas answering about locations is likely to include both numbers and letters to specify street numbers, street names, provinces, etc.

\begin{figure}[hbt]
    \centering
    \includegraphics[width=.9\linewidth]{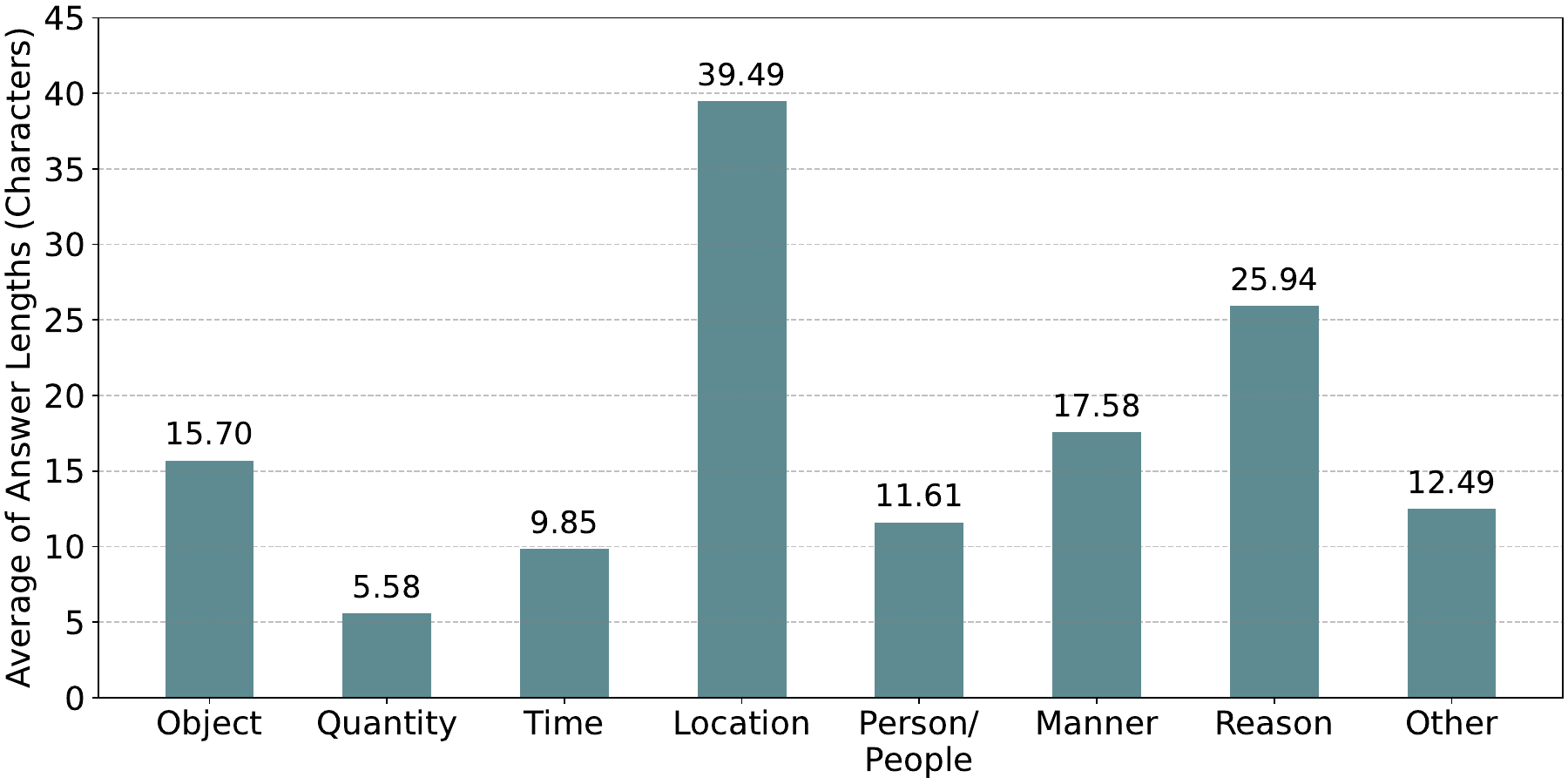}
    \caption{Average of answer lengths, counted in characters, in each question type}
    \label{fig:Alen_Qtype}
\end{figure}

\begin{figure}[hbt]
    \centering
    \includegraphics[width=.9\linewidth]{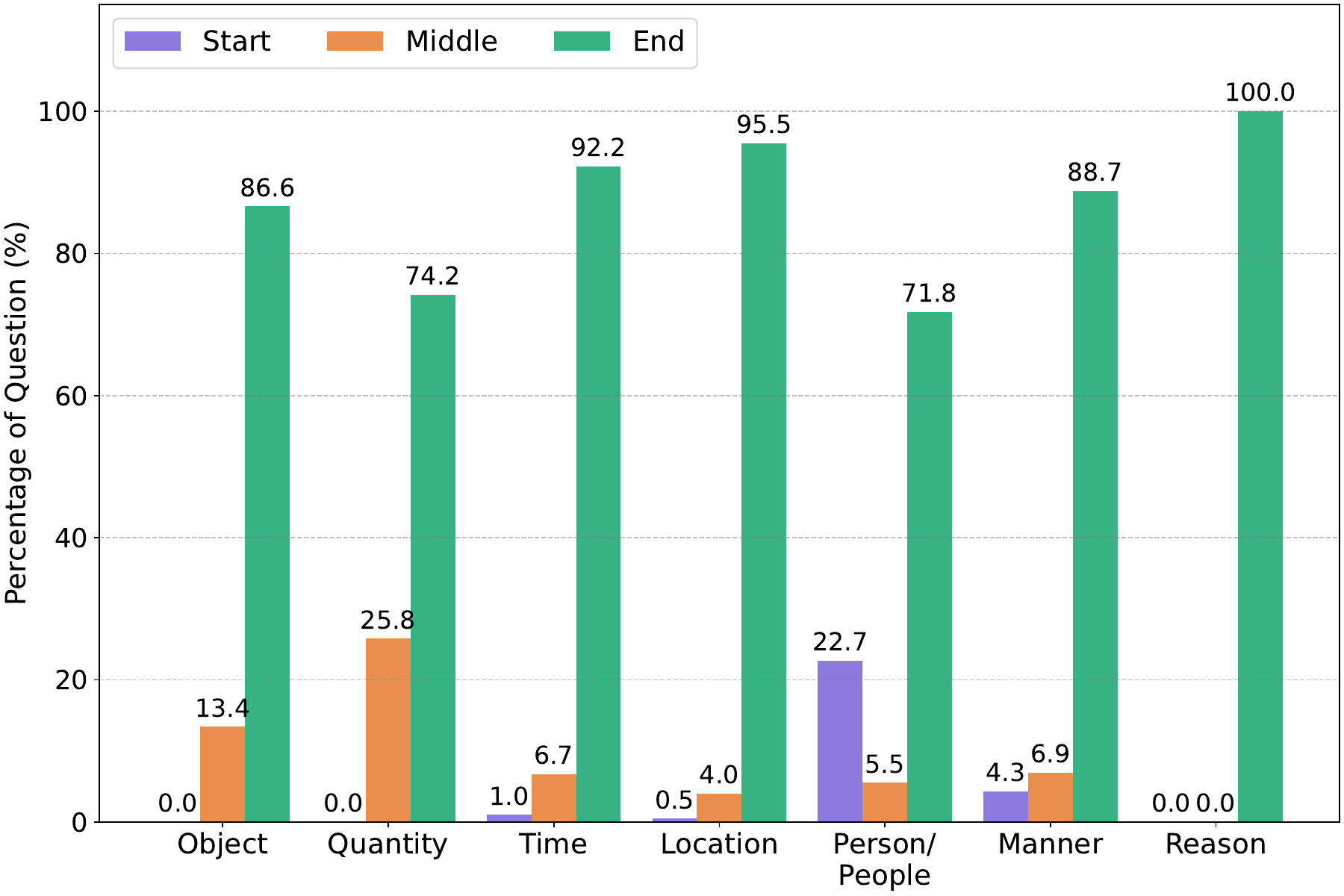}
    \caption{Distribution of keyword position of each question type ordered by their frequency from the highest to the lowest (excluded Other type)}
    \label{fig:Questiontypesdist}
\end{figure}

As mentioned in Section \ref{sec:Q_classify}, the position of Vietnamese interrogative words can vary from the start to the end of a question. To further explore this phenomenon in receipt questions more particularly, we categorized the positions into three labels, wherein the Start and End label denoted whether the keyword is at the beginning or at the end of the question, and the Middle label denoted the remaining cases. Figure \ref{fig:Questiontypesdist} illustrates the proportion of each keyword position in each question type, except the Other type. It can be seen that the majority of our questions has keywords at the end of the sentence; all types have more than 70\% share of the End label, which may indicate that one tends to mention the context before finishing with the interrogative part when it comes to inquiring about receipts. 

\begin{figure}[hbt]
    \centering
    \includegraphics[width=.9\linewidth]{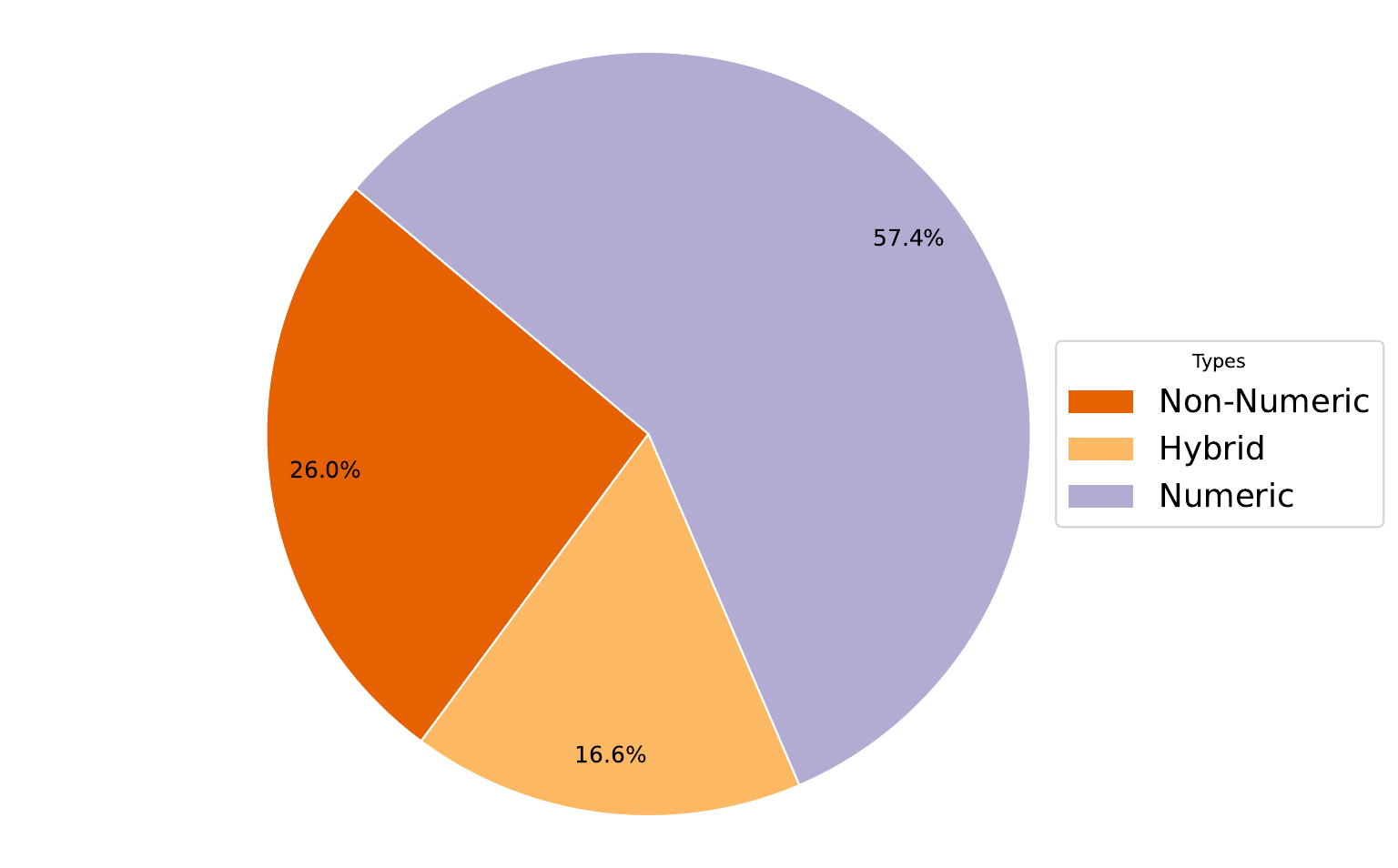}
    \caption{Percentage of each answer type}
    \label{fig:ATypePie}
\end{figure}

In terms of responding to the dataset questions, the answer must be verbatim text spans from the receipt image, where typical information is number, such as product price, total price, and product quantity. To get a clearer view of how one uses spans to answer, we divided our dataset answers into three types: Numeric, Non-Numeric, and Hybrid. Regardless of the punctuation, Numeric answers are answers that contain only numbers, while Non-Numeric answers involve elements that are not numbers, and Hybrid answers encompass the combination of both types. For instance, answer \textit{50 000} or \textit{50.000} or \textit{30/04/2024} is Numeric, while answer \textit{tiền mặt} (\textit{cash}) or \textit{Trà đào (L)} (\textit{Peach tea (L)}) is Non-Numeric, and answer \textit{50.000 VND} or \textit{24/12/2022(Thứ bảy)} (\textit{24/12/2022(Saturday)}) is Hybrid. Figure \ref{fig:ATypePie} shows that the Numeric answer is the most populous type, accounting for more than half of the total answers, while the proportion of Hybrid answers is the smallest, making up 16.6\%. This suggests that Numeric answers are either easier for crowd-workers to annotate or actually the most vital form of answer for receipts.

\begin{figure}[hbt]
    \centering
    \includegraphics[width=.9\linewidth]{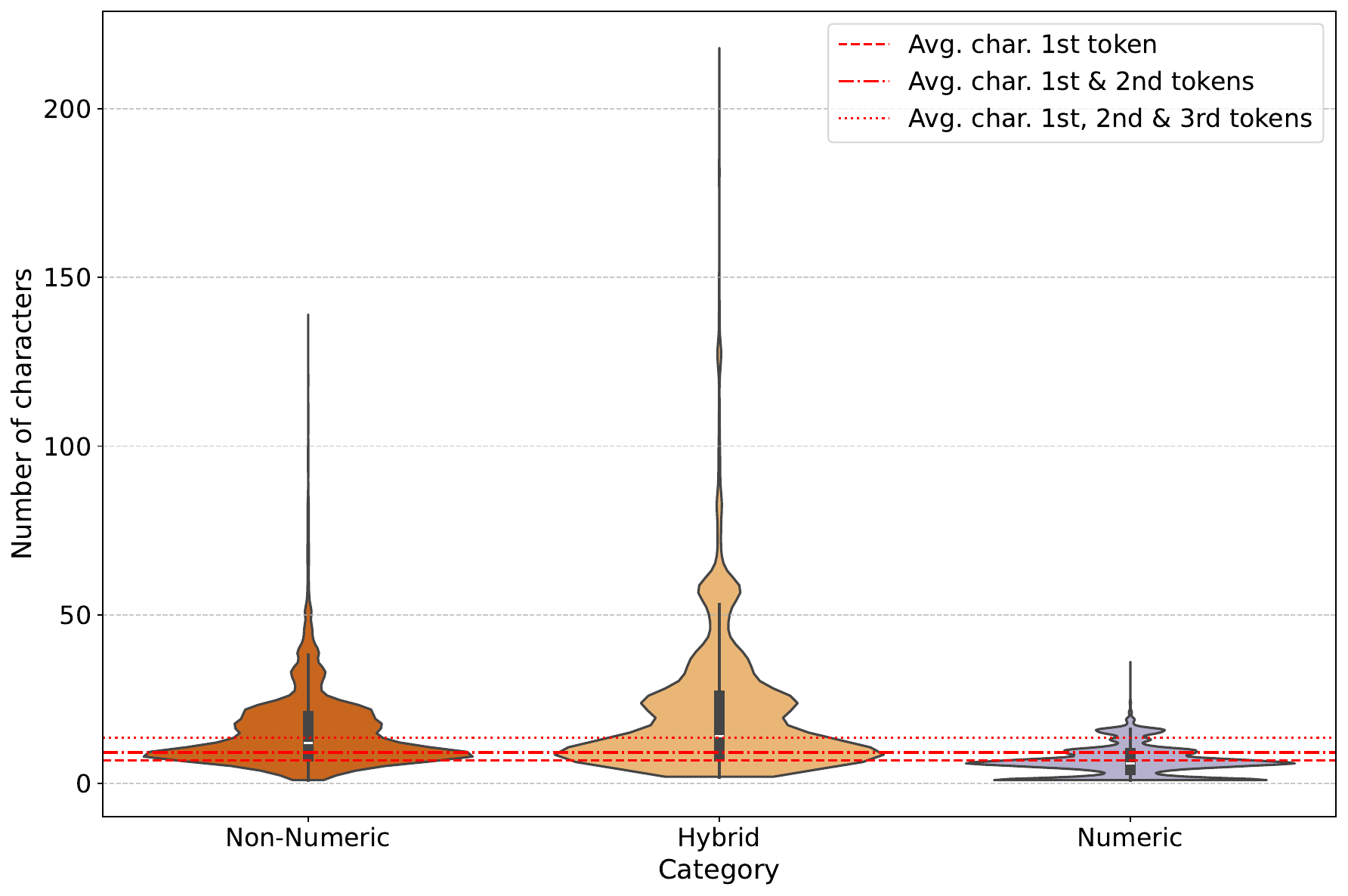}
    \caption{Distribution of answer types against numbers of characters}
    \label{fig:ATypeVio}
\end{figure}

Furthermore, we explored the distribution of each answer type on the character-level lengths, shown in Figure \ref{fig:ATypeVio}. The red lines represent the average number of characters in one first token (6.8 characters), two first tokens (9.2 characters), and three first tokens (13.5 characters) in the answers. It is apparent that Numeric answers are generally shorter than the others, and mainly used by from one to three tokens. Whereas, Hybrid answers have the largest range of lengths, having a visible density ranging from zero to approximately 55 characters. The Non-Numeric answers also have a relatively large range, reigning the middle position between the other two types. This indicates that annotators may have to put more effort into Hybrid and Non-Numeric answer types than the Numeric type, intending to invest more time in Numeric answers to optimize the amount of money they can receive. Therefore, identifying and grouping difficulties of the annotations in advance for balancing annotation credit could result in more diverse datasets in future works. Additionally, we explored the Part-of-Speech (POS) tagging perspective on our dataset in Appendix \ref{sec:secA1}.


\section{Our Proposed Method} \label{sec:ProposedMethod}

Document images are characterized by an immense density of texts. Thus, using regular 2D locations for such document tasks usually means not only handling 2D information of vertical and horizontal locations of both top-left and bottom-right points of all bounding boxes but also reasoning semantic information of both the given question and OCR texts comprehensively. On the other hand, document VQA tasks often use a relatively small portion of the total text spans to provide answers, which may not be necessary to process precisely all the locations of OCR bounding boxes. Therefore, in our layout approach, instead of directly using the locations of OCR bounding boxes, our model first hashes them into simple symbols to represent the areas they belong to and then uses them to operate layout embeddings via embedding layers of the pretrained model. OCR texts in the same area at the examined hashing level have the same representation. By hashing through multiple levels, OCR elements are initiated relative 2D connections to each other, constituting the 2D layout modality. 

\begin{figure*}[hbt]
  \begin{minipage}[c]{0.12\linewidth}
    \caption{Overview of LiGT model architecture}
    \label{fig:ModelArch}
  \end{minipage}
  \hspace{0.005\linewidth}
  \begin{minipage}[c]{0.85\linewidth}
    \centering
    \includegraphics[width=.95\linewidth]{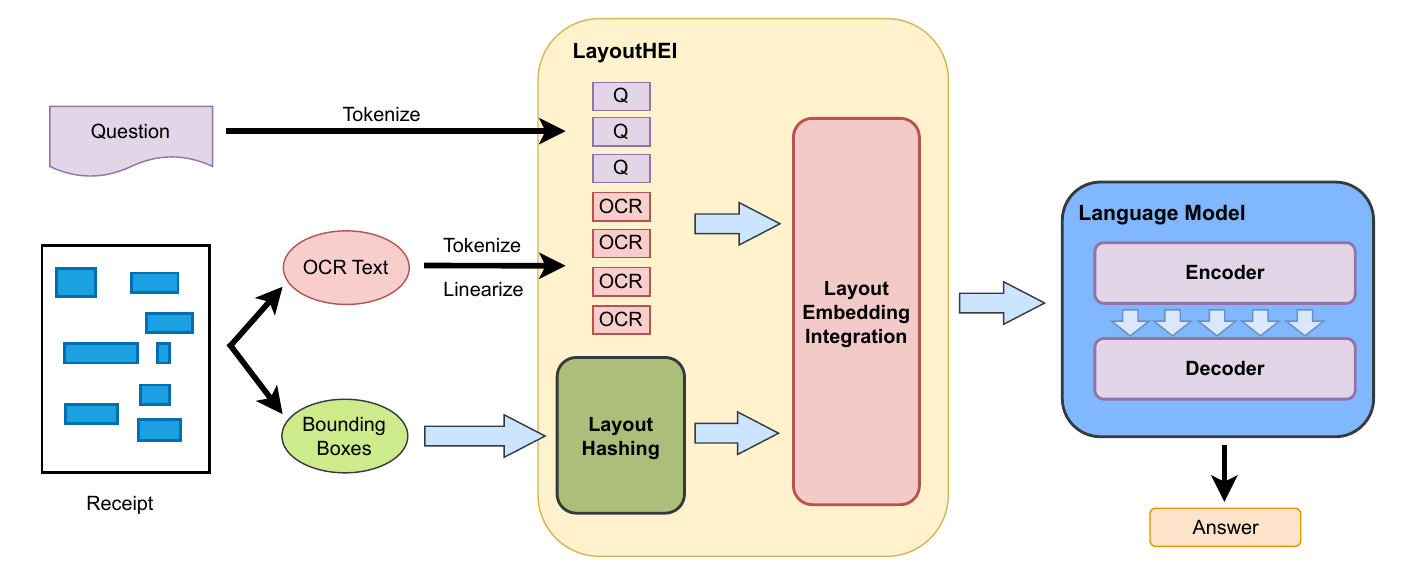}
  \end{minipage}
\end{figure*}

Figure \ref{fig:ModelArch} schematically shows the overview of LiGT architecture. Our proposed method has two main components: LayoutHEI (\textbf{Layout} \textbf{H}ashing and \textbf{E}mbedding \textbf{I}ntegration), and Language Model, which are detailed in the following subsections.

\subsection{LayoutHEI} \label{sec:LayoutHEI}

We conceptualized our LayoutHEI component based on two principles: (i) gradually specifying the 2D location of text via their relative areas throughout multiple levels, and (ii) leveraging inherent properties of language models for 2D representing. LayoutHEI focuses on extending the semantic embeddings of pretrained models, complementing 2D layout information to the language embedding input. This proposed component consists of two phases, \textit{Layout Hashing} and \textit{Layout Embedding Integration}.

\subsubsection{Layout Hashing}

\begin{figure}[b]
    \centering
    \includegraphics[width=1\linewidth]{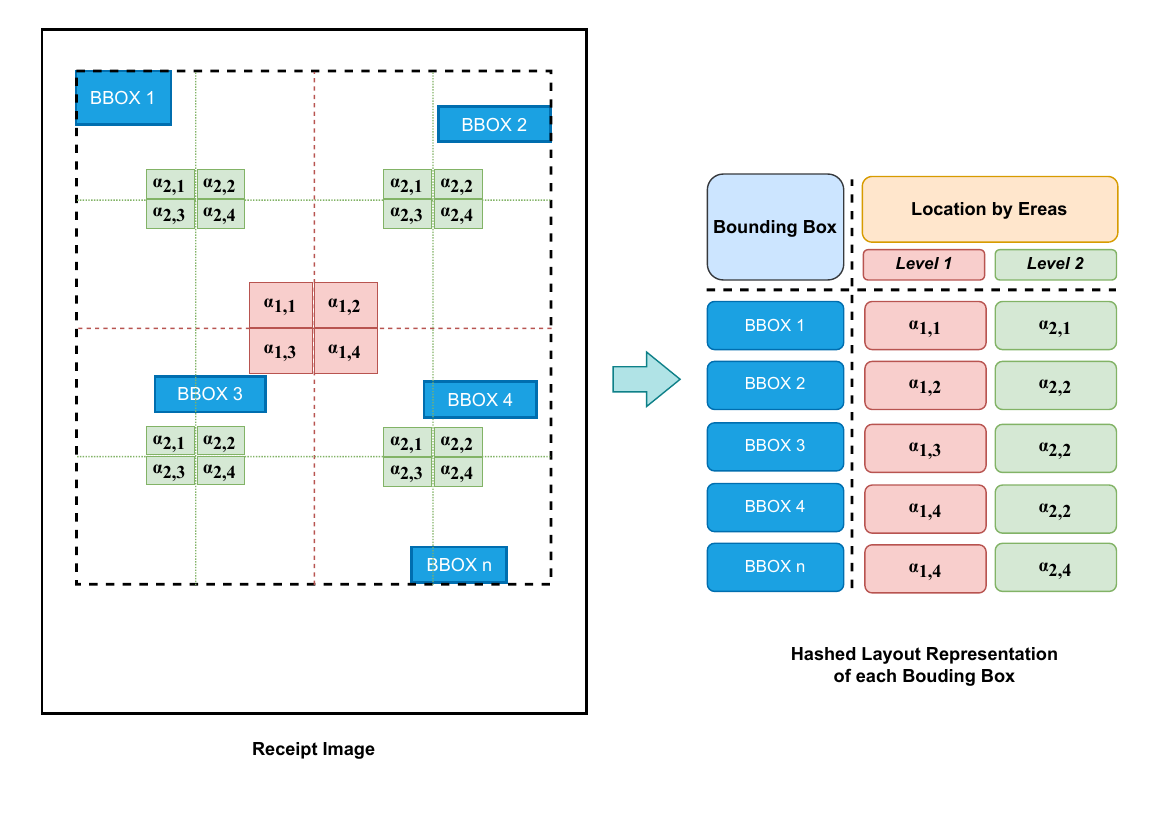}
    \caption{Illustration of the Layout Hashing phase in our LayoutHEI component in the LiGT architecture. For simplicity, only two hashing levels are presented}
    \label{fig:layouthashing}
\end{figure}

Figure \ref{fig:layouthashing} presents the overview of the layout hashing phase, wherein given a set of \textit{n} OCR bounding boxes and a number of hashing levels \textit{L}, the system provides \textit{L} sequences of \textit{n} symbols ($\alpha$$_{i,j}$) representing the areas the bounding boxes belong to. The order of symbols in each hashing level follows the reading order. Each symbol $\alpha$$_{i,j}$, where $\textit{i} = \{1,..., L\}$, and $\textit{j} = \{1, ..., 4\}$, gives information that the current bounding box location is assessed in the i$^{th}$ of \textit{L} level, at the j$^{th}$ of 4 divided areas (quarters). 

At the first hashing level, the examined area is initiated by identifying a rectangle having max and min values of the x and y axis of all the bounding boxes, i.e, the rectangle has (x$_{min}$, y$_{min}$, x$_{max}$, y$_{max}$) location in which the x, y values are from all bounding boxes' locations. As a result, the hashing scope only depends on the locations of the given bounding boxes.

After the examined area is initiated, it is then separated into four equal parts (quarters). These quarters are used to identify the relative positions of the bounding boxes. We denote each quarter with a corresponding symbol, $\alpha$$_{i,j}$. 

Bounding boxes are assigned with the symbol corresponding to the quarters they belong to. We identify the area containing each bounding box by identifying which quarter the location of the bounding box's center point is in.

Subsequently, we gain a sequence of \textit{n} symbols representing the quarters containing the bounding boxes. The process is recursively executed \textit{L} times for \textit{L} hashing levels. From the second hashing level, the examined areas are pivoted to the quarters of the previous hashing level. 

At the end of this phase, we gain \textit{L} sequences of n symbols, $\alpha$$_{i,j}$, hierarchically representing the relative location of each bounding box. The 2D layout information of each bounding box is conveyed by a combination of its L symbols. In other words, each bounding box position is identified by \textit{L} symbols, wherein the higher the level of the symbol, the more specific the bounding box position is.

\subsubsection{Layout Embedding Integration}

\begin{figure}[b]
    \centering
    \includegraphics[width=1\linewidth]{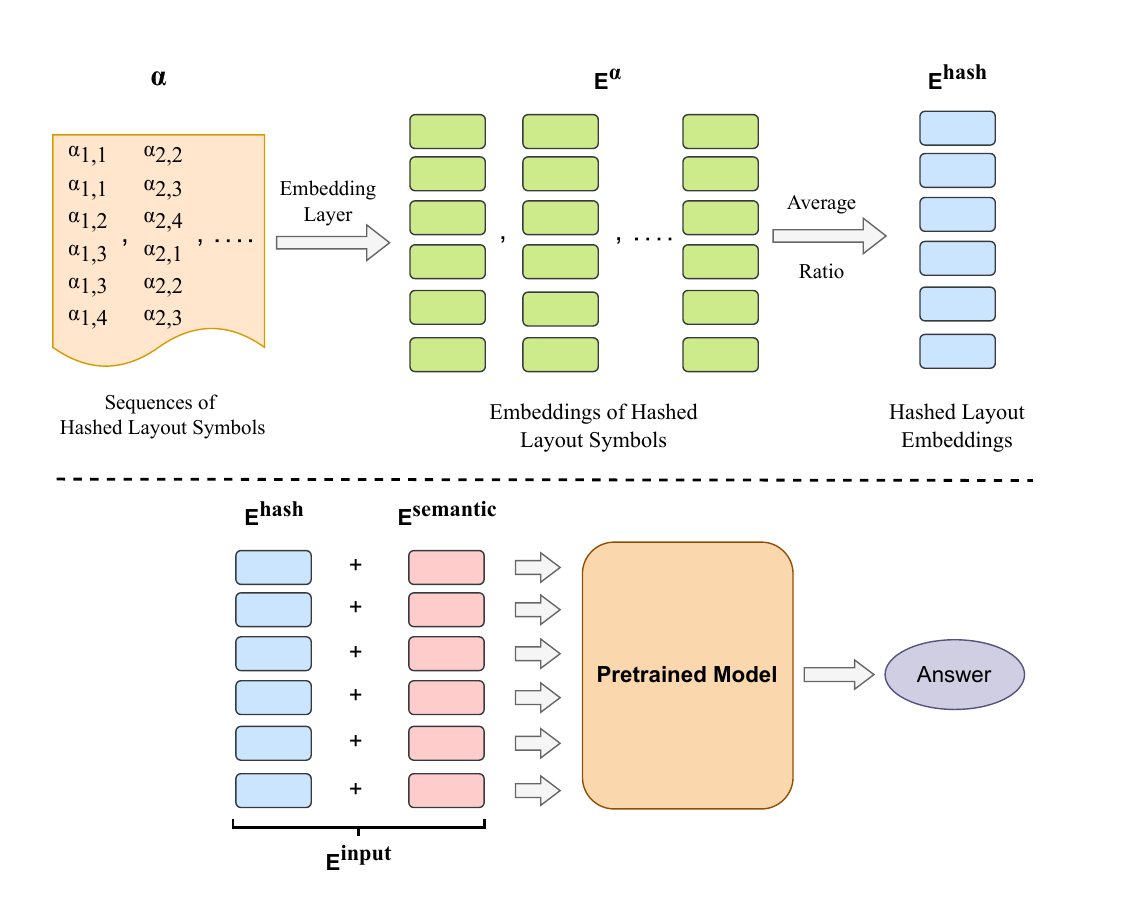}
    \caption{Illustration of the Layout Integration phase in our LayoutHEI component}
    \label{fig:embeddingintegration}
\end{figure}

Each hashing level has four symbols to denote the corresponding quarters. In practice, we use upper-cased letters in the English alphabet to simply present the symbols. Following the alphabetical order, four continual letters are used for four symbols in each hashing level. Each letter is assigned to a quarter following the reading order. In other words, we associate $\alpha$$_{1,1}$ with letter \textit{A}, $\alpha$$_{1,2}$ with letter \textit{B}, $\alpha$$_{2,1}$ with letter \textit{E}, $\alpha$$_{2,4}$ with letter \textit{H}, etc.. In addition, tokens from questions are considered as special cases of 2D hashing, and we assign them with the letter zero (\textit{0}). 

Figure \ref{fig:embeddingintegration} shows the overview of the layout integration phase. To leverage the properties of pretrained models, we utilize embeddings of the applied pretrained model. In particular, embeddings corresponding to the letters denoting 2D hashing areas are used to create 2D layout embeddings. \textit{L} sequences of hashed layout letters ($\alpha$) are projected through the embedding layer of the language model to gain \textit{L} embeddings ($E^{\alpha}$). The embeddings are then element-wise computed to obtain the average embedding and multiplied with a learnable ratio which ranges from 0 to 1. The computed result ($E^{hash}$) is added to the semantic embedding to become the input embedding, which is projected to the model for training and inferring.

Particularly, given a sequence of \textit{T} input tokens, including tokenized questions and tokenized OCR texts, the t$^{th}$ embedding of the corresponding token ($E_{t}^{semantic}$) is integrated with the layout embedding as follows:

\begin{equation}
    \omega = \sigma(\rho)
\end{equation}

\begin{equation}
    E_{t}^{hash} = \omega \odot \frac{1}{L} \sum_{i=1}^L E_{t}^{\alpha_{i}}
\end{equation}

\begin{equation}
    E_{t}^{input} = E_{t}^{semantic} + E_{t}^{hash}
\end{equation}

\noindent where $\rho$ is a learnable parameter, $\sigma$ is the sigmoid function, and $\odot$ is element-wise multiplication. The average of hashed embeddings corresponding to the t$^{th}$ token is computed and multiplied with the ratio $\omega$ to create the 2D layout embedding $E_{t}^{hash}$. The semantic embedding $E_{t}^{semantic}$ is then added with the layout embedding to create the t$^{th}$ input embedding, being ready to project through the pretrained model.

Besides, regarding our choices of the symbols, alphabetical letters are likely to have the \textit{in-ordered} semantic in pretrained language models, making them suitable to represent arranged and hierarchical elements. Additionally, these letters are basic characters that return in one token after the tokenization process. Thereby, each hashed layout symbol results in one vector of embedding, easily aligning with semantic embeddings of the corresponding tokens.

\subsection{Language Model}

We choose the Text-to-Text Transfer Transformer (T5 \cite{t5model}) as the language model backbone for our architecture. T5 is a widely used model to tackle VQA tasks, easily adapting to process additional modalities. Based on the T5 architecture, many multimodal extensions have proposed and achieved state-of-the-art performance not only in document VQA \cite{tanaka2021visualmrc,tilt,SlideVQA2023,udop} but also in scene-text VQA \cite{latr,SaL,prestu}, and many other multimodal studies. Empirical studies show that T5 architecture has a certain level of resistance against OCR errors with its capability to generate texts by using the model's \textit{built-in} vocabulary. Applying the T5 model to our Language Model component is projected to enhance 2D spatial processing ability and equip our LiGT model the flexibility in producing answers. To effectively handle the Vietnamese language, we made use of the Vietnamese version of T5, named ViT5 \cite{vit5}, pretrained on a large resource of high-quality and diverse Vietnamese texts.


\section{Evaluation} \label{sec:Evaluation}

This section presents the performance of our LiGT model and comparative baselines on the ReceiptVQA dataset. We follow the regular training paradigm that improves models' scores on the training and development set before comparing the overall results on the test set. 

Due to the lack of pretrained layout-aware models designed for Vietnamese, we made use of multilingual models and architectures that can adapt to the Vietnamese language. To extract OCR, we used Google Cloud Vision API\footnote{\url{https://cloud.google.com/vision}}, which results in an average of 142 bounding boxes per image. The following subsections describe chosen methods, metrics, the setup for baseline experiments, experimental results, and further analysis in methodologies.

\subsection{Comparative Baselines}

\subsubsection{Extractive baselines} \label{sec:Ex_baseline}

\begin{figure}[hbt]
    \centering
    \includegraphics[width=1\linewidth]{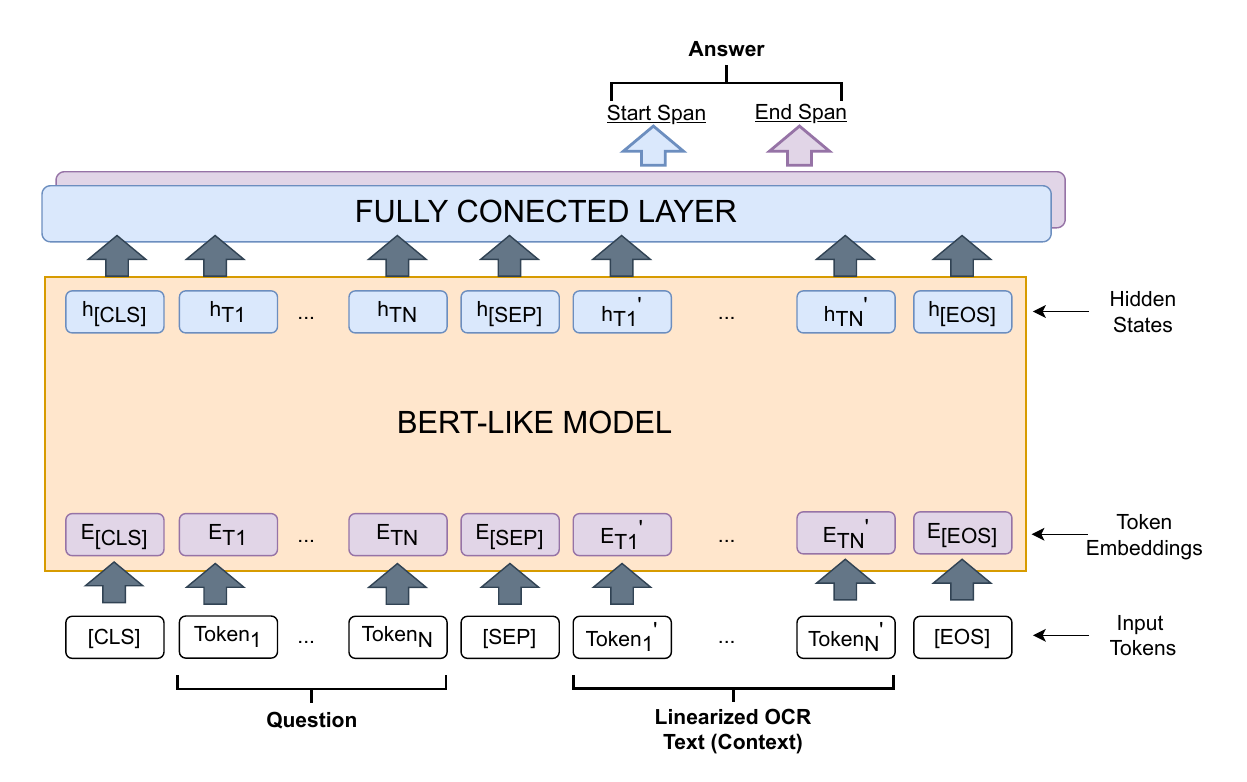}
    \caption{Overview of our Text modality employment for the extractive baselines}
    \label{fig:bertlikeqa}
\end{figure}

Figure \ref{fig:bertlikeqa} schematically shows the overview of our employment for the extractive baselines. All textual inputs and outputs of extractive models follow the same process as illustrated, while visual and layout inputs are particularly extended depending on the model architectures. The outputs are the start index and the end index of the answer span, which are predicted by using two fully connected layers to process the encoded input. 

Preparing answers for extractive models was conducted in an automated manner. We identify the start and end index of the answers by finding the first occurrence of the answers in the context string. Since there often exists errors in OCR text recognition, we also include cases having only output texts that lose one character, if there are two or more characters in their actual outputs. Other cases, such as one character being changed or added, or losing more than one character, are not allowed, as they have a higher possibility of causing misinformation. Table \ref{tab:extractive_procedures} shows percentages of the answers that are able to support the extractive approach. In terms of training, we also include the answers that do not support the extractive approach by assigning them to the index of the \textit{[CLS]} token (the zero index in most cases) and consider the output that only contains \textit{[CLS]} token as the model can not find any appropriate span to answer. 

\begin{table}[hbt]
    \centering
    \caption{Proportions of answers that can produce literal extractive outputs regarding two matching procedures}
    \begin{tabularx}{\columnwidth}{>{\centering\arraybackslash}X >{\centering\arraybackslash}X >{\centering\arraybackslash}X}
        \hline
        \multirow{2}{*}{Set name} & Fully-matched & Our approach\\
        & approach (\%) & (\%) \\
        \hline
        train  & 75.86 & 80.93\\
        \hline
        dev    & 76.17 & 81.26\\
        \hline
        test   & 76.77 & 81.66\\
        \hline
        total  & 75.98 & 81.04\\
        \hline
    \end{tabularx}
    \label{tab:extractive_procedures}
\end{table}

Our used extractive models are listed as follows.
    \begin{itemize}
      \item \textit{Text-only pretrained language models}: we employed two multilingual models, XLM-Roberta \cite{xlmroberta}, and InfoXLM \cite{infoxlm}, and two Vietnamese models, PhoBERT \cite{phobert} and CafeBERT \cite{cafebert}. These are widely used language models that show outstanding performance on various NLP benchmarks in Vietnamese.
      
      \item \textit{LiLT} \cite{lilt}: LiLT is a layout pretrained model that can easily incorporate with an available RoBERTa-based language model, regardless of its pretrained languages, and perform competitively on document understanding tasks. We incorporate LiLT with XLM-Roberta, InfoXLM, and PhoBERT to create three LiLT-based baselines.
      
      \item \textit{LayoutXLM} \cite{layoutxlm}: LayoutXLM is the first multilingual layout-aware pretrained models. Based on LayoutLMv2 \cite{layoutxlm}, it provides a strong multilingual baseline that combines textual, visual, and layout information for document understanding tasks in 53 languages.
    \end{itemize}

\subsubsection{Generative baselines}

Since there is neither a Vietnamese layout-aware generative pretrained model nor a multilingual one, our chosen baselines for this approach are deep neural network architectures that have a backbone of pretrained generative language model, which can be replaced by a corresponding Vietnamese version. In other words, we did not apply any pre-training phase and each generative baseline only underwent one phase of training on our ReceiptVQA dataset.

The chosen architectures are as follows.
    \begin{itemize}
          \item \textit{ViT5} \cite{vit5}: ViT5 is a Vietnamese version of T5 \cite{t5model}, which is a pretrained Transformer-based encoder-decoder model. ViT5 has shown remarkable performance not only in text generation tasks such as text summarization but also in text classification tasks in Vietnamese.
          
          \item \textit{ViT5+2D, ViT5+U, ViT5+2D+U} \cite{tilt}: are originated from the research of the TILT model \cite{tilt} and its implementation in the DUE benchmark \cite{due}. These architectures extend the T5 model \cite{t5model} to understand layout information and visual content. The layout is handled by adding 2D layout embeddings to the model attention bias (denoted by \textit{+2D}), while the model's visual ability is from the utilization of U-Net \cite{unet} (denoted by \textit{+U}). The U-Net is trained together with the T5 backbone in the training phase as a whole architecture. Despite being introduced in 2021, these methods still stand as standard baselines for generative approaches in document understanding, and have been widely used till recent research \cite{dude,udop,docformerv2}. With regard to our experiments, we apply ViT5 \cite{vit5} as the T5 backbone for these architectures.
          
          \item \textit{LaTr} \cite{latr}: is a method initially designed for scene-text VQA tasks. The model processes input from a concatenation of text embedding, 2D layout embedding, and the spatial embedding extracted from Vision Transformer encoder (ViT) \cite{imageworth16x16words}, and feeds them to its T5 \cite{t5model} backbone. While most scene-text VQA architectures involve an object detection module, such as Faster R-CNN \cite{fasterrcnn}, which provides specific visual information of objects in the image, LaTr utilizes patched features encoded from ViT \cite{imageworth16x16words}, which may benefit in document tasks primarily focusing on text in the image. The employment of our LaTr version also made use of ViT5 \cite{vit5} as the T5 backbone.
        \end{itemize}
        
\subsubsection{Upper bounds and heuristics}

We employed fundamental upper bound and heuristic baselines to facilitate the comparison of our chosen models on the dataset. For our upper bound, we leveraged results from fully-matched answers in Table \ref{tab:extractive_procedures}, and named it \textit{MATCHED\_OCR}. We also made use of fully-matched answers in cases where the OCR contexts are limited to 142 bounding boxes, the average number of OCR bounding boxes, \textit{MATCHED\_OCR\_AVG}. The upper bound represents the ability to provide answers precisely if they exist in the given OCR text input, otherwise return empty strings. 

Regarding the heuristic baselines, we provided two baselines that randomly select an answer from the top 10, and top 100 most frequent answers in the train set, named \textit{RAND\_TOP10}, \textit{RAND\_TOP100} respectively. These heuristics reflect whether our dataset is biased and could be handled by merely selecting some random answers.

\subsection{Evaluation Metrics}

With regard to the receipt domain where the majority of answers contain numbers, we value the preciseness of each single character in the results. Thus, we made use of the F1 metric and the Accuracy metric. Moreover, following previous research in document VQA, we also evaluated the Average Normalized Levenshtein Similarity (ANLS) \cite{biten2019scene}.

\subsubsection{F1} \label{sec:f1metric}

Our applied F1 version is similar to the F1 in EVJVQA Challenge of VLSP 2022 \cite{Luu_Thuy_Nguyen_2023}. Answers from the prediction set and ground-truth set are first separated by white spaces to gain sequences of tokens. The final scores of the metric are the average score of each pair of predictions and corresponding ground truth sequences. Each pair is computed following the formula:

\begin{equation}
    \textit{Precision} = \frac{\textit{number of shared tokens}}{\textit{number of tokens in the evaluated set}}
\end{equation}

\begin{equation}
    \textit{Recall} = \frac{\textit{number of shared tokens}}{\textit{number of tokens in the ground truth set}}
\end{equation}

\begin{equation}
    \textit{F1} = \frac{2 \times \textit{Precision} \times \textit{Recall}}{\textit{Precision} + \textit{Recall}}
\end{equation}

\subsubsection{Accuracy}

Accuracy is computed by determining whether the predicted answer is mapped correctly to its ground truth. The final Accuracy score is the percentage of the shared answers in the ground truth that exactly match their inferred answers.

\subsubsection{ANLS}

Proposed along with STVQA dataset \cite{biten2019scene}, ANLS utilizes normalized Levenshtein similarity \cite{levenshtein1966binary} to reduce the strictness of the Accuracy metric, which will return zero even if the prediction has minor flaws. As our dataset has one ground truth answer for one question, we simplify the applied ANLS formula as follows:

\begin{equation}
    \textit{ANLS} = \frac{1}{N} \sum_{i=0}^N \textit{s}(a_{i},o_{i})
\end{equation}

\begin{equation}
    \textit{s}(a_{i},o_{i}) = \left\{
  \begin{array}{@{}ll@{}}
    (\text{1}-\textit{NL}(a_{i},o_{i})), & \text{if}\ \textit{NL}(a_{i},o_{i}) < \tau \\
    0, & \text{if}\ \textit{NL}(a_{i},o_{i}) \geq \tau 
  \end{array}\right.
\end{equation}

\noindent
where \textit{N} is the number of the evaluated answers, $a_{i}$ and $o_{i}$ are the $i^{th}$ prediction and ground truth respectively where $\textit{i} = \{0,..., N\}$, $\textit{NL}(a_{i},o_{i})$ is the normalized Levenshtein distance between $a_{i}$ and $o_{i}$, which is a value between 0 and 1. The threshold $\tau$ determines whether the ANLS score is 0 or the value based on the normalized Levenshtein distance. We choose the original threshold of 0.5 for all evaluations. The final scores of the metric are the average score of each pair of the prediction and the corresponding ground truth.

\subsection{Experimental Setup}

We evaluated the baselines on both their base and large versions. However, LiLT and LayoutXLM were the exceptions since only their base versions are published. Also, CafeBERT, trained from the large version of XLM-Roberta, was considered to have only a large version. For generative models, base and large versions indicate the version of the ViT5 backbone. Regarding our LiGT model, we apply four hashing levels (\textit{L} = 4) and set an initial value of the learnable parameter ($\rho$) to 0.5.

All baselines are trained on batch size of 8, throughout 10 epochs. Adam optimizer \cite{adam} and linear scheduler are employed. All models are used $5e-5$ learning rate, 5,000 warm up steps, and cross entropy loss. The weights of all pretrained models were applied from HuggingFace \cite{huggingface}.

Due to high computational cost, we limit the max length of input to not exceed 180 tokens, including question and linearized OCR texts. This is a reasonable constraint since our extracted OCR contains 142 bounding boxes on average and the mean of our question lengths is 11.8.

In terms of categorizing the modalities, we refer to textual semantics (T), layout content (L), and visual information (V), presented in Table \ref{tab:main_eval}, as the abilities of a model capable of processing the corresponding modality. In other words, models having V modality means they process visual features either from a pretrained visual model or from their trainable visual components, L modality in essence is the utilization of bounding boxes to create 2D spatial features.

\subsection{Experimental Results}

\begin{table*}[hbt]
    \centering
    \small
    \caption{Main evaluation results of baselines on the ReceiptVQA test set. Modality T, L, and V denote the utilization of Textual semantics, Layout content, and Visual information respectively}
    \resizebox{\textwidth}{!}{
    \begin{tabular}{ccccccccc}
         \hline
         \multirow{2}{*}{\textbf{Model}} & \multirow{2}{*}{\textbf{Version}} & \multicolumn{3}{c}{\textbf{Modality}} && \multicolumn{3}{c}{\textbf{Metrics}} \\
         \cmidrule{3-5} \cmidrule{7-9}
           & & \textbf{T} & \textbf{L} & \textbf{V} & & \textbf{ANLS} & \textbf{F1} & \textbf{Accuracy}\\
         \hline
         \multicolumn{9}{l}{\textbf{\textit{Upper bound (UB) \& Heuristic (H)}}} \\
         \hline
         MATCHED\_OCR (UB)& & & & & & 76.77 & 76.72 & 76.77 \\
         
         MATCHED\_OCR\_AVG (UB)& & & & & & 74.63 & 74.59 & 74.63 \\
         
         RAND\_TOP10 (H)& & & & & & 4.62 & 1.54 & 1.48 \\
         
         RAND\_TOP100 (H)& & & & & & 2.71 & 0.41 & 0.31 \\
         \hline
         \multicolumn{9}{l}{\textbf{\textit{Extractive approach}}} \\
         \hline
         CafeBERT & \textit{large} & \ding{51} & & & & 56.59 & 52.67 & 50.40 \\
         \hline
         \multirow{2}{*}{PhoBERT} & \textit{base} & \multirow{2}{*}{\ding{51}} & & & & \underline{61.39} & \underline{57.55} & \underline{54.91} \\
                                                & \textit{large} & & & & & 57.87 & 53.93 & 51.34 \\
         \hline
         \multirow{2}{*}{XLM-Roberta} & \textit{base} & \multirow{2}{*}{\ding{51}} & & & & 58.00 & 54.49 & 52.28 \\
                                                       & \textit{large} & & & & & 57.81 & 53.76 & 50.97 \\
         \hline
         \multirow{2}{*}{InfoXLM} & \textit{base} & \multirow{2}{*}{\ding{51}} & & & & 58.33 & 54.49 & 51.89 \\
                                                & \textit{large} & & & & & 57.56 & 52.92 & 49.08\\
         \hline
         LiLT$_{[PhoBERT]}$ & \multirow{3}{*}{\textit{base}} & \multirow{3}{*}{\ding{51}} & \multirow{3}{*}{\ding{51}} & & & 59.87 & 56.29 & 54.05 \\
         
         LiLT$_{[XLM-Roberta]}$ & & & & & & 58.34 & 54.49 & 51.92\\
         
         LiLT$_{[InfoXLM]}$ & & & & & & 58.77 & 54.69 & 51.34 \\
        \hline
         LayoutXLM & \textit{base} & \ding{51} & \ding{51} & \ding{51} & & 59.11 & 55.45 & 53.05 \\
         \hline
         \multicolumn{9}{l}{\textbf{\textit{Generative approach}}} \\
         \hline
         \multirow{2}{*}{ViT5} & \textit{base} & \multirow{2}{*}{\ding{51}} & & & & 78.08 & 67.04 & 61.82 \\
                                            & \textit{large} & & & & &  77.22 & 66.69 & 61.60 \\
         \hline
         \multirow{2}{*}{ViT5+2D} & \textit{base} & \multirow{2}{*}{\ding{51}} & \multirow{2}{*}{\ding{51}} & & &  75.03 & 63.56 & 58.20  \\
                                            & \textit{large} & & & & &  74.08 & 63.17 & 58.08 \\
         \hline
         \multirow{2}{*}{ViT5+U} & \textit{base} & \multirow{2}{*}{\ding{51}} & \multirow{2}{*}{\ding{51}} & \multirow{2}{*}{\ding{51}} & &  \textbf{78.98} & \underline{67.89} & 62.46  \\
                                            & \textit{large} & & & & &  \underline{78.78} & \textbf{68.10} & \underline{62.80} \\
         \hline
         \multirow{2}{*}{ViT5+2D+U} & \textit{base} & \multirow{2}{*}{\ding{51}} & \multirow{2}{*}{\ding{51}} & \multirow{2}{*}{\ding{51}} & &  75.91 & 65.08 & 59.92 \\
                                                & \textit{large} & & & & &  73.56 & 62.41 & 57.43\\
        \hline
         \multirow{2}{*}{LaTr} & \textit{base} & \multirow{2}{*}{\ding{51}} & \multirow{2}{*}{\ding{51}} & \multirow{2}{*}{\ding{51}} & &  73.98 & 61.53 & 56.37\\
                                            & \textit{large} & & & & & 73.47 & 61.79 & 56.95\\
         \hdashline
         \multirow{2}{*}{\textbf{LiGT} (Ours)} & \textit{base} & \multirow{2}{*}{\ding{51}} & \multirow{2}{*}{\ding{51}} & & & \underline{78.78} & \underline{67.9} & 62.63  \\
                                                & \textit{large} & & & & & 78.64 & \textbf{68.09} & \textbf{63.02}\\
         \hline
      
    \end{tabular}}
    \label{tab:main_eval}
\end{table*}

Our main results are shown in Table \ref{tab:main_eval}. It can be seen that our LiGT model presented promising results. In particular, LiGT attained the highest F1, and Accuracy score compared to other baselines, which were competitive with ViT5+U which needs to train the U-Net architecture along with the ViT5 model. Furthermore, it is clear that generative models outperform extractive models significantly, demonstrating the feasibility of future investigations on architectures that can generate answers. In addition, heuristic baselines performed poorly on our dataset. This indicates that our dataset has a certain level of complication and requires properly-developed methodologies to handle it. Also, in terms of preciseness, there existed a remarkable performance gap between the upper bounds and the models, which suggests a certain level of challenge in our dataset.

\subsection{Experimental Analysis}

\subsubsection{Impact of Modalities on Extractive Baselines}

Regarding the extractive approach in Table \ref{tab:main_eval}, text-only baselines showed competitive results compared with other multimodal baselines. In particular, there was only a marginal gap between the two multilingual models, XLM-Roberta, and InfoXLM, and the multimodal models. Moreover, the native pretrained PhoBERT achieved a significant performance, surpassing LayoutXLM, and its LiLT-based layout extension, LiLT$_{[PhoBERT]}$. 

In addition, all large extractive models performed considerably worse than their base versions, which is not a regular phenomenon in NLP tasks. For this reason, we speculate that the linearized OCR context could hinder the models' inherent capabilities of understanding semantic properties, since the OCR context lacks natural language formats which are often characterized by phrases, sentences, and paragraphs.

These observations indicate that Textual modality plays a crucial role in understanding the receipt domain but it is not sufficient for models to comprehend the OCR context. Therefore, investigating extractive multimodal methodologies is necessary, and there still remains plenty of room for improvement.

\subsubsection{Impact of Modalities on Generative Baselines}

In terms of generative performances, Table \ref{tab:main_eval} shows that ViT5, a plain-text model, gained remarkably high scores in comparison to other multimodal models. In particular, ViT5 outperformed ViT5+2D, ViT5+2D+U, and LaTr, despite having lower scores than ViT5+U and LiGT. Observing ViT5+2D, ViT5+2D+U, and LaTr, we found that these models have one similarity of applying fully new additional embedding layers for tackling layout understanding. This suggests that inducing fully new embedding layers to pretrained models might result in incompatibility between the neural modules when tackling the receipt domain in Vietnamese, causing difficulties for the models to converge. On the other hand, it can be seen that leveraging trained embedding layers of pretrained models to incorporate layout understanding, such as LiGT, gained promising improvements in comparison with the text-only ViT5 model.

In spite of achieving a competitive performance, ViT5 still had its large version limited under its base version, which is similar to the extractive text-only models. Meanwhile, ViT5+U and LiGT showed prominent performances on their large models, which suggests that generative models might also require multimodal solutions to tackle the Vietnamese receipt domain comprehensively.

Results in Table \ref{tab:main_eval} also suggest that inducing multimodal methodologies could be necessary but might need thorough development to adapt to particular Vietnamese contexts. As mentioned in Section \ref{sec:Q_classify}, there are considerable differences between English and Vietnamese, from the grammatical perspectives of questions to the variety of interrogative words. On that account, understanding the inquiries and OCR context could be dissimilar in English pretrained models and Vietnamese pretrained models. This could be the reason why some multimodal models such as ViT5+2D+U, and ViT5+2D had poor performances in comparison with other generative models, although they have shown outstanding results in many document datasets in English. 

Furthermore, in our research scope, we excluded the multimodal pre-training process of these model architectures due to its high computational cost. Hence, results when the models have pre-training process could be different. We hope future extensions could investigate these aspects more comprehensively. 

\subsubsection{Analysis of Experimental Results on Question Types} \label{sec:details_q}

We further analyzed the results via question types mentioned in Section \ref{sec:Q_classify} and Section \ref{sec:Further_Stat}. As the number of examined baselines is quite huge, we selectively assessed on four extractive models (PhoBERT$_{base}$, PhoBERT$_{large}$, LiLT$_{[PhoBERT]}$, and LayoutXLM), and four generative models (ViT5+U$_{base}$, ViT5+U$_{large}$, LiGT$_{base}$, and LiGT$_{large}$) that have prominent performances. For ease of visualization, we also reduced the number of question types down to five types by considering types Reason, Manner, and Other, which have few samples, as one type named Minors.

\begin{figure*}[hbt]
\centering
\begin{subfigure}{0.23\linewidth}
  \centering
  \includegraphics[width=\linewidth]{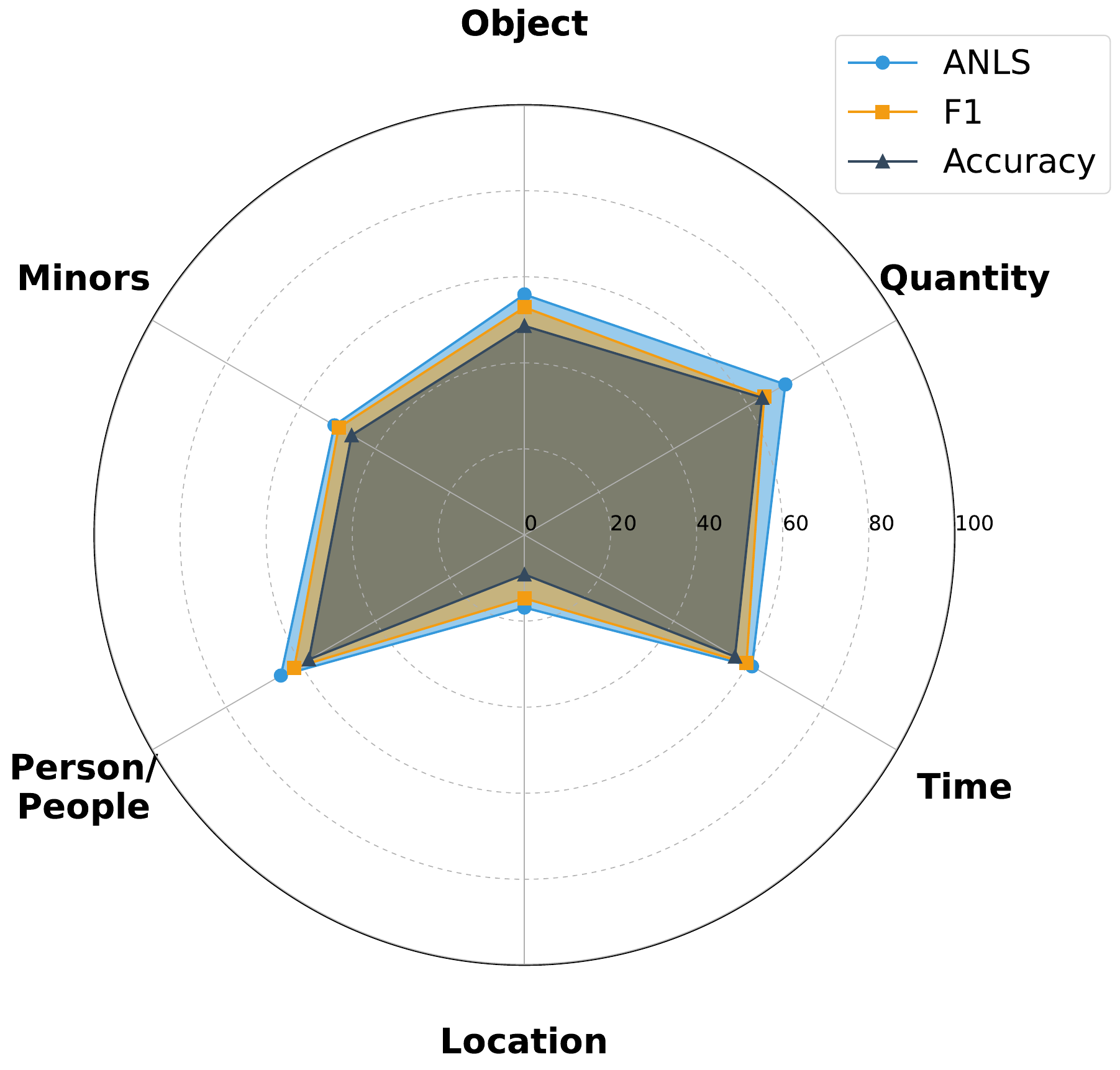}
  \caption{PhoBERT$_{base}$}
  \label{fig:Q_Phobert_Base}
\end{subfigure}%
\hspace{0.02\linewidth}
\begin{subfigure}{0.23\linewidth}
  \centering
  \includegraphics[width=\linewidth]{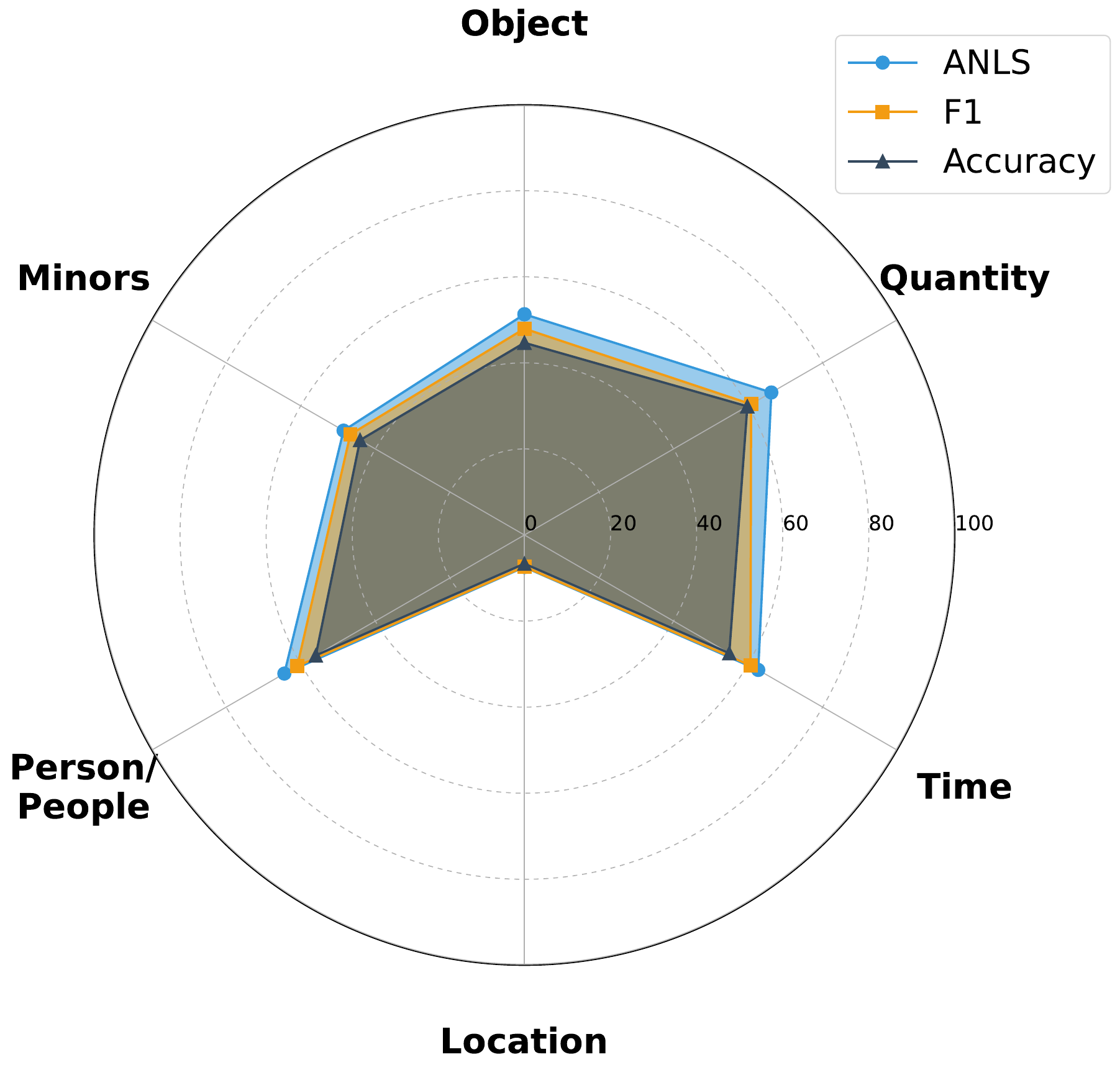}
  \caption{PhoBERT$_{large}$}
  \label{fig:Q_Phobert_Large}
\end{subfigure}%
\hspace{0.02\linewidth}
\begin{subfigure}{0.23\linewidth}
  \centering
  \includegraphics[width=\linewidth]{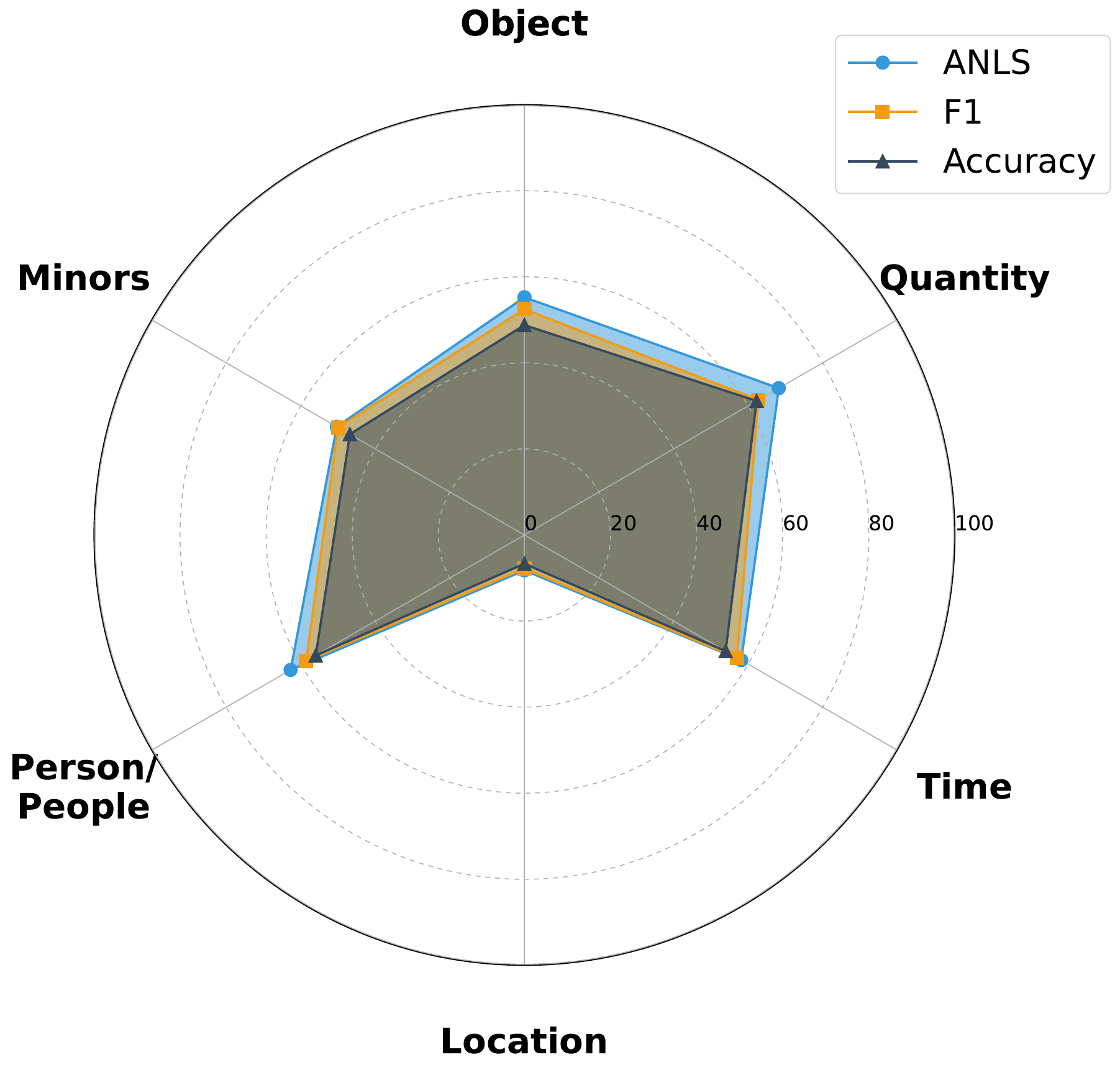}
  \caption{LiLT$_{[PhoBERT]}$}
  \label{fig:Q_LiLT_PhoBERT}
\end{subfigure}%
\hspace{0.02\linewidth}
\begin{subfigure}{0.23\linewidth}
  \centering
  \includegraphics[width=\linewidth]{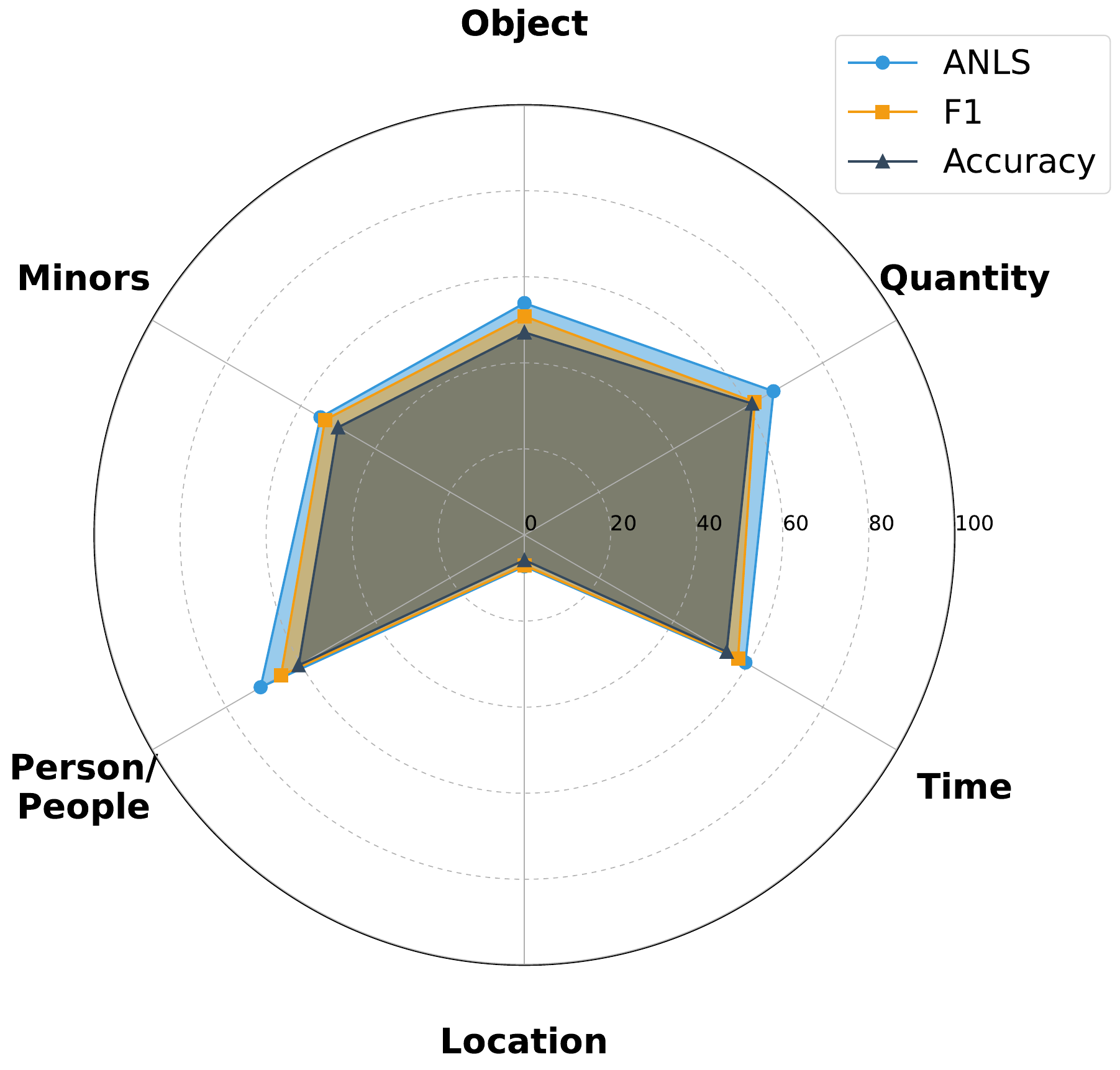}
  \caption{LayoutXLM}
  \label{fig:Q_LayoutXLM}
\end{subfigure}

\caption*{\centering Extractive Methods}\label{fig:Q_R_A_ex} 
\vspace{0.01\linewidth}

\begin{subfigure}{0.23\linewidth}
  \centering
  \includegraphics[width=\linewidth]{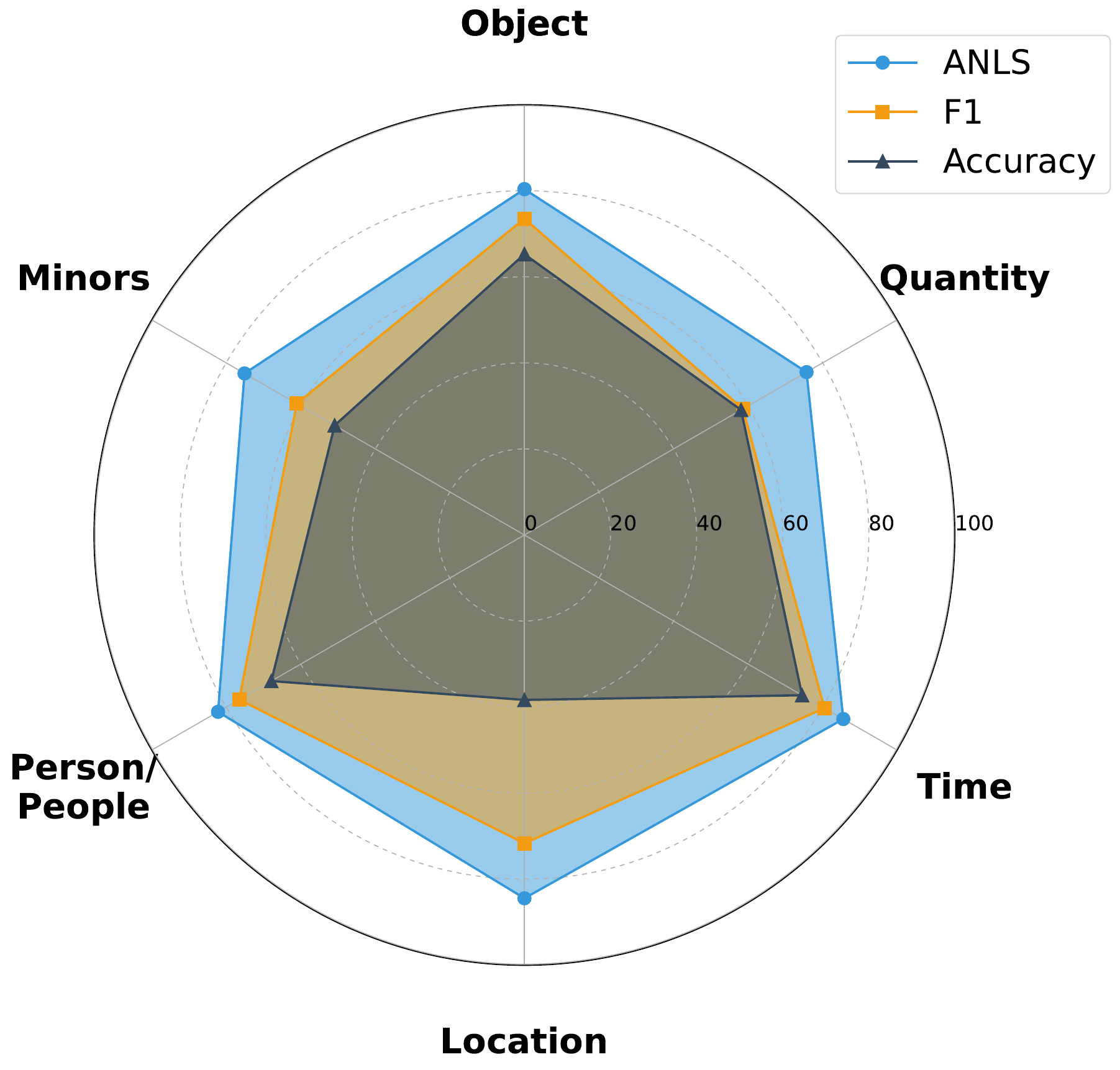}
  \caption{ViT5+U$_{base}$}
  \label{fig:Q_ViT5U_Base}
\end{subfigure}%
\hspace{0.02\linewidth}
\begin{subfigure}{0.23\linewidth}
  \centering
  \includegraphics[width=\linewidth]{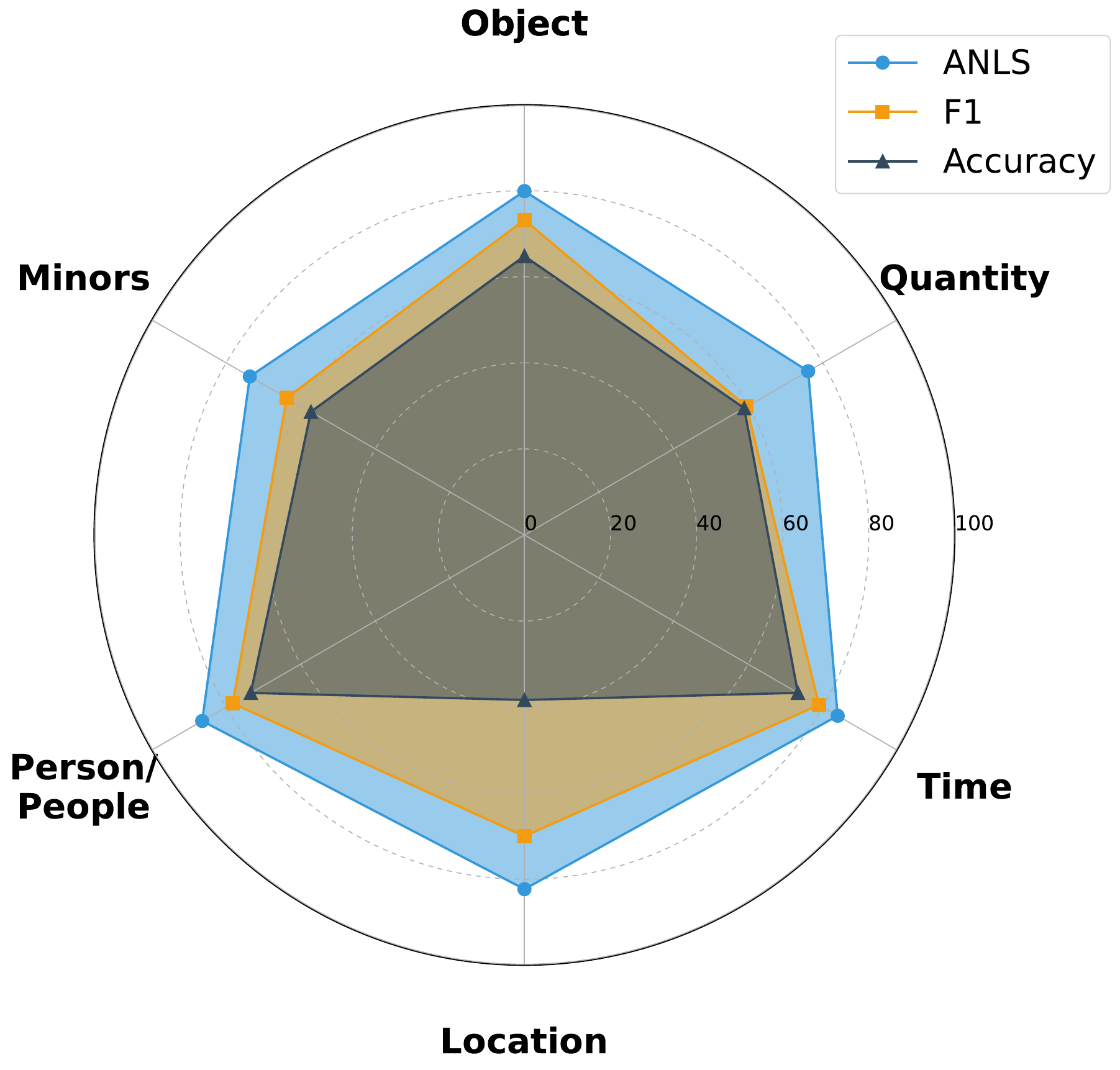}
  \caption{ViT5+U$_{large}$}
  \label{fig:Q_ViT5U_Large}
\end{subfigure}%
\hspace{0.02\linewidth}
\begin{subfigure}{0.23\linewidth}
  \centering
  \includegraphics[width=\linewidth]{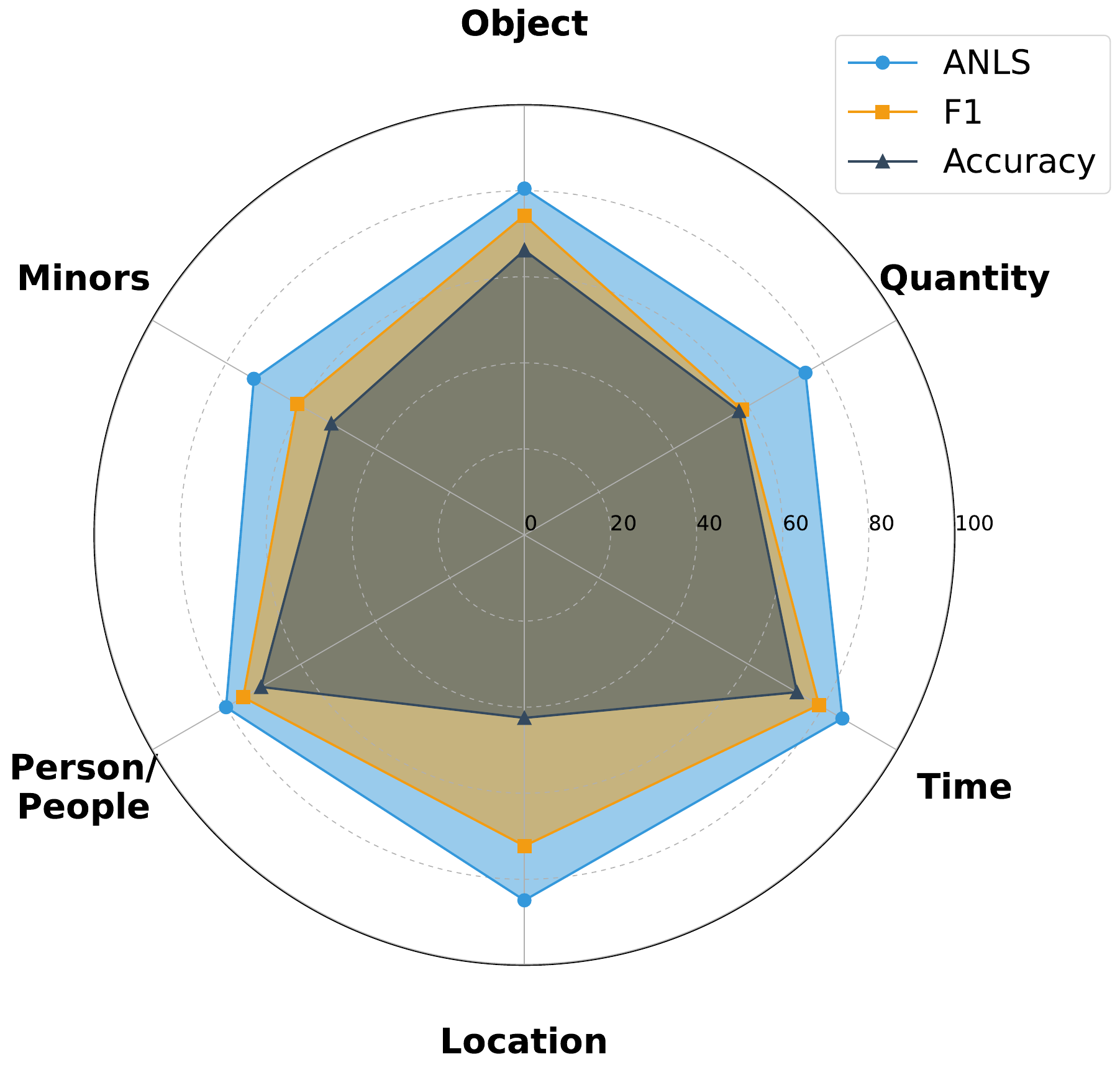}
  \caption{LiGT$_{base}$}
  \label{fig:Q_LiGT_base}
\end{subfigure}%
\hspace{0.02\linewidth}
\begin{subfigure}{0.23\linewidth}
  \centering
  \includegraphics[width=\linewidth]{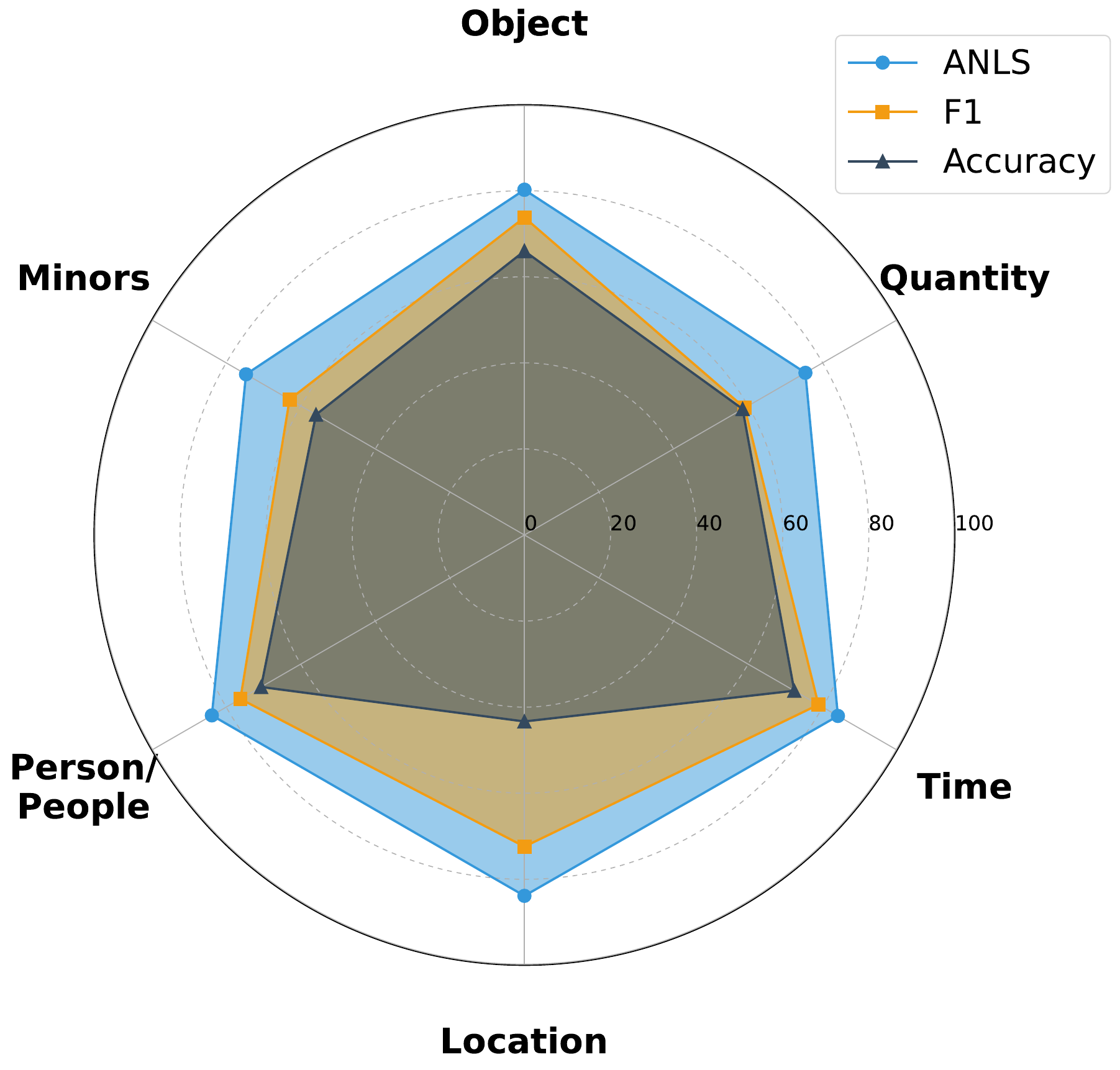}
  \caption{LiGT$_{large}$}
  \label{fig:Q_LiGT_large}
\end{subfigure}

\caption*{\centering Generative Methods}\label{fig:Q_R_A_gen} 
\vspace{0.02\linewidth}

\caption{Main results on question types according to selected extractive and generative methods. Minors type encompasses Reason, Manner, and Other, which are types having few samples}
\label{fig:Q_R_A}
\end{figure*}

Figure \ref{fig:Q_R_A} shows that extractive models have similar patterns. Referring to Figure \ref{fig:Questiontypes_Questionnums}, the models performed better on question types such as Object, Quantity, and Time, which have a high frequency of annotations. Interestingly, the People/Person type, which has remarkably lower frequency than the aforementioned types, was inferred with higher scores than the others. It could be explained that answers about a person or a group of people are often either human names or short ID codes, which are apparently more recognizable than other forms of text. On the other hand, models struggled with questions about locations although they have relatively equal frequency compared with questions in the People/Person type. Considering Figure \ref{fig:Alen_Qtype}, it can be seen that Location questions require significantly longer answers. This could be the main reason causing poor performance of the models, especially with such an extractive approach.

Regarding the generative approach, Figure \ref{fig:Q_R_A} shows that generative models have remarkably high potential to compensate for the downside of the extractive models. In particular, performances on Location questions were handled with significant improvements. Considering the ANLS metric, it can be observed that generative models evenly handle all types of questions, which include Minors questions with few samples. Nonetheless, while generative models' performance in ANLS approximated 80\%, the preciseness of the models (represented by Accuracy, and F1) still requires considerable refinement and enhancement. Accuracy scores in Location questions were only around 40\%, remaining a significant gap between them and the F1, and ANLS metrics. Additionally, Accuracy and F1 scores in Quantity questions were only around 60\% although this type has the shortest average length of the answers.

Besides, Figure \ref{fig:Q_R_A} shows that LiGT was competitive with ViT5+U in all question types, considering ViT5+U is applied to all the Text, Layout, and Visual modalities. However, our model has not provided exclusive solutions for current problems that the baseline models have faced, also struggling with the Location type. This could be a potential path for our future extensions.

\subsubsection{Analysis of Experimental Results on Answer Types}

\begin{figure*}[hbt]
\centering
\begin{subfigure}[t]{\textwidth}
    \centering
    \begin{subfigure}{0.23\linewidth}
      \centering
      \includegraphics[width=\linewidth]{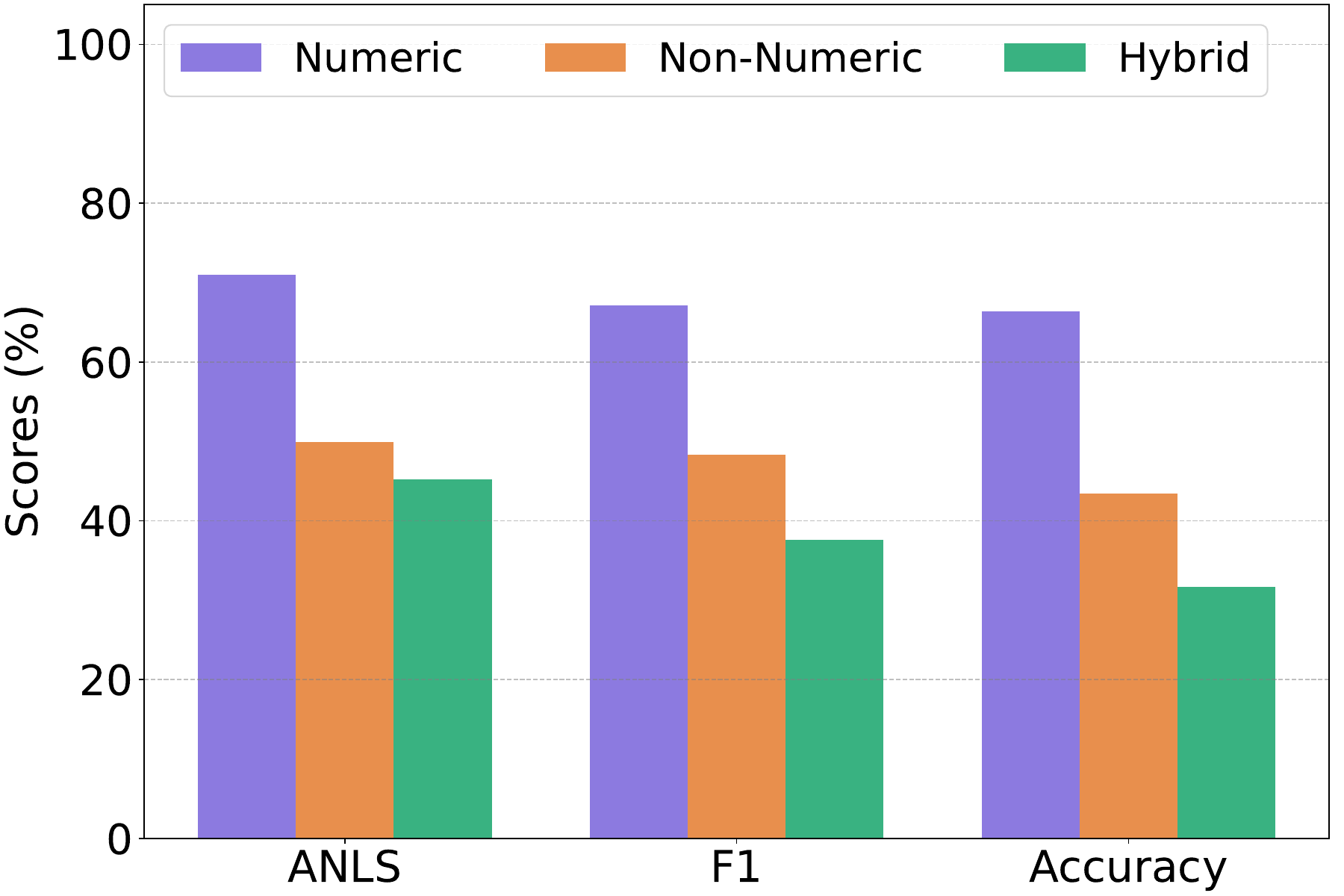}
      \caption{PhoBERT$_{base}$}
      \label{fig:A_Phobert_base}
    \end{subfigure}%
    \hspace{0.02\linewidth}
    \begin{subfigure}{0.23\linewidth}
      \centering
      \includegraphics[width=\linewidth]{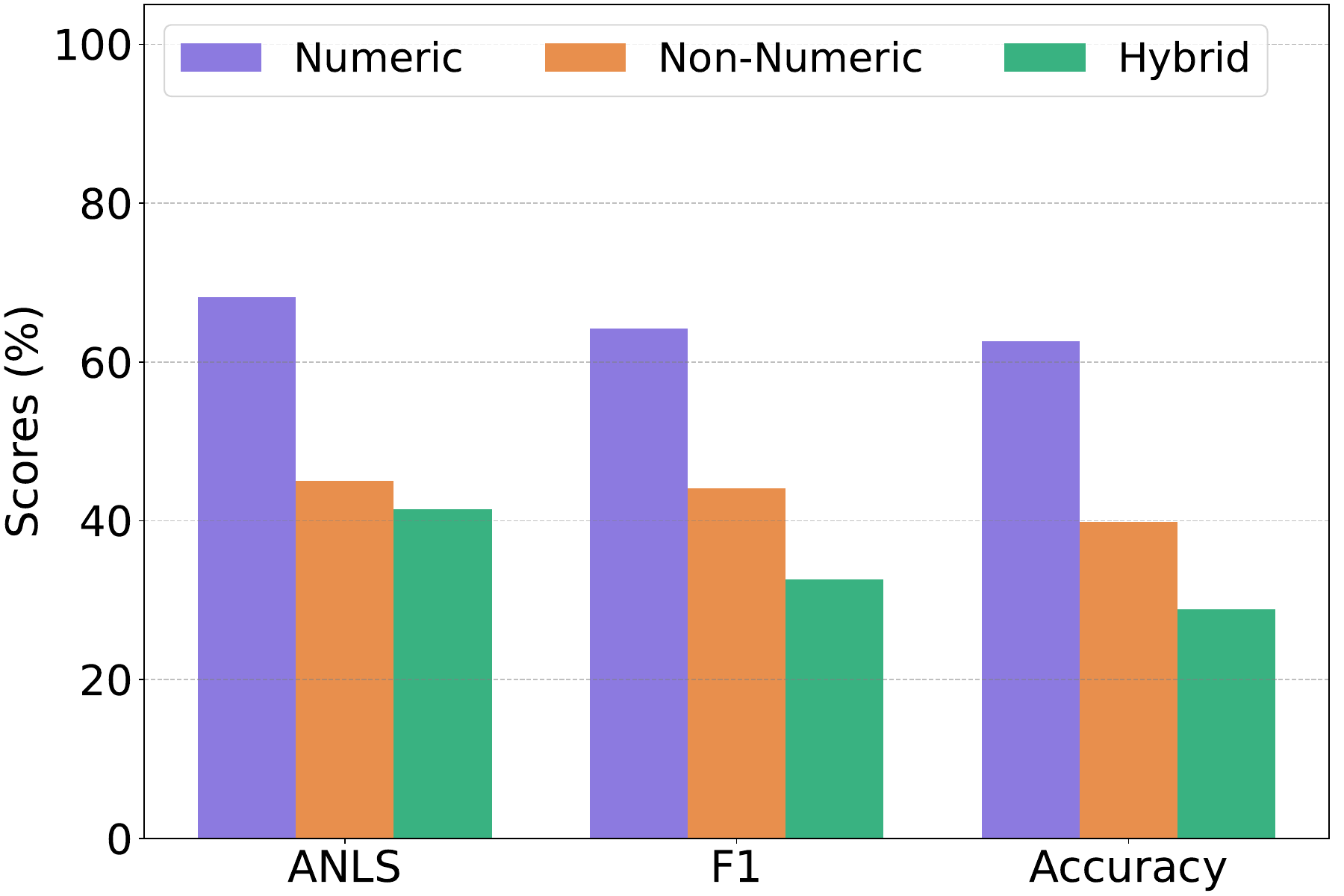}
      \caption{PhoBERT$_{large}$}
      \label{fig:A_Phobert_large}
    \end{subfigure}%
    \hspace{0.02\linewidth}
    \begin{subfigure}{0.23\linewidth}
      \centering
      \includegraphics[width=\linewidth]{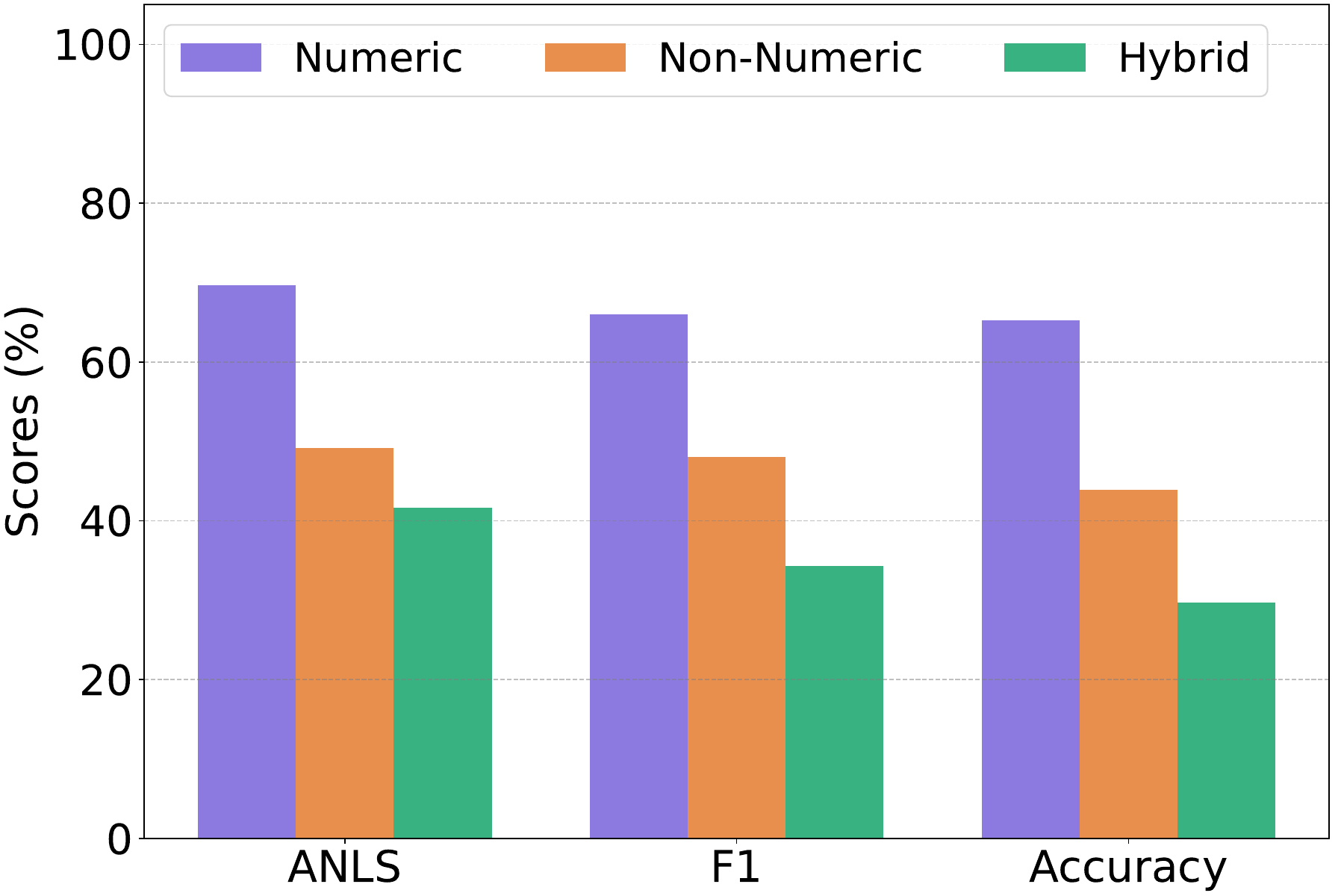}
      \caption{LiLT$_{[PhoBERT]}$}
      \label{fig:A_LiLT_PhoBERT}
    \end{subfigure}%
    \hspace{0.02\linewidth}
    \begin{subfigure}{0.23\linewidth}
      \centering
      \includegraphics[width=\linewidth]{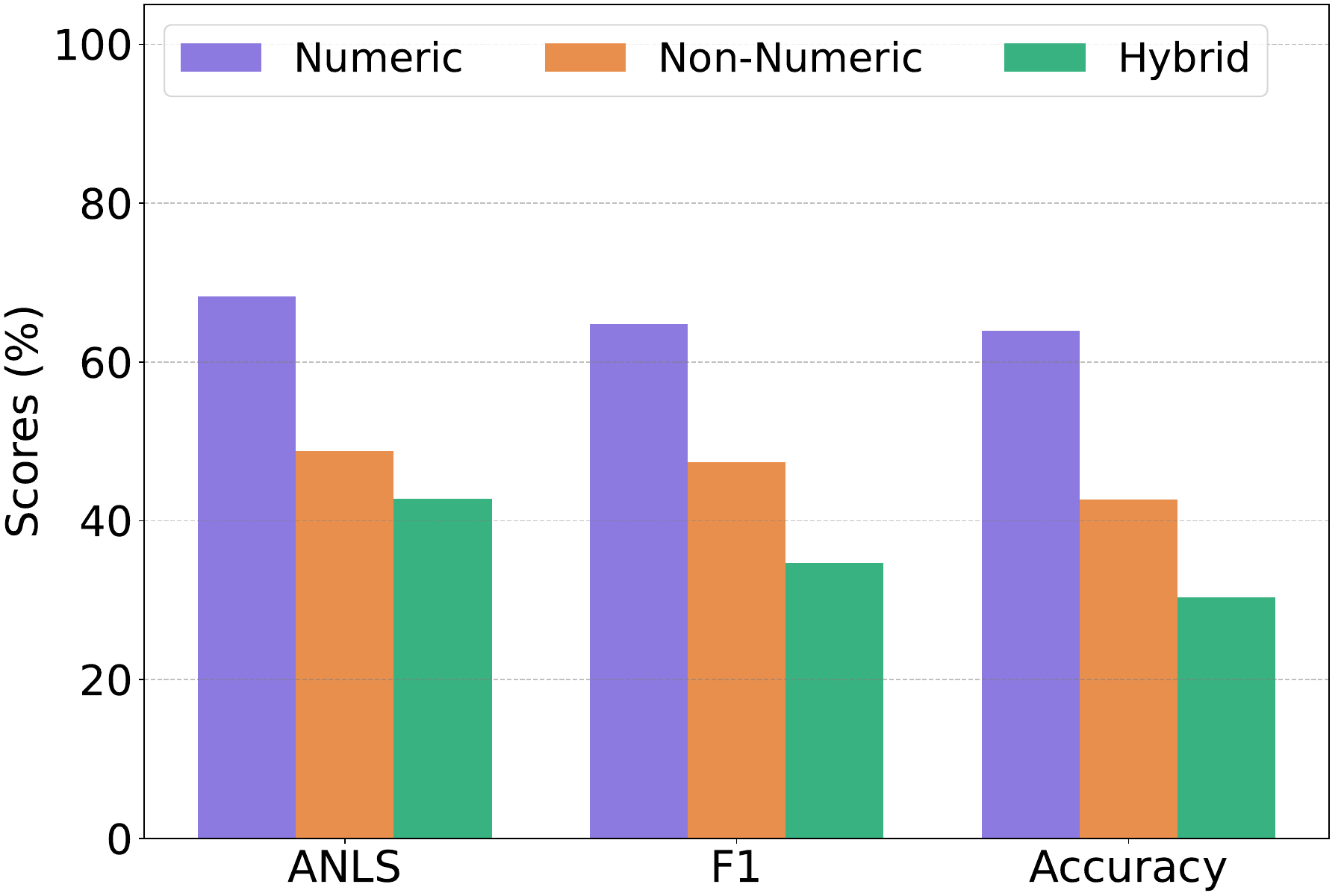}
      \caption{LayoutXLM}
      \label{fig:A_LayoutXLM}
    \end{subfigure}
    \caption*{\centering Extractive Methods}
    \label{fig:extractive_methods}
\end{subfigure}

\vspace{0.01\linewidth}

\begin{subfigure}[t]{\textwidth}
    \centering
    \begin{subfigure}{0.23\linewidth}
      \centering
      \includegraphics[width=\linewidth]{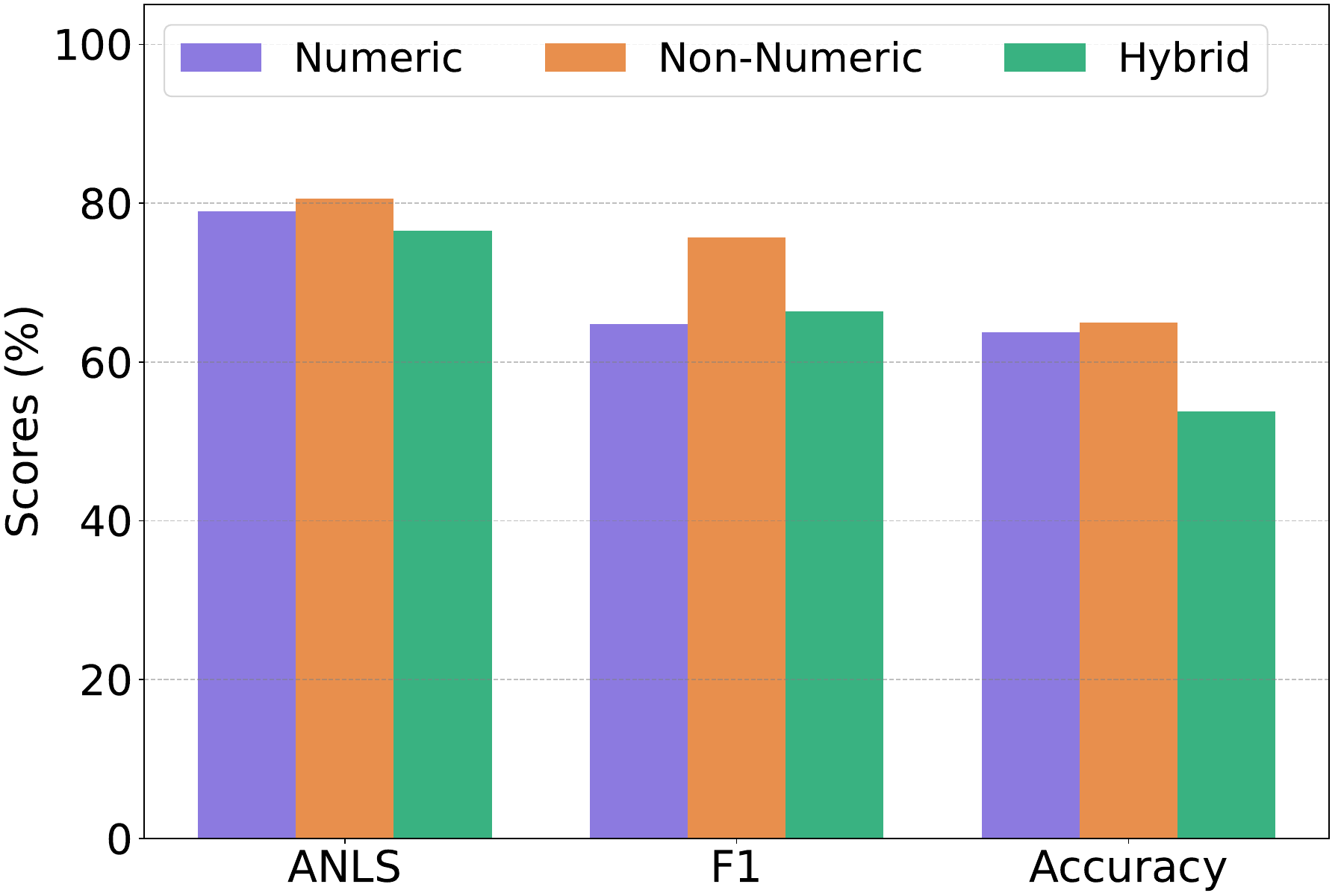}
      \caption{ViT5+U$_{base}$}
      \label{fig:A_ViT5U_base}
    \end{subfigure}%
    \hspace{0.02\linewidth}
    \begin{subfigure}{0.23\linewidth}
      \centering
      \includegraphics[width=\linewidth]{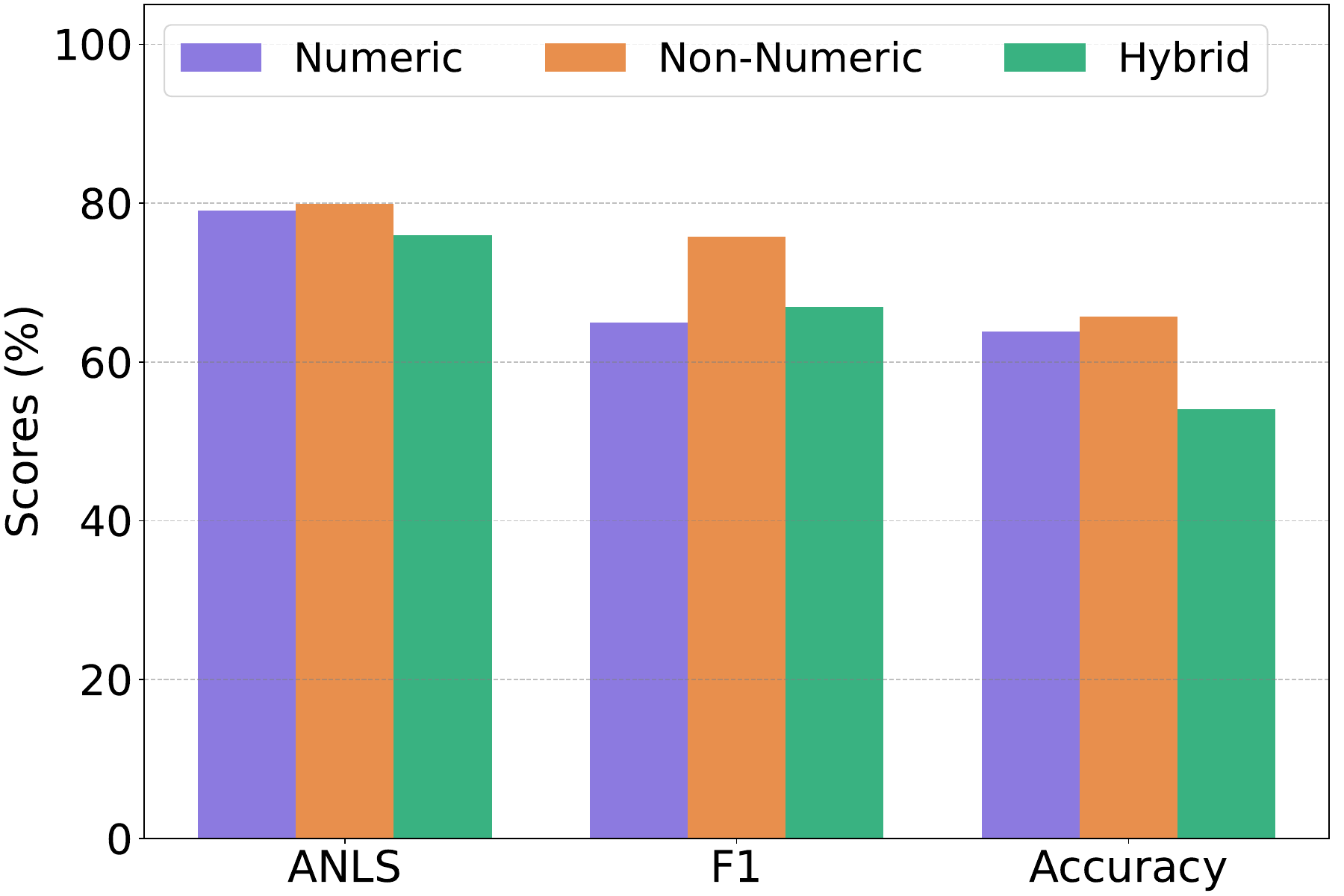}
      \caption{ViT5+U$_{large}$}
      \label{fig:A_ViT5U_large}
    \end{subfigure}%
    \hspace{0.02\linewidth}
    \begin{subfigure}{0.23\linewidth}
      \centering
      \includegraphics[width=\linewidth]{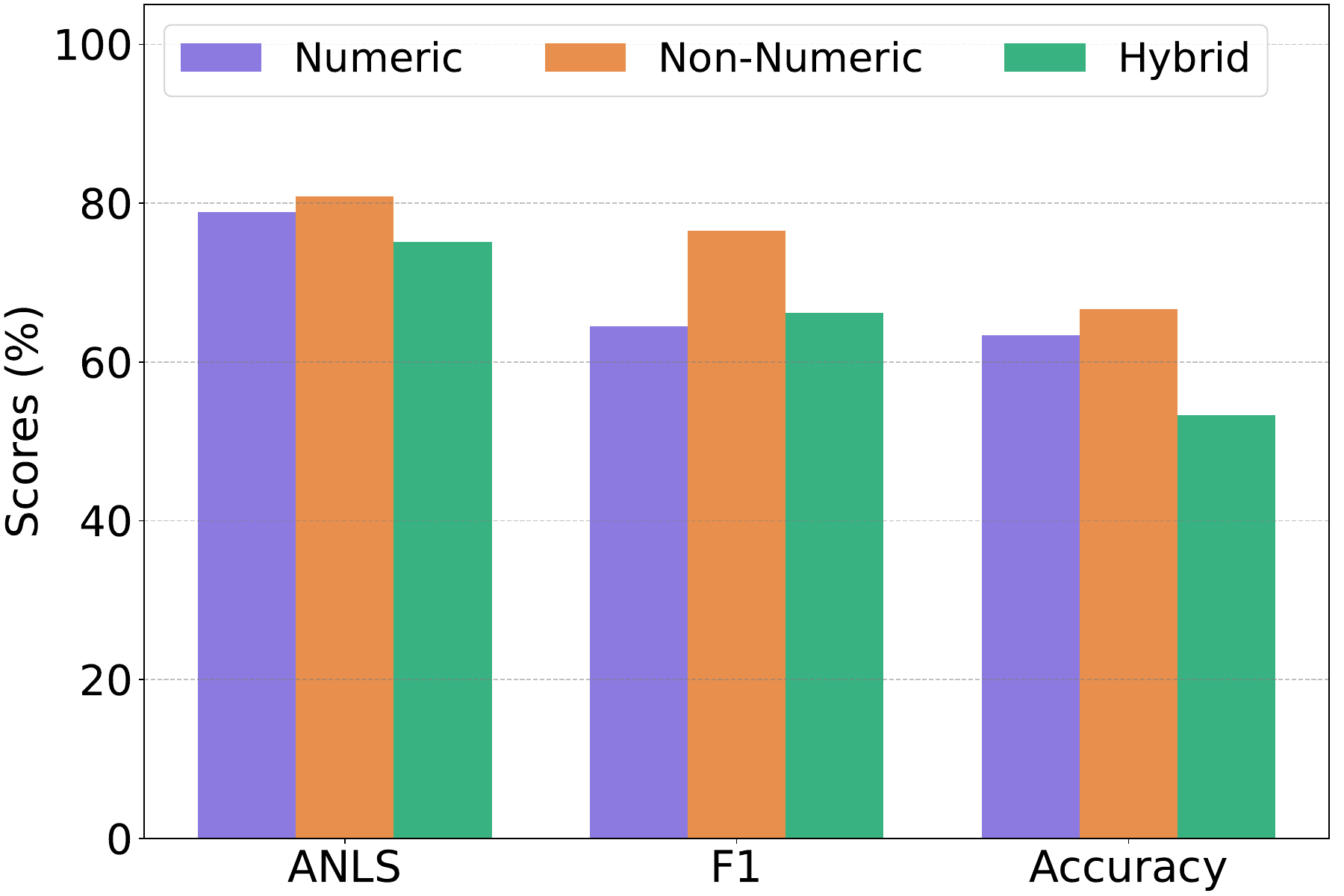}
      \caption{LiGT$_{base}$}
      \label{fig:A_LiGT_base}
    \end{subfigure}%
    \hspace{0.02\linewidth}
    \begin{subfigure}{0.23\linewidth}
      \centering
      \includegraphics[width=\linewidth]{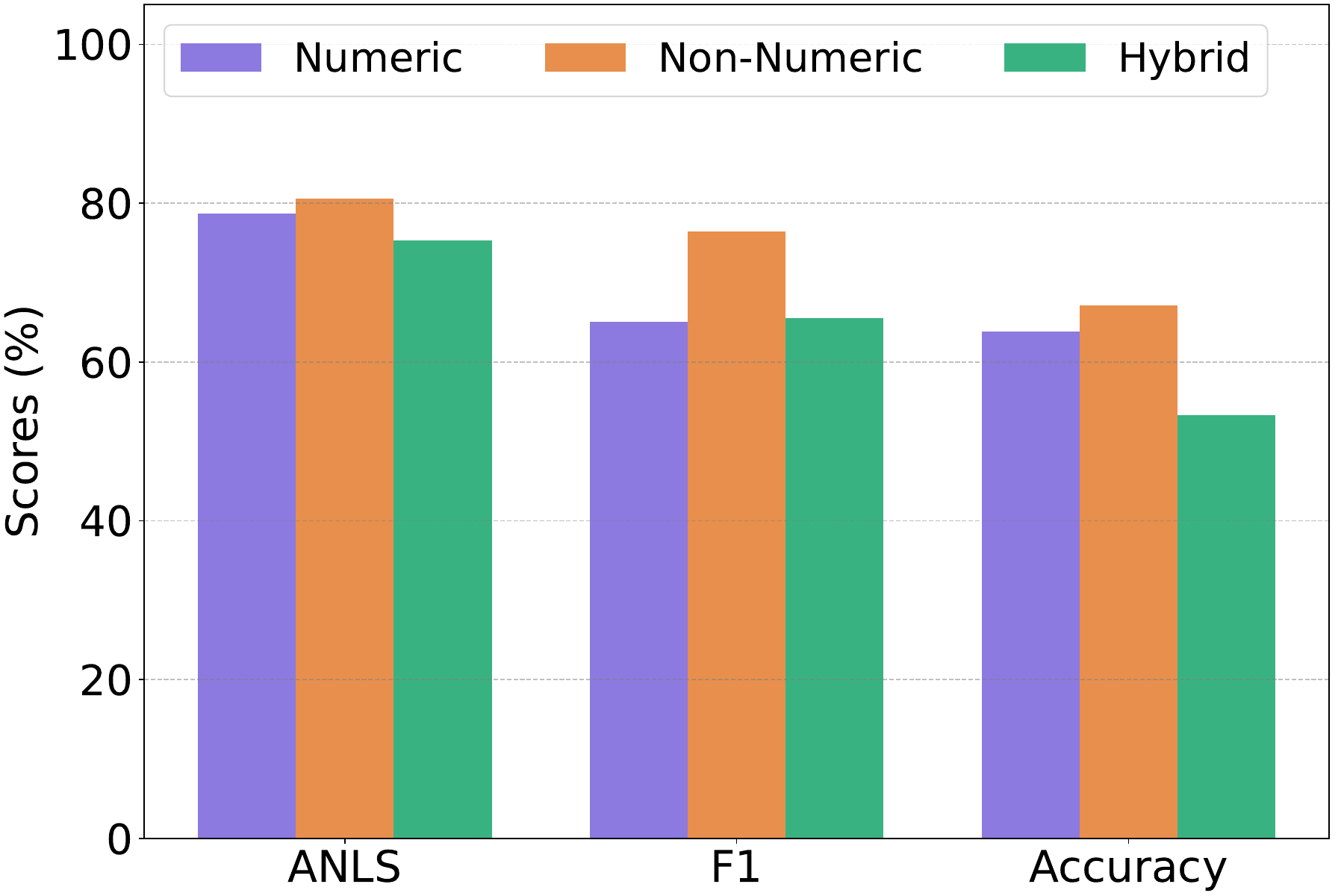}
      \caption{LiGT$_{large}$}
      \label{fig:A_LiGT_large}
    \end{subfigure}
    \caption*{\centering Generative Methods}
    \label{fig:generative_methods}
\end{subfigure}

\caption{Main results on answer types according to selected extractive and generative methods. The extractive methods are presented in the first row, and the generative methods are shown in the second row}
\label{fig:A_R_A}
\end{figure*}

Referring to the answer types mentioned in Section \ref{sec:Further_Stat}, we continued to analyze our main results of the aforementioned selected models on the three answer types. Figure \ref{fig:A_R_A} shows that the extractive models' results resemble each other. The models gained the highest scores when using Numeric answers, and the lowest scores when facing Hybrid answers. Observing Figure \ref{fig:ATypeVio}, the Hybrid type has the widest range of length, followed by the Non-Numeric type and Numeric type. This indicates that extractive models perform worse when answer lengths increase. Considering similar issues in the question analysis \ref{sec:details_q}, handling long answers seems to be one of the most considerable challenges for extractive models.

With regard to generative models, Figure \ref{fig:A_R_A} presents an astonishing pattern in comparison to the extractive results' pattern, wherein all selected generative models achieved the highest scores in Non-Numeric answers. A possible explanation for these results might be the ability to generate letters from ViT5, which is pretrained on a large corpus of texts. Moreover, although most of the models still had the lowest performances in Hybrid answers, the gaps between the Hybrid's scores and Numeric's scores are relatively small, compared with that of extractive models. This suggests that generative models have a certain level of resilience against answer lengths.

\subsubsection{Effect of LayoutHEI on text-only extractive baselines} \label{sec:Ef_LayoutHEI}

To have a more comprehensive view of our layout approach, we applied the LayoutHEI component, presented in Section \ref{sec:LayoutHEI}, of our LiGT model on extractive baselines. We replaced the Language Model component with chosen text-only extractive baselines in our main evaluation. The procedure for preparing answers was followed by Section \ref{sec:Ex_baseline}. Each LayoutHEI-employed variants (denoted by \textit{+LayoutHEI}) were compared with its text-only version in Table \ref{tab:main_eval} to view the differences in performance.

\begin{table*}[hbt]
    \centering
    \caption{Effect of LayoutHEI on text-only extractive baselines. Values in parentheses show differences in model performances compared to corresponding scores of text-only models}
    \resizebox{\textwidth}{!}{%
    \begin{tabular}{cccccc}
         \hline
         \multirow{2}{*}{\textbf{Model}} & \multirow{2}{*}{\textbf{Version}} & & \multicolumn{3}{c}{\textbf{Metrics}} \\
         \cmidrule{4-6}
           & & & \textbf{ANLS} & \textbf{F1} & \textbf{Accuracy}\\
         \hline
         \multirow{2}{*}{PhoBERT+LayoutHEI} & \textit{base} & & 63.11 (\textcolor{blue}{$\uparrow$ 1.72}) & 58.83 (\textcolor{blue}{$\uparrow$ 1.28}) & 55.26 (\textcolor{blue}{$\uparrow$ 0.35})\\  
                                                & \textit{large} & & 61.06 (\textcolor{blue}{$\uparrow$ 3.19}) & 56.69 (\textcolor{blue}{$\uparrow$ 2.76}) & 53.15 (\textcolor{blue}{$\uparrow$ 1.81})\\
         \hline
         \multirow{2}{*}{XLM-Roberta+LayoutHEI} & \textit{base} & & 58.75 (\textcolor{blue}{$\uparrow$ 0.75}) & 54.95 (\textcolor{blue}{$\uparrow$ 0.46}) & 52.34 (\textcolor{blue}{$\uparrow$ 0.06})\\
                                                       & \textit{large} & & 57.38 (\textcolor{red}{$\downarrow$ 0.43}) & 53.01 (\textcolor{red}{$\downarrow$ 0.75}) & 50.48 (\textcolor{red}{$\downarrow$ 0.49})\\
         \hline
         \multirow{2}{*}{InfoXLM+LayoutHEI} & \textit{base} & & 59.06 (\textcolor{blue}{$\uparrow$ 0.73}) & 54.63 (\textcolor{blue}{$\uparrow$ 0.14}) & 52.00 (\textcolor{blue}{$\uparrow$ 0.11})\\
                                                & \textit{large} & & 57.88 (\textcolor{blue}{$\uparrow$ 0.32}) & 54.32 (\textcolor{blue}{$\uparrow$ 1.40}) & 50.38 (\textcolor{blue}{$\uparrow$ 1.30})\\
         \hline
         CafeBERT+LayoutHEI & \textit{large} & & 58.36 (\textcolor{blue}{$\uparrow$ 1.77}) & 54.15 (\textcolor{blue}{$\uparrow$ 1.48}) & 51.17 (\textcolor{blue}{$\uparrow$ 0.77})\\
         \hline
    \end{tabular}%
    }
    \label{tab:effect_ex}
\end{table*}

Table \ref{tab:effect_ex} demonstrates that our layout approach has significant effects on various text-only baselines, achieving enhancements on most of the models. In particular, the large version PhoBERT+LayoutHEI gained a marked improvement of 3.19\% ANLS, 2.76\% F1, and 1.81\% Accuracy, while its base version had the highest score of all extractive baselines in both Table \ref{tab:main_eval} and Table \ref{tab:effect_ex}. Moreover, it can be seen that models pretrained on Vietnamese data benefit more clearly than the multilingual models. This could be the reason why large XLM-Roberta+LayoutHEI witnessed a performance decrease in all metrics, while CafeBERT, the large XLM-Roberta additionally pretrained on Vietnamese data, gained a significant improvement on its LayoutHEI-employed variant, CafeBERT+LayoutHEI. Besides, the results in Table \ref{tab:effect_ex} were also evidence supporting the need for multimodal methodologies for tackling the Vietnamese receipt domain. In addition, we examined the effect of using more specific sources for the visual modality in Appendix \ref{sec:secA3}.

Besides, we conducted ablation analyses on the efficiency of layout understanding of LiGT in Appendix \ref{sec:secA2}. The analyses explored the effect of hashing levels and the impact of the learnable ratio ($\omega$) on the performance of LayoutHEI-employed models.

\subsubsection{LiGT performance on standard document VQA benchmarks} \label{sec:standard_docvqa}

We further evaluate our LiGT architecture on two standard document VQA datasets, DocVQA \cite{mathew2021docvqa} and InfographicVQA \cite{InfographicVQA}. We applied three English pre-trained models, T5 \cite{t5model}, BART \cite{bart}, and Flan-T5 \cite{flant5}, to our LiGT language model component. While T5 and BART are two of the most reputed encoder-decoder models, Flan-T5 models are T5 extensions that applied instruction finetuning on various aspects, achieving outperforming results compared with their T5 counterparts. To reduce the computational cost and time, we only used base versions of the chosen models.

The annotations and extracted OCR properties available were obtained from the website of the Robust Reading Competition\footnote{\url{https://rrc.cvc.uab.es/?ch=17}}. All LiGT models only underwent one training phase using the training and development sets (detailed configurations in Appendix \ref{sec:secA4_1}) and were evaluated on the test sets directly from the evaluation system of the website using the original ANLS \cite{biten2019scene}. In these accounts, we only compare our results with those of baselines without pre-training phases from the datasets' paper. We leave more comprehensive assessments for future investigations, where pre-training procedures and the effect of the extracted OCR quality would be taken into consideration.

Table \ref{tab:standard_docvqa} presents the performance of our LiGT models in comparison to those from the chosen baselines, M4C \cite{m4c}, BERT \cite{bert}, and LayoutLM \cite{LayoutLM}. Details of LiGT performance on different aspects of the datasets are also provided in Appendix \ref{sec:secA4_2}. It can be seen that our models showed promising results under the circumstances of one training phase. In particular, while LiGT$_{[T5-base]}$ had marginal impact, LiGT$_{[Flan-T5-base]}$ attained 63.8\% ANLS on DocVQA (surpassing BERT$_{large}$ by 2.8\%) and 30.7\% ANLS on InfographicVQA (8.2\% higher than that of LayoutLM). Interestingly, although BART$_{base}$ ($\sim$140M) has fewer trainable parameters than T5$_{base}$ ($\sim$220M) has, its BART's LiGT variant had remarkably higher performance than that of T5, which would indicate the potential of BART architecture for layout understanding. Overall, the performance of LiGT demonstrates the feasibility of our proposed architecture. Besides that, these results also emphasized the effectiveness of generative models in comparison to extractive models such as BERT and LayoutLM.

\begin{table}[hbt]
    \centering
    \caption{Performance of our LiGT architecture with different language models on the test sets of the DocVQA and InfographicVQA (InfoVQA) benchmark}
        \begin{tabularx}{\columnwidth}{ccc}
             \hline
             \multirow{2}{*}{\textbf{Models}} & \textbf{DocVQA} & \textbf{InfoVQA}\\
             \cmidrule{2-2} \cmidrule{3-3} 
              &  \textit{ANLS} & \textit{ANLS}\\   
             \hline
             \multicolumn{3}{l}{\textbf{\textit{Baselines}}} \\
             \hline
             M4C \cite{m4c}  & 39.1 &  14.7 \\
             \hline
             BERT$_{base}$/BERT$_{large}$ \cite{bert} & 57.4/61.0  & -  \\
             \hline
             LayoutLM \cite{LayoutLM} & - & 22.5 \\
             \hline
             \multicolumn{3}{l}{\textbf{\textit{LiGT models}}} \\
             \hline
             LiGT$_{[T5-base]}$  & 57.6 & 22.4 \\
             \hline
             LiGT$_{[BART-base]}$ & 59.3 & 27.1 \\
             \hline
             LiGT$_{[Flan-T5-base]}$ & \textbf{63.8} & \textbf{30.7} \\
             \hline
        \end{tabularx}
    \label{tab:standard_docvqa}
\end{table}

\section{Conclusion and Future Works}\label{sec:Conclusions}

In this work, we presented ReceiptVQA, the first Vietnamese document VQA dataset for receipts. ReceiptVQA is a large-scale dataset containing 9,500 captured receipt images and 64,812 manually annotated question-answer pairs. Also, we introduced LiGT, a novel spatial-aware architecture with a backbone of an encoder-decoder transformer. Throughout the experiments, our LiGT model showed promising results, achieving competitive performances in comparison with other baselines. Moreover, applying LayoutHEI, the layout operating component of LiGT, to text-only extractive models showed significant enhancements. 

By comparing the extractive approach with the generative approach, we observed that the generative models outperformed the extractive models in many aspects, especially in handling long answers. Furthermore, we found that tackling the Vietnamese receipt domain requires proper multimodal extension, although text modality was shown to play a highly vital role in the task.

In spite of having a remarkably large number of annotations, our dataset may not be sufficient to reflect real-world scenarios as there exists imbalances in question types and answer types, and a relatively lower number of images compared to English document VQA datasets. Future work could address these limitations and expand the dataset to encompass more images with diverse inquiries. 

In addition, evaluating our LiGT architecture on two standard document datasets, DocVQA and InfographicVQA, also showed promising results, demonstrating that our method has a certain level of applicability to different document scenarios. Nonetheless, as our main research scope was dedicated to the ReceiptVQA dataset and to explore its characteristics comprehensively, the assessment of the two benchmarks was confined to fundamental models with one phase of training. We leave for future work the investigation of diverging our architecture development to various training strategies, OCR qualities, domains, and languages.

\section*{Acknowledgement}
This research was supported by The VNUHCM-University of Information Technology's Scientific Research Support Fund.

\section*{Author Contributions Statement}

\textbf{Thanh-Phong Le:} Conceptualization; Data curation; Formal analysis; Investigation; Methodology; Validation; Visualization; Writing - original draft. \textbf{Trung Phan Le Chi:} Conceptualization; Data curation; Formal analysis; Investigation; Validation. \textbf{Nghia Hieu Nguyen:} Conceptualization; Data curation; Formal analysis; Investigation; Validation; Writing - review \& editing. \textbf{Kiet Van Nguyen:} Conceptualization; Formal analysis; Investigation; Methodology; Validation; Supervision; Writing - review \& editing.

\section*{Declaration of Competing Interest}
The authors declare that they have no known competing financial interests or personal relationships that could have appeared to influence the work reported in this paper.

\section*{Data Availability}
Data will be made available on reasonable request.


\begin{appendices}


\setcounter{table}{6}
\renewcommand{\thetable}{\arabic{table}}

\setcounter{figure}{16}
\renewcommand{\thefigure}{\arabic{figure}}

\section{Further dataset analysis on the Part-of-Speech tagging perspective}\label{sec:secA1}

To have an overview of language properties in our questions, we made use of PhoNLP \cite{phonlp}, a PhoBERT-based \cite{phobert} NLP model designed for core NLP tasks (Part-of-Speech (POS) tagging, named entity recognition (NER) and dependency parsing), to obtain POS tags in the questions. We additionally used VnCoreNLP tool \cite{vncorenlp} to segment the questions before feeding them to PhoNLP. A segmentation step essentially groups white-space-separated syllables into one multi-syllable word. It is a crucial pre-processing step to conduct Vietnamese core NLP tasks due to the monosyllabic attribute of the Vietnamese language, mentioned in \ref{sec:Q_classify}. 

\begin{figure*}[hbt]
\begin{subfigure}{.5\linewidth}
  \centering
  \includegraphics[width=1\linewidth]{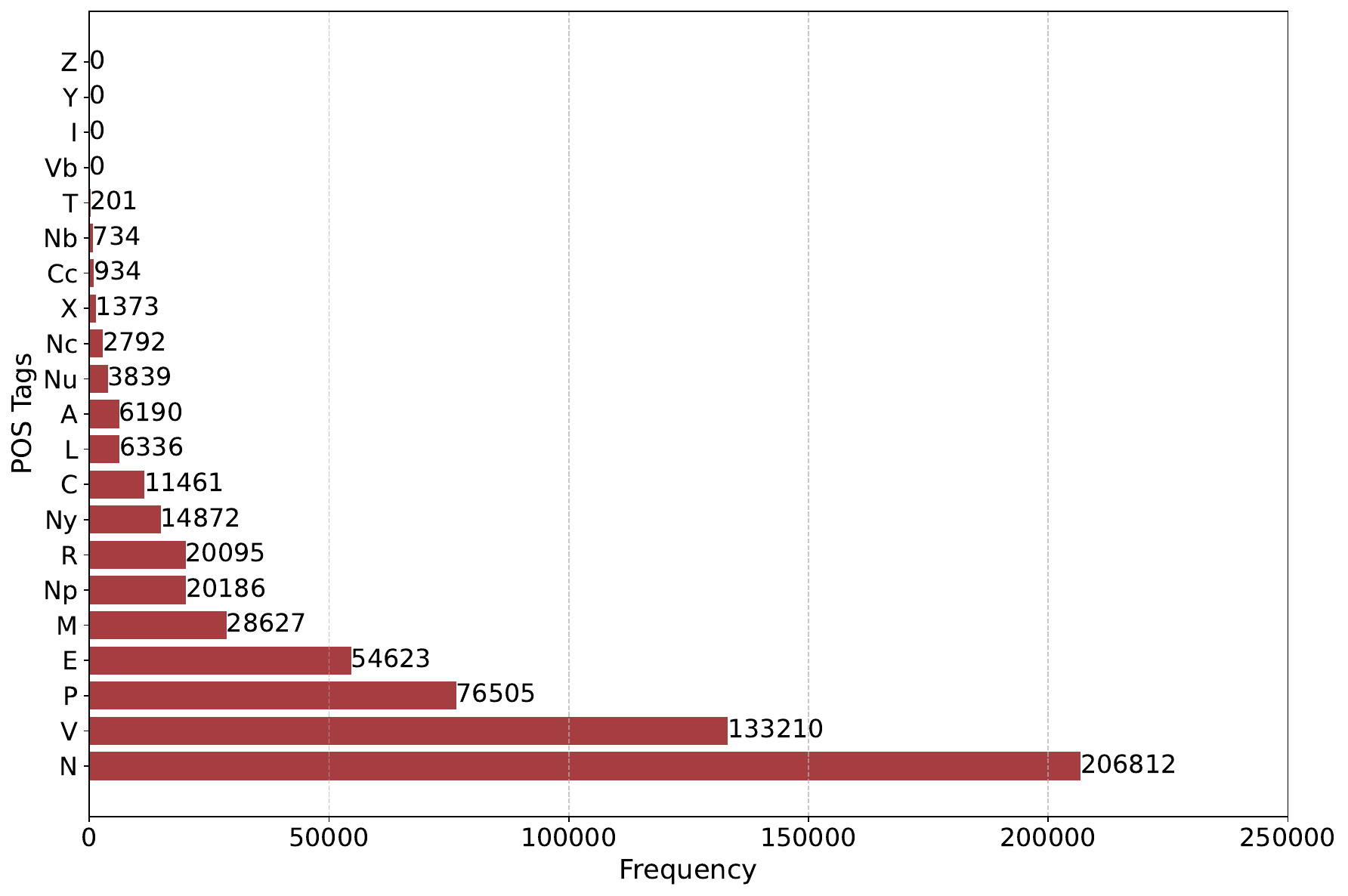}
  \caption{Distribution of each POS tag}
  \label{fig:POSQ1}
\end{subfigure}%
\begin{subfigure}{.5\linewidth}
  \centering
  \includegraphics[width=1\linewidth]{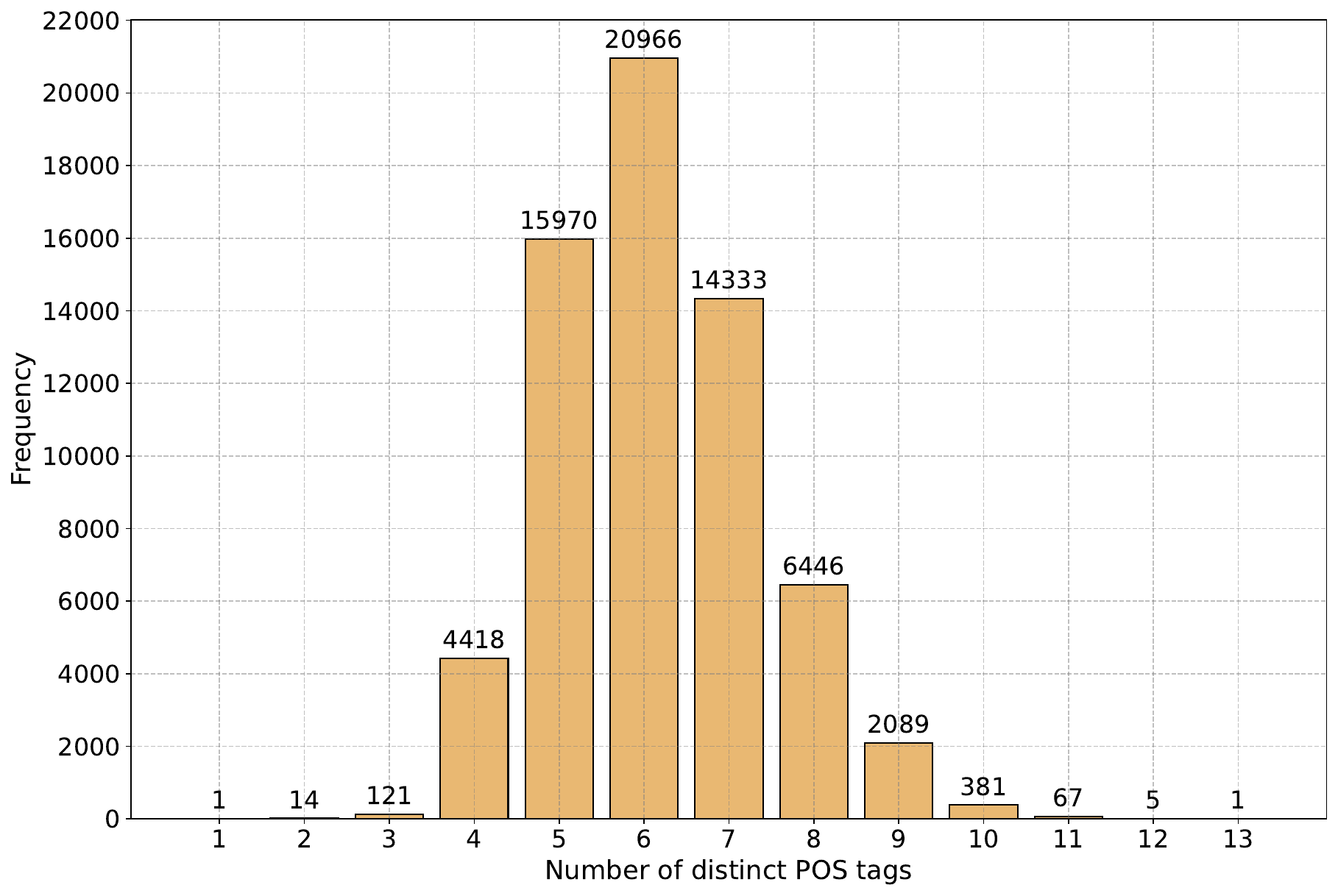}
  \caption{Distribution of distinct POS tags in a question}
  \label{fig:POSQ2}
\end{subfigure}
\caption{Distributions of POS tag in the questions}
\label{fig:QPOSdis}
\end{figure*}

Figure \ref{fig:QPOSdis} reveals characteristics of our dataset in the Part-of-Speech (POS) tagging perspective; definitions of the POS tags, originated from the VLSP 2013 POS tagging dataset \footnote{\url{https://vlsp.org.vn/vlsp2013/eval}}, are shown in Table \ref{tab:pos_defs}. As illustrated in Figure \ref{fig:POSQ1}, the questions encompass beyond 200,000 nouns (N), more than 125,000 verbs (V), and more than 75,000 pronouns (P), while excluding foreign/borrowed verb (Vb), interjection (I), abbreviation (Y), and bound morpheme (Z). Moreover, Figure \ref{fig:POSQ2} shows that there is a variety of POS tags in the questions with the majority consisting of from five to seven distinct tags.

\begin{table}[hbt]
    \centering
    \caption{POS Tags and Definitions, originated from the VLSP 2013 POS tagging dataset}
    \begin{tabularx}{\columnwidth}{c|l}
        \hline
        \textbf{Tag} & \textbf{Definition} \\
        \hline
        Np & Proper noun \\
        Nc & Classifier noun \\
        Nu & Unit noun \\
        N & Noun \\
        Ny & Abbreviated noun \\
        Nb & (Foreign) borrowed noun \\
        V & Verb \\
        Vb & (Foreign) borrowed verb \\
        A & Adjective \\
        P & Pronoun \\
        R & Adverb \\
        L & Determiner \\
        M & Numeral/Quantity \\
        E & Preposition \\
        C & Subordinating conjunction \\
        Cc & Coordinating conjunction \\
        I & Interjection/Exclamation \\
        T & Particle/Auxiliary, modal words \\
        Y & Abbreviation \\
        Z & Bound morpheme \\
        X & Un-definition/Other \\
        CH & Punctuation and symbols \\
        \hline
    \end{tabularx}
    \label{tab:pos_defs}
\end{table}

\section{Ablation Analysis on the layout understanding of LiGT model} \label{sec:secA2}

We further analyzed important factors of the LayoutHEI component, e.g., the number of levels for layout hashing and the learnable ratio ($\omega$) on both extractive and generative perspectives. To reduce computational cost and time, all models were applied with their base versions to create variants for the experiments.

\subsection{Impact of Hashed Levels on LayoutHEI-Employed Models}

Technically, the more hashed levels are employed, the more specifically the layout information could be represented. However, the correlation between specificity and performance could be different. For this reason, we created four variants applied from two to five hashed levels for each examined baseline and compared them with their corresponding text-only results. For extractive models, we used PhoBERT+LayoutHEI, mentioned in Section \ref{sec:Ef_LayoutHEI}, to compare with PhoBERT. For generative models, we compared our proposed model LiGT with its employed language model ViT5. We also considered the value of the ratio ($\omega$) after the training process to see how it adapts to each variant.

\begin{table*}[hbt]
    \centering
    \small
    \caption{LayoutHEI-Employed models' performances on different numbers of hashed levels. Values in parentheses show differences in model performances compared to corresponding scores of text-only models. All models are base versions}
    \resizebox{\textwidth}{!}{%
    \begin{tabular}{ccccccc}
         \hline
         \multirow{2}{*}{\textbf{Model}} & \textbf{Level} & \multicolumn{3}{c}{\textbf{Metrics}} & Ratio\\
         \cmidrule{3-5}
           & \textbf{(L)} & \textbf{ANLS} & \textbf{F1} & \textbf{Accuracy} & ($\omega$)\\
         \hline
         \multirow{4}{*}{PhoBERT+LayoutHEI} & \textit{2} & 60.70 (\textcolor{red}{$\downarrow$ 0.69}) & 56.20 (\textcolor{red}{$\downarrow$ 1.35}) & 53.18 (\textcolor{red}{$\downarrow$ 1.73}) & 0.5949\\  
                                                
                                            & \textit{3} & 61.55 (\textcolor{blue}{$\uparrow$ 0.16}) & 57.35 (\textcolor{red}{$\downarrow$ 0.20}) & 54.75 (\textcolor{red}{$\downarrow$ 0.16}) & 0.5792\\

                                            & \textit{4} & 63.11 (\textcolor{blue}{$\uparrow$ 1.72}) & 58.83 (\textcolor{blue}{$\uparrow$ 1.28}) & 55.26 (\textcolor{blue}{$\uparrow$ 0.35}) & 0.5737\\

                                            & \textit{5} & 61.40 (\textcolor{blue}{$\uparrow$ 0.01}) & 57.32 (\textcolor{red}{$\downarrow$ 0.23}) & 54.49 (\textcolor{red}{$\downarrow$ 0.42}) & 0.5654\\

         \hline
         \multirow{4}{*}{LiGT} & \textit{2} & 78.36 (\textcolor{blue}{$\uparrow$ 0.28}) & 67.60 (\textcolor{blue}{$\uparrow$ 0.56}) & 62.35 (\textcolor{blue}{$\uparrow$ 0.53}) & 0.5476\\  
                                                
                                            & \textit{3} & 79.07 (\textcolor{blue}{$\uparrow$ 0.99}) & 67.72 (\textcolor{blue}{$\uparrow$ 0.68}) & 62.32 (\textcolor{blue}{$\uparrow$ 0.50}) & 0.5491\\

                                            & \textit{4} & 78.78 (\textcolor{blue}{$\uparrow$ 0.70}) & 67.90 (\textcolor{blue}{$\uparrow$ 0.86}) & 62.63 (\textcolor{blue}{$\uparrow$ 0.81}) & 0.5492\\

                                            & \textit{5} & 78.32 (\textcolor{blue}{$\uparrow$ 0.24}) & 67.60 (\textcolor{blue}{$\uparrow$ 0.56}) & 62.29 (\textcolor{blue}{$\uparrow$ 0.47}) & 0.5529\\
         \hline
    \end{tabular}%
    }
    \label{tab:effect_hash_level}
\end{table*}

Table \ref{tab:effect_hash_level} shows that an adequate number of hashed levels enhanced the performances. In particular, performances of variants of the two architectures increased when applying from two levels to four levels, and decreased when we continued to apply five levels. Applying four hashed levels yielded the highest results, followed by applying three levels. In addition, it can be seen that the generative model was more stable than the extractive model, presented by the performance gaps between the variants.

Besides, the ratio ($\omega$) shows an interesting pattern apart from its distribution between 0.54 to 0.6. While the ratio of PhoBERT+LayoutHEI dropped slightly as we increased the number of levels, the ratio of LiGT had an upward trend.

\subsection{Impact of Applying Ratio ($\omega$) in the LayoutHEI-Employed Models}

To have a clearer view of using the ratio ($\omega$), we continued to evaluate variants of our PhoBERT+LayoutHEI and LiGT, wherein the ratios are removed. Our configuration in this experiment is similar to setting the ratio to 1.0 all the time. The experiments were examined on three and four hashed levels due to their prominent performances in prior evaluation, shown in Table \ref{tab:effect_hash_level}. In this experiment, we compared the variants without the ratio ($\omega$) with those having full settings from Table \ref{tab:effect_hash_level}.

\begin{table}[hbt]
    \centering
    \caption{Performances of 3-level LayoutHEI-employed models without the ratio $\omega$, denoted by \textit{*}. Values in parentheses show differences in the model performances compared to scores of their corresponding models applied the ratio. All models are base versions}
    
        \begin{tabularx}{\columnwidth}{>{\centering\arraybackslash}X >{\centering\arraybackslash}X >{\centering\arraybackslash}X}
             \hline
             \multirow{3}{*}{\textbf{Metric}}  & \multicolumn{2}{c}{\textbf{Models}}\\
             \cmidrule{2-3}
              &  \textbf{PhoBERT} & \textbf{LiGT*}\\  
              & \textbf{LayoutHEI*} & \\
             \hline
             ANLS & 62.17 (\textcolor{blue}{$\uparrow$ 0.62}) & 77.99 (\textcolor{red}{$\downarrow$ 1.08}) \\
             F1 & 58.14 (\textcolor{blue}{$\uparrow$ 0.79}) & 67.28 (\textcolor{red}{$\downarrow$ 0.44}) \\
             Accuracy & 54.89 (\textcolor{blue}{$\uparrow$ 0.14}) & 61.58 (\textcolor{red}{$\downarrow$ 0.74})\\
             \hline
        \end{tabularx}
    \label{tab:effect_3hash_wo_ratio}
\end{table}

\begin{table}[hbt]
    \centering
    \caption{Performances of 4-level LayoutHEI-employed models without the ratio $\omega$, denoted by \textit{*}. Values in parentheses show differences in the model performances compared to scores of their corresponding models applied the ratio. All models are base versions}
    
        \begin{tabularx}{\columnwidth}{>{\centering\arraybackslash}X >{\centering\arraybackslash}X >{\centering\arraybackslash}X}
             \hline
             \multirow{3}{*}{\textbf{Metric}}  & \multicolumn{2}{c}{\textbf{Models}}\\
             \cmidrule{2-3}
              &  \textbf{PhoBERT} & \textbf{LiGT*}\\  
              & \textbf{LayoutHEI*} & \\
             \hline
             ANLS & 61.70 (\textcolor{red}{$\downarrow$ 1.41}) & 78.45 (\textcolor{red}{$\downarrow$ 0.33}) \\
             F1 & 57.59 (\textcolor{red}{$\downarrow$ 1.24}) & 67.68 (\textcolor{red}{$\downarrow$ 0.22}) \\
             Accuracy & 54.35 (\textcolor{red}{$\downarrow$ 0.91}) & 62.26 (\textcolor{red}{$\downarrow$ 0.37})\\
             \hline
        \end{tabularx}
    \label{tab:effect_4hash_wo_ratio}
\end{table}

Table \ref{tab:effect_3hash_wo_ratio} and Table \ref{tab:effect_4hash_wo_ratio} present that removing the adaptive ratio could hinder models' performance. In particular, LiGT witnessed a considerable decline in its both two variants in all metrics. Additionally, while the variant of PhoBERT+LayoutHEI having three levels achieved marked improvements, there was a significant drop in the variant applied four levels. The results demonstrated that the absence of ratio ($\omega$) could either cause unstable or poor performances, making the layout understanding ability sub-optimal.

\section{Effect of Visual Specialties on Generative Baselines} \label{sec:secA3}

For experiments on visual impact, we replace the visual components of ViT5+U, ViT5+2D+U, and LaTr with our chosen visual sources. In particular, we made use of features of VietOCR \footnote{\url{https://github.com/pbcquoc/vietocr}}, a widely-used OCR extraction tool for Vietnamese, to compare with U-Net in ViT5+U and ViT5+2D+U; we applied DiT \cite{dit} in comparison with ViT in LaTr. We leverage VietOCR's pretrained model having the highest performance (the combination of VGG net \cite{vgg} and Transformer \cite{transformer}). We cropped out portions of text in the image. These cropped images were then projected through VGG net and encoder layers of the chosen model to gain visual features to replace U-Net. About LaTr, DiT is a transformer-based model pretrained on the document domain. We replaced ViT in LaTr with DiT to assess the impact of domain-specific factors on the model. To reduce computational cost and time, all models were applied with their base versions.

\begin{table}[hbt]
    \centering
    \caption{Effect of varying sources of vision modality when replacing U-Net with VietOCR features in ViT5+U and ViT5+2D+U. All models are base versions}
    
        \begin{tabularx}{\columnwidth}{>{\centering\arraybackslash}X >{\centering\arraybackslash}X >{\centering\arraybackslash}X}
             \hline
             \multirow{3}{*}{\textbf{Metric}}  & \multicolumn{2}{c}{\textbf{Models}}\\
             \cmidrule{2-3}
              &  \multirow{2}{*}{\textbf{ViT5+VietOCR}} & \textbf{ViT5+2D}\\
              & & \textbf{+VietOCR}\\
             \hline
             ANLS & 79.06 (\textcolor{blue}{$\uparrow$ 0.08}) & 76.39 (\textcolor{blue}{$\uparrow$ 0.48}) \\
             F1 & 67.75 (\textcolor{red}{$\downarrow$ 0.14}) & 64.93 (\textcolor{red}{$\downarrow$ 0.15}) \\
             Accuracy & 62.38 (\textcolor{red}{$\downarrow$ 0.08}) & 59.66 (\textcolor{red}{$\downarrow$ 0.26})\\
             \hline
        \end{tabularx}
    \label{tab:effect_unet}
\end{table}

\begin{table}[hbt]
    \centering
    \caption{Effect of varying sources of vision modality when replacing ViT with DiT in LaTr. All models are base versions}
    
        \begin{tabularx}{\columnwidth}{>{\centering\arraybackslash}X >{\centering\arraybackslash}X >{\centering\arraybackslash}X >{\centering\arraybackslash}X}
             \hline
             \textbf{Model} & \textbf{ANLS} & \textbf{F1} & \textbf{Accuracy} \\
             \hline
             \multirow{2}{*}{LaTr(DiT)} & 74.27  & 62.00  & 56.65 \\
             & (\textcolor{blue}{$\uparrow$ 0.29}) & (\textcolor{blue}{$\uparrow$ 0.47}) & (\textcolor{blue}{$\uparrow$ 0.28}) \\
             \hline
        \end{tabularx}
    \label{tab:effect_vit}
\end{table}

As shown in Table \ref{tab:effect_unet} and Table \ref{tab:effect_vit}, although ViT5+VietOCR achieved ANLS higher than 79\%, its F1 and Accuracy score had a slight drop. ViT5+2D+VietOCR also followed the same pattern despite having an increase in ANLS by nearly 0.5\%. Regarding LaTr, while using a domain-specific visual module like DiT showed improvements in all metrics, the differences did not make it competitive with other generative baselines. These observations suggest that the impact of visual modalities on the receipt domain in Vietnamese might have certain limits and need more investigation on both enriching visual sources and developing visual integrating methodologies.

\section{Detail of the evaluation of LiGT on DocVQA and InfographicVQA} \label{sec:secA4}

\subsection{Configuration of LiGT language model components on DocVQA and InfographicVQA} \label{sec:secA4_1}

\begin{table}[hbt]
    \centering
    \caption{Configuration of LiGT language models for DocVQA and InfographicVQA. All models are base versions}
    
        \begin{tabularx}{\columnwidth}{>{\centering\arraybackslash}X >{\centering\arraybackslash}X >{\centering\arraybackslash}X >{\centering\arraybackslash}X}
             \hline
             \textbf{Config} & \textbf{T5} & \textbf{BART} & \textbf{Flan-T5} \\
             \hline
             Learning rate & 1e-3 & 5e-5 & 1e-4 \\
             \hline
             Max token & 512 & 512 & 450 \\
             \hline
             Warmup step & \multicolumn{3}{c}{1000} \\
             \hline
             Epoch & \multicolumn{3}{c}{10} \\
             \hline
             Hashed level & \multicolumn{3}{c}{4} \\
             \hline
        \end{tabularx}
    \label{tab:config_standard_docvqa}
\end{table}

\subsection{LiGT performance on different aspects of DocVQA and InfographicVQA} \label{sec:secA4_2}

\begin{table}[hbt]
    \centering
    \caption{Details of LiGT performance using different language models on the aspects of DocVQA. Names of the aspects are from the Robust Reading Competition website. All models are base versions}
    
        \begin{tabularx}{\columnwidth}{>{\centering\arraybackslash}X >{\centering\arraybackslash}X >{\centering\arraybackslash}X >{\centering\arraybackslash}X}
             \hline
             \textbf{Aspect} & \textbf{T5} & \textbf{BART} & \textbf{Flan-T5} \\
             \hline
             Figure/Diagram & 26.28 & 30.90 & 32.44 \\
            \hline
            Form & 72.74 & 72.32 & 75.80\\
            \hline
            Table/List & 51.78 & 55.13 & 57.51\\
            \hline
            Layout & 61.73 & 61.19 & 67.73\\
            \hline
            Free text & 57.61 & 60.71 & 68.07\\
            \hline
            Image/Photo & 24.89 & 36.48 & 38.43\\
            \hline
            Handwritten & 51.44 & 51.15 & 51.51\\
            \hline
            Yes/No & 37.93 & 44.83 & 62.07\\
            \hline
            Others & 47.67 & 46.23 & 55.63\\
             \hline
        \end{tabularx}
    \label{tab:detail_docvqa}
\end{table}

\begin{table}[hbt]
    \centering
    \caption{Details of LiGT performance using different language models on the aspects of InfographicVQA. Names of the aspects are from the Robust Reading Competition website. All models are base versions}
    
        \begin{tabularx}{\columnwidth}{>{\centering\arraybackslash}X >{\centering\arraybackslash}X >{\centering\arraybackslash}X >{\centering\arraybackslash}X}
             \hline
             \textbf{Aspect} & \textbf{T5} & \textbf{BART} & \textbf{Flan-T5} \\
             \hline
             \multicolumn{4}{c}{\textit{Answer type}} \\
             \hline
            Image span & 22.85 & 29.08 & 32.57\\
            Question span & 48.53 & 46.76 & 54.09\\
            Multiple spans & 6.44 & 10.96 & 15.82\\
            Non span & 18.53 & 17.75 & 20.49\\
            \hline
             \multicolumn{4}{c}{\textit{Evidence}} \\
             \hline	
            Table/List & 20.48 & 23.74 & 29.2\\
            Textual & 27.10 & 32.60 & 36.12\\
            Visual object & 20.07 & 20.01 & 26.91\\
            Figure & 20.57 & 26.08 & 28.46\\
            Map & 16.54 & 18.49 & 29.49\\
            \hline
             \multicolumn{4}{c}{\textit{Operation}} \\
             \hline			
            Comparison & 20.51 & 23.93 & 31.70\\
            Arithmetic & 19.40 & 18.29 & 20.40\\
            Counting & 17.47 & 17.37 & 20.77\\
             \hline
        \end{tabularx}
    \label{tab:detail_infovqax}
\end{table}




\end{appendices}


\bibliography{sn-article}

\end{document}